\documentclass{article}

\usepackage{arxiv}
\usepackage[square,numbers, compress]{natbib}
\usepackage[utf8]{inputenc} 
\usepackage[T1]{fontenc}    
\usepackage{hyperref}       
\usepackage{url}            
\usepackage{booktabs}       
\usepackage{amsmath,amssymb,amsfonts,mathtools,bbm}       
\usepackage{nicefrac}       
\usepackage{microtype}      
\usepackage{xcolor}         
\usepackage{graphicx}
\usepackage{algorithm}
\usepackage[noend]{algorithmic}
\usepackage{subcaption}

\newcommand{\RR}{\mathbb{R}}

\newcommand{\PP}{\mathbb{P}}

\newcommand{\EE}{\mathbb{E}}

\newcommand{\calC}{\mathcal{C}}
\newcommand{\calE}{\mathcal{E}}
\newcommand{\calF}{\mathcal{F}}
\newcommand{\calL}{\mathcal{L}}

\newcommand{\bb}[1]{\left( #1\right)}
\newcommand{\closedb}[1]{\left[ #1\right]}
\newcommand{\bs}[1]{\left\{ #1\right\}}

\newcommand{\norm}[1]{\left\| #1 \right\|}
\newcommand{\innerprod}[1]{\left\langle #1 \right\rangle}

\title{Deep-MacroFin: Informed Equilibrium Neural Network for Continuous-Time Economic Models}

\author{%
Yuntao Wu\\
Electrical and Computer Engineering\\
University of Toronto\\
Toronto, Ontario, Canada\\
  \texttt{winstonyt.wu@mail.utoronto.ca} \\
\And
Jiayuan Guo \\
Joseph L. Rotman School of Management\\
University of Toronto\\
Toronto, Ontario, Canada \\
\texttt{guoflora5@gmail.com} \\
\And
Goutham Gopalakrishna \\
Joseph L. Rotman School of Management\\
University of Toronto\\
Toronto, Ontario, Canada \\
\texttt{goutham.gopalakrishna@rotman.utoronto.ca} \\
\And
Zissis Poulos \\
School of Information Technology\\
York University\\
Toronto, Ontario, Canada \\
\texttt{zpoulos@yorku.ca} \\
}

\begin{document}

\maketitle

\begin{abstract}
In this paper, we present Deep-MacroFin, a comprehensive framework designed to solve partial differential equations, with a particular focus on models in continuous time economics. This framework leverages deep learning methodologies, including Multi-Layer Perceptrons and the newly developed Kolmogorov-Arnold Networks. It is optimized using economic information encapsulated by Hamilton-Jacobi-Bellman (HJB) equations and coupled algebraic equations.
The application of neural networks holds the promise of accurately resolving high-dimensional problems with fewer computational demands and limitations compared to other numerical methods.
This framework can be readily adapted for systems of partial differential equations in high dimensions. Importantly, it offers a more efficient (\textbf{5}$\times$ less CUDA memory and \textbf{40}$\times$ fewer FLOPs in \textbf{100D} problems) and user-friendly implementation than existing libraries. We also incorporate a time-stepping scheme to enhance training stability for nonlinear HJB equations, enabling the solution of \textbf{50D economic models}.
\end{abstract}

\section{Introduction}
Partial Differential Equations (PDEs) represent a class of mathematical equations that encapsulate the rates of change with respect to continuous variables. These equations are ubiquitous in the fields of physics and engineering, offering succinct insights into phenomena pertaining to acoustics, thermodynamics, electrodynamics, etc., where closed-form analytical solutions may not always be found.
In the realm of macroeconomics and finance, PDEs are used to model and forecast complex phenomena like economic growth, inflation, interest rates, and asset prices. General equilibrium problems in these fields are typically governed by nonlinear elliptical PDEs \cite{Brunnermeier2014, Brunnermeier2016, yves2017, gomez2017}.
The ability to numerically solve PDE systems allows us to scrutinize the effects of parameter alterations on the equilibrium state.

Given the inherent complexity in deriving exact analytic solutions to PDEs, particularly for nonlinear PDE problems, researchers frequently resort to numerical techniques, such as the Finite Difference Method (FDM) \cite{grossmann2007FDM, boyce2012FDM}, and the Finite Element Method (FEM) \cite{solín2005FEM, quarteroni2008FEM}. 
The fundamental concept behind these numerical solutions involves discretizing the continuous domain into a grid (mesh) of finite elements and approximating derivatives using differences between adjacent points. While these methods can yield accurate solutions with careful grid choices, they may encounter instability and high computational costs, particularly for higher-dimensional problems or complex physical systems \cite{stability1, stability2}.
D'Avernas et al. demonstrate that standard FDM can fail when solving economic problems, even with only two state variables \cite{pymacrofinsolutionmethod}. Moreover, FDM is not scalable, due to the high nonlinearity of PDEs governing economic equilibrium models and the computational infeasibility of solving large linear systems when an implicit scheme is employed. However, these traditional numerical schemes provide insights into handling nonlinearity by introducing a transient time dimension \cite{Brunnermeier2014,Brunnermeier2016,pymacrofinsolutionmethod,ditella2017}.

Deep learning, particularly deep neural networks, has been used for a variety of tasks across multiple domains, including regression, classification, and the generation of images and natural language \cite{deep-learning}. 
Recently, the use of deep learning to solve partial differential equations (PDEs) has emerged as a promising alternative to traditional numerical solutions \cite{Sirignano2018,scientific-ml,deepxde}. This approach leverages the theorem that neural networks can serve as universal function approximators \cite{universal-approx}. 
The primary methodology involves Physics-Informed Neural Networks (PINNs), which optimize neural networks using PDEs as loss functions to approximate solutions \cite{pinn2017, pinn2019}. 
However, to the best of our knowledge, deep learning has not been extensively utilized to solve equilibrium problems in continuous time economics, compared to the PINN literature.\footnote{This area is still nascent, albeit growing over time.} 
These problems typically involve optimizing Hamilton-Jacobi-Bellman (HJB) equations, which often lack classical smooth solutions \cite{hjb1, hjb2}. Additionally, they are coupled with a system of algebraic equations derived from market clearing conditions, binding constraints, and financial frictions. Solving such a system of PDEs can be numerically unstable, while neural networks could potentially offer improved approximations.

In this paper, we present Deep-MacroFin, a comprehensive framework for solving PDEs, with a specific focus on continuous-time economic models. 
Our approach offers several advantages over numerical methods: (1)  it accommodates higher dimensionality with sparse sampling. Specifically, we demonstrate the ability to solve a 100-dimensional Laplace equation using 4.3GB CUDA memory, 12 GFLOPs, with an average epoch time of 0.16 seconds; (2) it handles differentiation more accurately without explicit discretization; and (3) once the problem is solved, the solution can be accurately extrapolated to a larger domain, free from grid space constraints.
Furthermore, our methods outperform existing neural network techniques in several ways: (1) they offer simpler, more user-friendly implementation, with support for string and \LaTeX\ input; (2) they can readily solve various types of PDEs such as free boundary models; and (3) they allow for flexible initial / boundary conditions, allowing for a shift in learning regions and accommodating various boundary conditions in high dimensions. Finally, (4) we incorporate a time-stepping scheme from traditional numerical methods, inspired by \cite{ALIENs}, leading to more stable training in macro-finance models involving HJB equations. We solve economics models upto 50D with time-stepping scheme.

\section{Related work}
While several libraries exist for numerically solving PDEs, many either struggle with high-dimensional settings or are tailored to physical sciences rather than economics. Deep-MacroFin addresses this gap by offering a user-friendly, end-to-end solution for solving economic equilibrium models governed by PDEs. We benchmark our method against two representative libraries:

\textbf{PyMacroFin \cite{pymacrofin}.} This library targets macro-finance equilibrium models in continuous time, typically with one or two state variables, using finite difference methods and implicit schemes based on \cite{Brunnermeier2014}. To address the nonlinearity of HJB equations, it reformulates them as auxiliary linear parabolic PDEs, solved through transient time iterations. However, PyMacroFin is limited in scope—it supports only low-dimensional problems and a narrow class of economic PDE systems. Numerical instability and the curse of dimensionality make it unsuitable for more general or higher-dimensional models. In some of our experiments, we implement finite difference methods directly, rather than relying on PyMacroFin.

\textbf{DeepXDE \cite{deepxde}.} This is a general-purpose PINN library for solving both forward and inverse PDE problems with given initial and boundary conditions. It has been successfully applied in areas such as option pricing \cite{deepxde-option-pricing} and some economic dynamics \cite{deepxde-macro-problem}. Despite its flexibility, DeepXDE struggles with high-dimensional problems and lacks tailored features for economic models, limiting its applicability to complex equilibrium systems (see Section~\ref{sec:experiments}).

\section{Methodology}

\subsection{Problem formulation}\label{sec:problem}
Following \cite{Brunnermeier2014, Brunnermeier2016}, we consider two types of agents: experts ($e$) and households ($h$), each with distinct consumption preferences and stochastic differential utility \cite{Duffie1992StochasticDU}. Time is continuous and the horizon is infinite. In equilibrium, (1) agents maximize expected utility subject to consumption, asset holdings, and budget constraints, and (2) markets clear.
Let $Z_t$ be a Brownian motion on a filtered probability space $\bb{\Omega, \calF, \bs{\calF_t}_{t\geq 0}, \PP}$. The exogenous state variable $x\in\RR^n$ evolves according to the SDE: 
\begin{equation}
    dx_t = \mu_x dt + \sigma_x dZ_t
\end{equation}
Each agent solves the following generic optimization problem: 
\begin{align*}
    \sup_{c_t^j}\; &\EE_T^{\PP} \closedb{\int_T^\infty e^{-\rho t} u(c_t)dt},\\
    \text{s.t. } & f(x) \leq 0, g(x) =0,
\end{align*}
where $c_t$ denotes the consumption policy, and $u(c_t)$ the instantaneous utility. The constraint functions $f: \RR^n \to \RR^m$ and $g: \RR^n \to \RR^k$ encode budget, asset, and market clearing conditions, depending on endogenous variables such as capital prices $q_t$.
Let $V_j(x)$ be the value function for agent $j$. Applying Itô’s lemma yields the associated HJB equation:
\begin{equation}\label{eq:hjb}
    \rho V_j(x) = \sup_{c} \bs{u(c) + \nabla_x V_j^T \mu_x + \frac{1}{2} \sigma_x^T H_x(V_j) \sigma_x},
\end{equation}
where $\nabla_x V_j$ and $H_x(V_j)$ are the gradient and Hessian of $V_j$, respectively.
To approximate equilibrium quantities $V$, $c$, and $q$, we reformulate the optimization problem as a residual system:
\begin{equation}
    \calE_k\left(x; V_j, c_j, q;\nabla_x V_j, H(V_j); \mathbf{\lambda} \right) = 0,
\end{equation} 
where $\lambda=\bs{a_e, a_h,...}$ are parameters. The goal is to minimize the residuals, $\calL_{\calE_k} = \norm{\calE_k}^2$.

\subsection{System design}

\begin{figure*}[tb]
    \centering
    \includegraphics[width=0.9\textwidth]{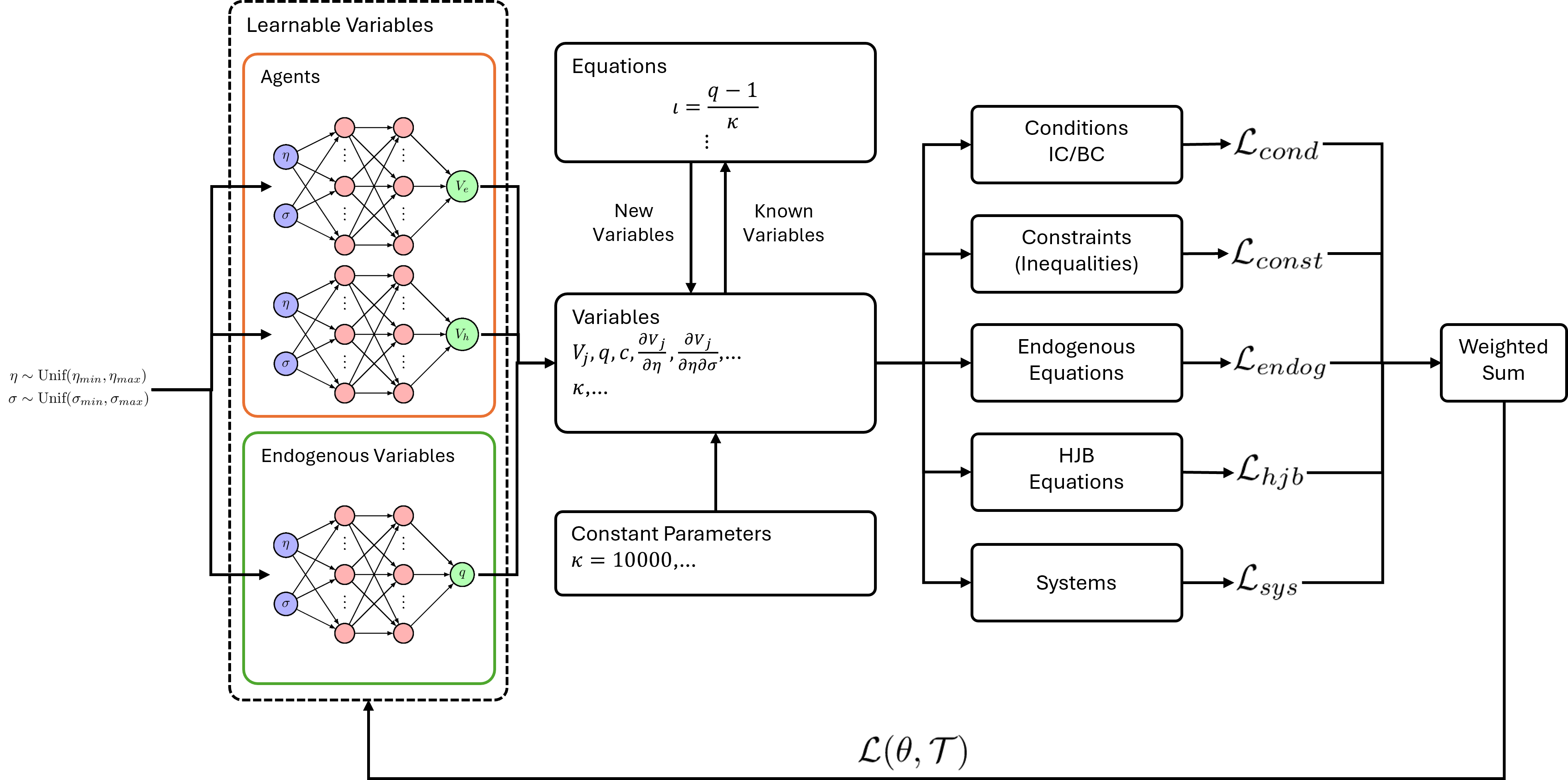}
    \caption{System overview. This model assumes the presence of two state variables: $\eta$ (experts' wealth share) and $\sigma$ (volatility), two agents: $V_e$ (experts) and $V_h$ (households), and one endogenous variable: $q$ (price of the capital).}
    \label{fig:system}
\end{figure*}

Agent value functions $V_j$, and endogenous variables such as $q$, as outlined in the previous section, are approximated by neural networks, guided by endogenous and HJB equations. 
The framework also supports initial/boundary conditions, inequality constraints, and systems activated by constraints, enabling applications to basic PDEs, free boundary and variational problems.
Let $\theta$ represent the neural network parameters and $\mathcal{T}$ be the training data. The total loss is:
\begin{align}\label{eq:total-loss}
    \mathcal{L}(\theta, \mathcal{T}) &= \sum_i \lambda_i \mathcal{L}_{i}(\theta, \mathcal{T}),
\end{align}
which is a weighted sum of losses from all conditions, constraints, endogenous equations, and HJB equations, with each loss computation detailed in subsequent paragraphs. This weighted loss guides the neural network to learn the equilibrium solution without any assumption about the solution itself. 
Weights $\lambda_i$ can be adjusted dynamically via learnable weights \cite{loss-attention} or heuristic adaptation \cite{loss-balancing}, though in this paper we manually select them for best empirical performance, as automatic methods proved less robust.
Models that minimize $\mathcal{L}(\theta,\mathcal{T})$ ensures the approximated functions satisfy the specified equations and constraints.
Figure~\ref{fig:system} offers an overview of the system with two state variables. Further details on each component and the training process will be elaborated in the following paragraphs.

\paragraph{State variables and parameters}
State variables $X=(x_1,...,x_d)\in\mathbb{R}^d$ define the dimensionality and domain of the problem. In physical problems, they may represent space-time coordinates, while in economic problems, they could denote wealth shares or capital return volatility. Users can specify a domain $x_i\in [low_i, high_i]$, and values are sampled during training from user-defined distributions. Sampling can be uniform or follow an ergodic distribution, as in \cite{dp-dynamic, hd-dp}. 
Models are also governed by constant hyper-parameters. In the context of economic models, these could encompass factors such as relative risk aversion ($\gamma$), discount rate ($\rho$), etc. Different parameters can yield different solutions for the same underlying model. With neural networks, it is also possible to extend to more practical applications by treating fixed parameters as pseudo-variables and estimating their domains using real-world data. 

\paragraph{Learnable variables}
Learnable variables, including agent value functions and endogenous variables, are unknown functions $f:\mathbb{R}^n\to\mathbb{R}^m$ to be approximated by neural networks. 
These learnable variables are modeled as configurable deep neural networks. The number of hidden units, hidden layers and outputs, and types of activation functions are customizable. 
Currently, Multi-Layer Perceptrons (MLPs) \cite{deep-learning} and Kolmogorov-Arnold Networks (KANs) \cite{kan} are supported. While other architectures like DeepONet \cite{deeponet} can be integrated, they are outside the scope of this work.

The universal approximation theorem underlies MLP \cite{universal-approx}. Automatic differentiation with batched Jacobian and Hessian enables precise and efficient calculation of derivatives \cite{pytorch2, pytorch-autodiff}. Furthermore, \citet{Sirignano2018} shows that MLPs can approximate PDE solutions without knowing the exact form. In comparison, KAN is based on Kolmogorov–Arnold representation theorem. It is claimed to outperform MLP in terms of accuracy and interpretability. The interpretability stems from the symbolic formulas in KAN approximations, which can be easily extracted and evaluated. However, recent research suggests that KAN requires further improvements to match MLP in solving PDEs due to its lack of robustness and computational parallelism \cite{kan-mlp}.

\paragraph{Equations}
The Equation module is used to define new variables. In economic models, the endogenous or HJB equations guiding the equilibrium are unlikely to directly depend on the agent value functions and endogenous variables. The Equation module provides a straightforward method to define intermediate variables. 
Given an equation $l=r(x)$, $l$ is stored as a new variable in the system with an initial value of zero. During each iteration of training and testing, $r(x)$ is evaluated using known variables, and the resulting value is assigned to $l$. 
For example, the equation $\iota = \frac{q-1}{\kappa}$ in Figure~\ref{fig:system} defines a new variable $\iota$ (investment). Its value is updated to $\frac{q-1}{\kappa}$ in each iteration.
User-defined formulas can be raw Python code or \LaTeX-based formula. The parsing of \LaTeX\ is based on regular expression, independent of external libraries. This allows users to easily transfer their formulas from \LaTeX\ documents to Python code.

\paragraph{Conditions}
Each learnable variable $v_i\in \{a_1,...,a_n, e_1,...,e_m\}$, either an agent value function or an endogenous variable, can be associated with specific conditions $\mathcal{C}(v_i, x)=0$. 
In the context of mathematical or physical PDE problems, these conditions could represent initial value conditions $v_i(x_0)=v_{i0}$ or boundary value conditions $\mathcal{B}(v_i, x) = 0$, where $x\in \partial \Omega$, and $\Omega\subset \mathbb{R}^d$ is the problem domain.
These conditions could be extended to any subsets $U\subset \Omega\cup \partial \Omega$ that require accurate approximation. 
These constraints are incorporated as a Mean Squared Error (MSE) over $U$: $\mathcal{L}_{cond} = \frac{1}{|U|} \sum_{x\in U}\|\mathcal{C}(v_i, x)\|_2^2$.

\paragraph{Endogenous equations}
The Endogenous Equation module is used to establish equalities that are required to pin down endogenous variables in the system. Typically, an endogenous equation takes the form of an algebraic (partial differential) equation $l(x)=r(x)$. Each endogenous equation is converted to a MSE loss over a batch of size $B$: $\mathcal{L}_{endog} = \frac{1}{B} \|l(x)-r(x)\|_2^2$.

\paragraph{HJB equations} 
The HJB Equation module is used to inform the neural networks of the heterogenous agent asset pricing models. Unlike PyMacroFin, which linearizes the HJB equation, Deep-MacroFin allows for direct input of HJB equations in the form of \eqref{eq:hjb} to pin down each agent. As we aim for the optimal value of the HJB equation $\sup\{\text{HJB}(x)\}$ to be zero, the MSE loss can be computed over a batch of size $B$ as: $\mathcal{L}_{hjb} = \frac{1}{B} \|\text{HJB}(x)\|^2$.
The optimality conditions are computed using first-order conditions and are enforced using equations and endogenous equations. With different loss functions, this module can also be applied to solve variational problems as shown in Section~\ref{sec:free-boundary}.

\paragraph{Constraints}
The Constraint module is used to impose inequality conditions that restrict the solution space of the model. It enables the neural network to account for scenarios where valid solutions must lie within a specific subset of the full function space spanned by the network. For $l(x)\leq r(x)$, a rectified MSE is computed over a batch of size $B$: $\mathcal{L}_{const} = \frac{1}{B} \|\text{ReLU}(l(x)-r(x))\|_2^2$, where $\text{ReLU}(x)=\max(x, 0)$. Therefore, loss is only computed for $x\in B$, where $l(x) > r(x)$, \textit{i.e.} when the constraint $l(x)\leq r(x)$ is violated. 
For $l(x)\geq r(x)$, the rectified MSE is computed as $\mathcal{L}_{const} = \frac{1}{B} \|\text{ReLU}(r(x)-l(x))\|_2^2$.
In the case of strict inequalities, $l(x)>r(x)$ or $l(x)<r(x)$, an additional $\epsilon = 10^{-8}$ is added to the difference within ReLU to ensure strict inequalities.

\paragraph{Systems}
Systems are activated only when the binding constraints are satisfied. For a batch of inputs, both constraint-governed equations and endogenous equations are computed for each input in the batch. If an input does not satisfy the constraint, it is excluded from the loss computation. Equations are used to assign new variables and losses are computed based on the associated endogenous equations. 
Let $\mathbbm{1}_{mask}$ be a vector of zeros and ones indicating which inputs in the batch meet the constraints. Then loss for a specific endogenous equation $i$ in the system is $\mathcal{L}_{endog, i} = \frac{1}{\sum \mathbbm{1}_{mask}} \langle (l-r)^2, \mathbbm{1}_{mask}\rangle$,
where $(l-j)^2$ is the element-wise square of $l-r$, and $\langle \cdot, \cdot \rangle$ denotes the inner product. Essentially, this is the MSE loss computed on mask-selected inputs.
Let $\lambda_i$ be the weight associated with each endogenous equation, and $N$ be the number of endogenous equations attached to the system. The total loss of the system is $\mathcal{L}_{sys} = \sum_{i=1}^N \lambda_i \mathcal{L}_{endog, i}$.

\subsection{Training}
If the learnable variables are exclusively defined using MLPs, the neural networks can be trained using L-BFGS \cite{lbfgs}, Adam \cite{Adam}, or AdamW \cite{AdamW} algorithms. 
If any learnable variables are defined using KANs, then the KAN-customized L-BFGS algorithm \cite{kan} is employed. The neural networks are trained for a pre-defined number of epochs. 
To ensure reproducibility, the random seed is set to zero before training. 
Various training strategies can be employed depending on the problem context. A standard approach for a single epoch is outlined in Algorithm~\ref{algo:basic-training-step} in Appendix~\ref{appendix:training-algorithms}. The objective is to approximate the optimal neural network parameters $\theta^* = \arg\min \mathcal{L}(\theta, \mathcal{T})$. Upon training completion, both the model with the lowest loss and the final epoch model are saved for analysis.

\subsection{Time-stepping scheme}\label{sec:time-stepping-scheme}
The standard training approach works well for many simple PDE systems but often fails for nonlinear HJB equations, as we demonstrate in Section~\ref{sec:ncg}. Traditional numerical schemes for macro-finance problems \cite{Brunnermeier2014,Brunnermeier2016,pymacrofinsolutionmethod,ditella2017,ALIENs} introduce a transient but finite time dimension, analogous to value function iteration in discrete time.
Rather than modeling an infinite-horizon economy, we consider a finite horizon $[0,T]$, with value functions and endogenous variables parametrized as $V_j(x, t)$ and $E_j(x,t)$. Given terminal conditions at $T$, we compute the equilibrium over $[0,T]$. By iteratively updating the boundary conditions, setting $V_j(x,T)\gets V_j(x, 0)$ and $E_j(x,T)\gets E_j(x,0)$, equivalently taking $T\to \infty$, the behavior at $t=0$ converges to the infinite-horizon equilibrium.
This approach modifies the HJB equation via Itô’s lemma, introducing a time derivative and making the HJB quasi-linear:
\begin{align*}
    \rho V_j(x, t) = \partial_t V_j(x,t) + u(c) + \nabla_x V_j^T \mu_x + \frac{1}{2} \sigma_x^T H_x(V_j) \sigma_x.
\end{align*}
The implementation is outlined in Algorithm~\ref{algo:time-stepping} in Appendix~\ref{appendix:training-algorithms}.

\section{Experiments}\label{sec:experiments}
To evaluate Deep-MacroFin's performance, we undertake various tasks: basic ODEs/PDEs, free boundary models derived from variational problems, and two standard economic problems. We benchmark with DeepXDE or numerical solutions.
This section highlights main results, with detailed descriptions of each model and additional examples available in Appendix~\ref{appendix:models}. Low dimension models are trained on a Windows 11 machine with an i7-12700H CPU, RTX3070Ti Laptop GPU and 64GB RAM, while high dimension models are trained on a Linux RTX3090 machine. The backend uses PyTorch 2.4.1.

\subsection{Basic PDEs and benchmarking with DeepXDE}\label{sec:basic-problems}

We first evaluate basic boundary value problems with known analytic solutions to benchmark Deep-MacroFin against DeepXDE. Let $\Omega \subset \RR^n$ be the domain with boundary $\partial\Omega$, and let $L$ be a differential operator with boundary condition $\calC$. We seek the unique solution $u$ satisfying:
\begin{align*}
    \begin{cases}
        L[u](x) = 0, x\in\Omega\\
        \calC(u, x) = 0, x\in\partial\Omega
    \end{cases}
\end{align*}
In low dimensions, we benchmark using Cauchy-Euler, diffusion, and Black-Scholes equations. For higher dimensions, we use Laplace equations with harmonic boundary conditions, where the solution is analytically known as $u(x)=g(x), \forall x\in\Omega$ by the maximum principle. 
All low-dimensional models are trained with 50 random seeds using fixed configurations (Appendix~\ref{appendix:basic-models}). DeepXDE uses the same setup as the baseline MLPs in Deep-MacroFin. We performed grid search for MLP hyperparameters, but the MSE variance across configurations was negligible ($\sim 10^{-6}$). In contrast, KAN models were more sensitive to hyperparameters, and we manually selected the best-performing configuration.
Table~\ref{tab:basic-pde} reports mean and standard deviation of MSE, $\norm{u-\hat{u}}_{L^\infty}$, $\norm{u-\hat{u}}_{W^{1,\infty}}$, and $\norm{u-\hat{u}}_{W^{2,\infty}}$. The Sobolev $W^{1,\infty}$ norm uses the max over batches of summed vector differences, and, and $W^{2,\infty}$ does the same with matrix differences. For low dimensions, errors are computed on fixed grids; for 100D Laplacian, on $10^5$ random points.
Our MLP implementation matches DeepXDE in MSE and $\norm{u-\hat{u}}_{L^\infty}$ confirming robustness. However, DeepXDE wraps inputs as NumPy arrays for inference, preventing automatic differentiation for Jacobians and Hessians. Therefore, $W^{1,\infty}$ and $W^{2,\infty}$ norms are not reported for DeepXDE. These Sobolev norms are often large, as documented in prior work \cite{high-order-derivatives,high-order-derivatives2}, suggesting a future direction in designing architectures with better derivative accuracy.
To evaluate the symbolic interpretability of KANs \cite{kan}, we extract symbolic expressions and compare their errors to both the analytic solution (KAN Symbolic) and model predictions (KAN–KAN Symbolic). The symbolic errors align closely with original KAN errors, showing that KANs can accurately recover simple PDE solutions and potentially support efficient symbolic computation and root finding.

\begin{figure}[!htb]
\centering
\begin{subfigure}[b]{0.3\linewidth}
\centering
\includegraphics[width=\linewidth]{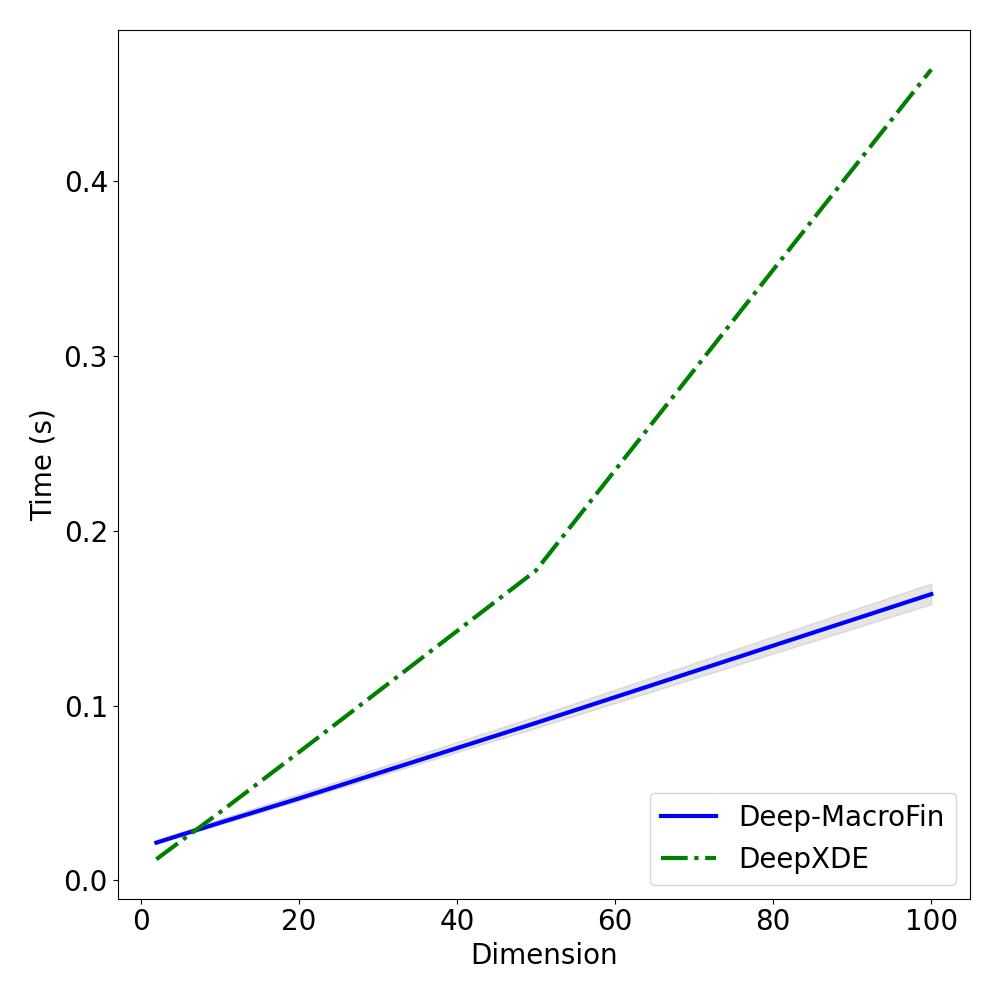}
\caption{Training Time per Epoch}
\end{subfigure}
\hfill
\begin{subfigure}[b]{0.3\linewidth}
\centering
\includegraphics[width=\linewidth]{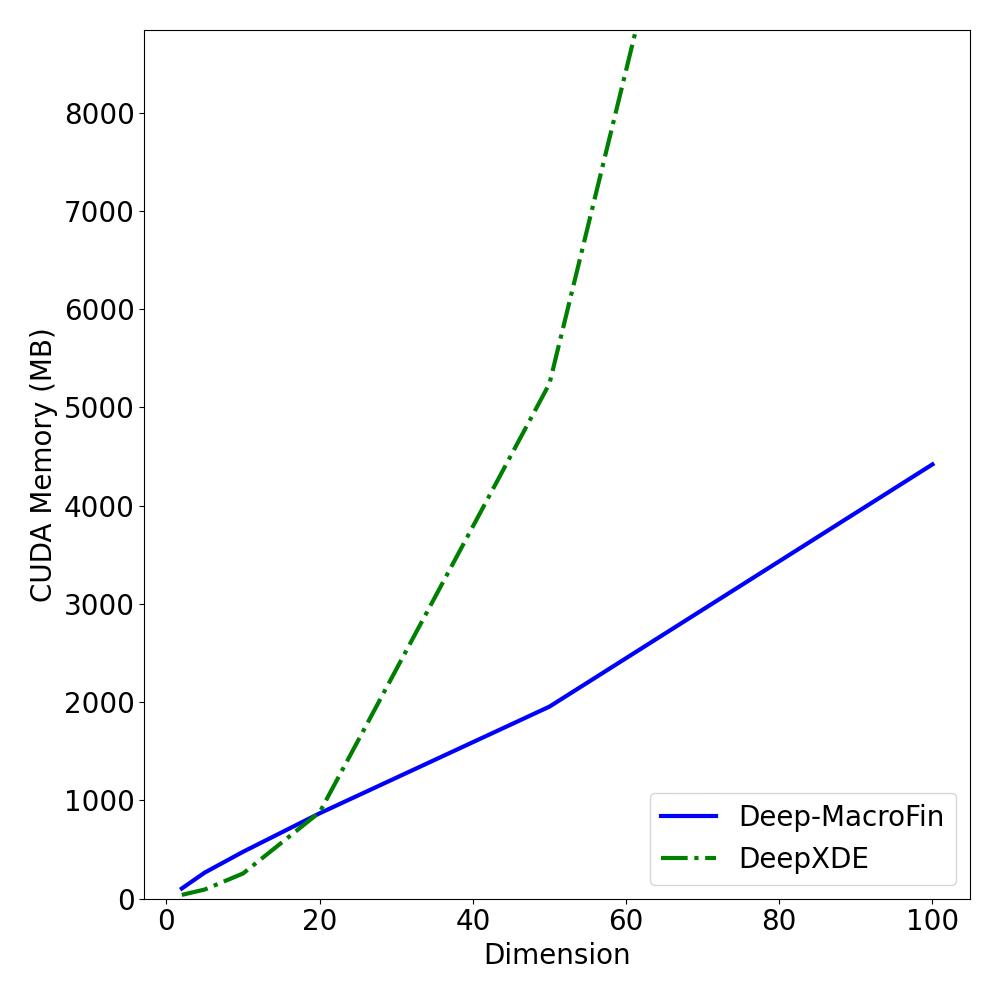}
\caption{Memory Usage}
\end{subfigure}
\hfill
\begin{subfigure}[b]{0.3\linewidth}
\centering
\includegraphics[width=\linewidth]{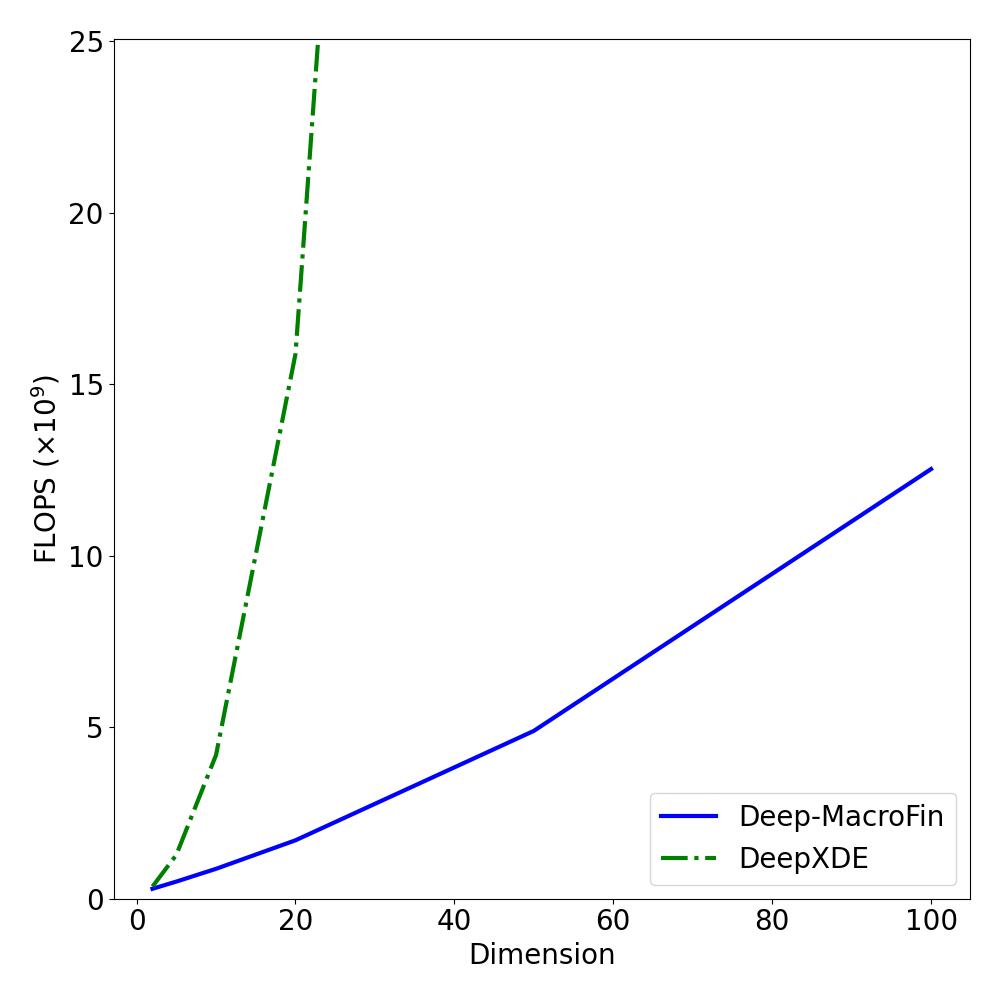}
\caption{FLOPs}
\end{subfigure}

\caption{Laplace equation benchmark \label{fig:laplace-equation}}
\end{figure}

To test scalability, we experiment with dimensions $n\in\bs{2,5,10,20,50,100}$ on a Linux machine with RTX 3090 GPU. Models are 4-layer MLPs with 30 neurons per layer and SiLU activations, using a batch size of 1000. For each boundary hyperplane, 100 points are sampled, totaling $1000+200n$ points per training step.
Figure~\ref{fig:laplace-equation} compares DeepXDE and Deep-MacroFin. Though Deep-MacroFin has slightly higher overhead in low dimensions (due to logging and derivative caching), it scales significantly better. It maintains stable runtime, memory usage, and FLOPs as dimensionality increases, requiring only 4.3GB CUDA memory ($\textbf{5}\times$ lower), 12 GFLOPs ($\textbf{40}\times$ lower), and 0.16s per epoch. Training for 100D model using Deep-MacroFin finishes within 30 minutes, while DeepXDE typically takes 75 minutes. DeepXDE’s rapidly increasing resource demands limit its scalability, which is a critical concern for high-dimensional economic applications like asset pricing models \cite{Martin2013}.

\begin{table}[!htb]
\caption{Basic PDE errors}\label{tab:basic-pde}

\centering
\resizebox*{\textwidth}{!}{%
\begin{tabular}{cccccc}
\toprule
PDE & Model & MSE & $\|u-\hat{u}\|_{L^\infty}$ & $\|u-\hat{u}\|_{W^{1,\infty}}$ & $\|u-\hat{u}\|_{W^{2,\infty}}$ \\
\midrule
Cauchy-Euler & DeepXDE & $1.56 \times 10^{-5}$ ($\pm$ $4.28 \times 10^{-5}$) & $3.14 \times 10^{-3}$ ($\pm$ $4.81 \times 10^{-3}$) &   &   \\
    & MLP & $5.82 \times 10^{-6}$ ($\pm$ $7.78 \times 10^{-6}$) & $2.60 \times 10^{-3}$ ($\pm$ $1.68 \times 10^{-3}$) & $7.72 \times 10^{-3}$ ($\pm$ $5.90 \times 10^{-3}$) & $1.83 \times 10^{-1}$ ($\pm$ $7.47 \times 10^{-2}$) \\
    & KAN & $2.79 \times 10^{-2}$ ($\pm$ $2.48 \times 10^{-2}$) & $4.29 \times 10^{-1}$ ($\pm$ $3.10 \times 10^{-1}$) & 6.02 ($\pm$ 4.16) & $5.43 \times 10^{1}$ ($\pm$ $3.17 \times 10^{1}$) \\
    & KAN Symbolic & $8.14 \times 10^{-2}$ ($\pm$ $1.91 \times 10^{-1}$) & $5.46 \times 10^{-1}$ ($\pm$ $3.86 \times 10^{-1}$) & 5.99 ($\pm$ 4.23) & $5.03 \times 10^{1}$ ($\pm$ $3.05 \times 10^{1}$) \\
    & KAN-KAN Symbolic & $5.13 \times 10^{-2}$ ($\pm$ $1.92 \times 10^{-1}$) & $2.65 \times 10^{-1}$ ($\pm$ $2.81 \times 10^{-1}$) & 1.20 ($\pm$ 1.32) & $1.04 \times 10^{1}$ ($\pm$ $2.40 \times 10^{1}$) \\
Diffusion & DeepXDE & $1.21 \times 10^{-5}$ ($\pm$ $1.41 \times 10^{-5}$) & $1.41 \times 10^{-2}$ ($\pm$ $4.25 \times 10^{-3}$) &   &   \\
    & MLP & $2.98 \times 10^{-5}$ ($\pm$ $2.87 \times 10^{-5}$) & $1.51 \times 10^{-2}$ ($\pm$ $4.90 \times 10^{-3}$) & $1.07 \times 10^{-1}$ ($\pm$ $3.28 \times 10^{-2}$) & 1.02 ($\pm$ $2.84 \times 10^{-1}$) \\
    & KAN & $9.26 \times 10^{-4}$ ($\pm$ $5.81 \times 10^{-6}$) & $9.80 \times 10^{-2}$ ($\pm$ $1.27 \times 10^{-3}$) & $7.19 \times 10^{-1}$ ($\pm$ $3.91 \times 10^{-2}$) & 3.12 ($\pm$ $4.47 \times 10^{-1}$) \\
    & KAN Symbolic & $9.26 \times 10^{-4}$ ($\pm$ $6.40 \times 10^{-6}$) & $9.81 \times 10^{-2}$ ($\pm$ $1.33 \times 10^{-3}$) & $7.19 \times 10^{-1}$ ($\pm$ $3.90 \times 10^{-2}$) & 3.12 ($\pm$ $4.47 \times 10^{-1}$) \\
    & KAN-KAN Symbolic & $4.06 \times 10^{-9}$ ($\pm$ $1.91 \times 10^{-8}$) & $1.02 \times 10^{-4}$ ($\pm$ $6.01 \times 10^{-5}$) & $2.95 \times 10^{-4}$ ($\pm$ $3.26 \times 10^{-5}$) & $1.04 \times 10^{-3}$ ($\pm$ $4.40 \times 10^{-5}$) \\
Black-Scholes & DeepXDE & $2.40 \times 10^{-4}$ ($\pm$ $1.05 \times 10^{-3}$) & $3.79 \times 10^{-2}$ ($\pm$ $4.42 \times 10^{-2}$) &   &   \\
    & MLP & $1.20 \times 10^{-5}$ ($\pm$ $5.87 \times 10^{-6}$) & $1.53 \times 10^{-2}$ ($\pm$ $3.60 \times 10^{-3}$) & $8.38 \times 10^{-1}$ ($\pm$ $1.11 \times 10^{-2}$) & $1.10 \times 10^{2}$ ($\pm$ $7.15 \times 10^{-1}$) \\
    & KAN & $5.87 \times 10^{-5}$ ($\pm$ $2.41 \times 10^{-5}$) & $2.43 \times 10^{-2}$ ($\pm$ $1.98 \times 10^{-3}$) & $8.42 \times 10^{-1}$ ($\pm$ $1.18 \times 10^{-2}$) & $1.14 \times 10^{2}$ ($\pm$ $2.25 \times 10^{-1}$) \\
    & KAN Symbolic & $5.88 \times 10^{-5}$ ($\pm$ $2.41 \times 10^{-5}$) & $2.43 \times 10^{-2}$ ($\pm$ $2.01 \times 10^{-3}$) & $8.42 \times 10^{-1}$ ($\pm$ $1.17 \times 10^{-2}$) & $1.14 \times 10^{2}$ ($\pm$ $2.25 \times 10^{-1}$) \\
    & KAN-KAN Symbolic & $6.81 \times 10^{-9}$ ($\pm$ $4.95 \times 10^{-9}$) & $1.21 \times 10^{-4}$ ($\pm$ $6.10 \times 10^{-5}$) & $3.27 \times 10^{-4}$ ($\pm$ $1.75 \times 10^{-4}$) & $2.27 \times 10^{-3}$ ($\pm$ $1.26 \times 10^{-3}$) \\
Laplacian (100D) & DeepXDE & $2.26 \times 10^{-9}$ ($\pm$ $1.33 \times 10^{-10}$) & $2.52 \times 10^{-4}$ ($\pm$ $4.37 \times 10^{-5}$) &    &    \\
(Zero boundary)  & MLP & $9.14 \times 10^{-10}$ ($\pm$ $4.99 \times 10^{-11}$) & $1.36 \times 10^{-4}$ ($\pm$ $2.31 \times 10^{-5}$) & $1.36 \times 10^{-4}$ ($\pm$ $2.31 \times 10^{-5}$) & $1.36 \times 10^{-4}$ ($\pm$ $2.31 \times 10^{-5}$) \\
Laplacian (100D) & DeepXDE & $2.56 \times 10^{-2}$ ($\pm$ $2.10 \times 10^{-4}$) & $2.25 \times 10^{-1}$ ($\pm$ $1.01 \times 10^{-2}$) &    &    \\
(Summation boundary)  & MLP & $1.60 \times 10^{-3}$ ($\pm$ $8.36 \times 10^{-5}$) & $1.81 \times 10^{-1}$ ($\pm$ $3.27 \times 10^{-2}$) & $1.81 \times 10^{-1}$ ($\pm$ $3.27 \times 10^{-2}$) & $1.81 \times 10^{-1}$ ($\pm$ $3.27 \times 10^{-2}$) \\
\bottomrule
\end{tabular}}
\end{table}

\subsection{Free boundary problems}\label{sec:free-boundary}

Many real-world problems in engineering, economics, and finance involve constraints and exhibit free boundaries. For example, in American option pricing, early exercise implies that the Black-Scholes equation holds only when the option price exceeds the payoff. Similar free-boundary structures appear in principal-agent models \cite{rochet-chone} and incomplete market equilibria \cite{Brunnermeier2014}. Formally, the goal is to find the unique minimizer $u$ of a functional $E[u]$ within a constrained subspace $V$ of neural network-approximated functions:
\begin{align*}
    \min_{u\in V} E[u] \quad\text{or equivalently,} \quad \min \bs{L[u](x), \calC(u, x)} = 0,
\end{align*}
where $L[u]$ is the PDE derived from the variational problem and $\calC(u,x)$ encodes the free boundary condition. Table~\ref{tab:free-boundary-models} reports results for the Rochet–Choné model in 1D with active ($a=0.5$) and inactive ($a=2$) free boundaries; and the American option with dividend yield. Errors (MSE, $L^\infty$, and free boundary violation) are measured against numerical solutions from a projected SOR method on a fixed grid. Means and standard deviations are computed over 50 random seeds.

\begin{figure}[!htb]
\centering

\begin{subfigure}[b]{0.24\linewidth}
\centering
\includegraphics[width=\linewidth]{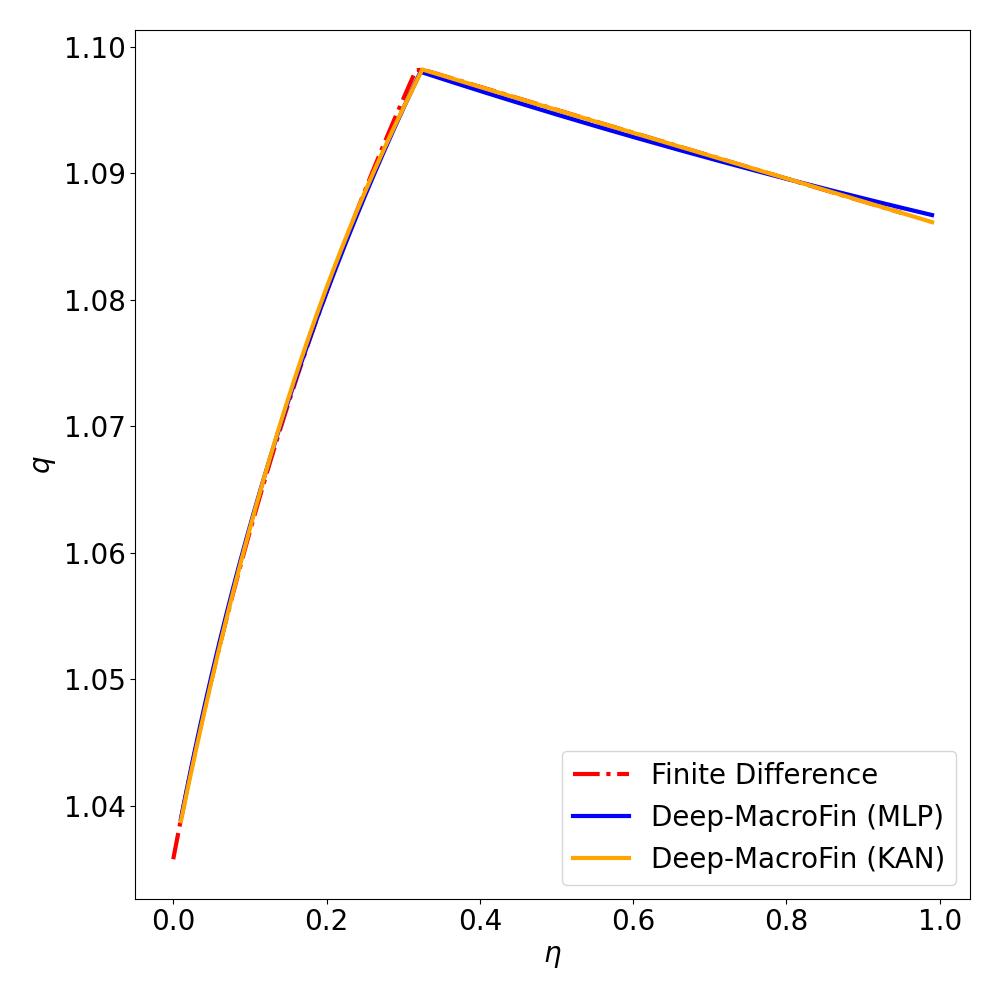}
\caption{Price}
\end{subfigure}
\begin{subfigure}[b]{0.24\linewidth}
\centering
\includegraphics[width=\linewidth]{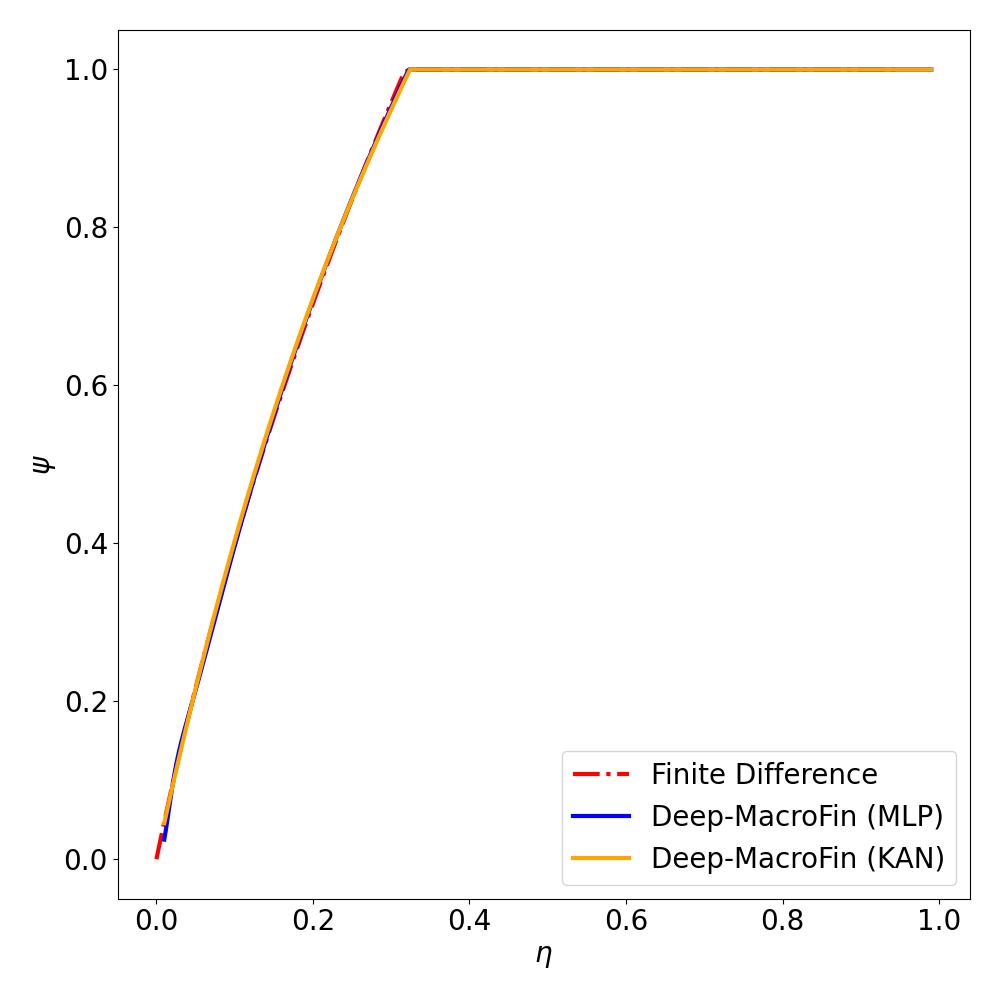}
\caption{Experts Capital Share}
\end{subfigure}

\caption{Brunnermeier \& Sannikov: Red shows the finite difference solutions by PyMacroFin; blue shows the fitted solution by Deep-MacroFin with MLP; orange shows the fitted solution by KAN.}
\label{fig:free-boundary-model}
\end{figure}

Our main focus is the incomplete market equilibrium (Proposition 4 in \cite{Brunnermeier2014}), where the goal is to identify a critical wealth share $\eta^{\psi}$, beyond which the expert controls all market capital. When $\eta \in (0, \eta^{\psi})$, agents can short-sell, leading to a nonlinear rise in the capital price $q$. When $\eta > \eta^{\psi}$, short-selling is prohibited, and $q$ decreases linearly. 
Unlike other free-boundary problems (e.g., principal-agent and Black-Scholes), where transitions are smooth, the incomplete market model exhibits a discontinuous first derivative. To address this, we adopt an upwinding scheme: we solve two systems, one assuming $\psi<1$, the other assuming $\psi=1$, then locate the boundary where $\psi=1$ in the first system and merge it with the second.
Table~\ref{tab:free-boundary-1d} and Figure~\ref{fig:free-boundary-model} show that both MLPs and KANs accurately approximate $q$ and $\psi$. A key advantage of KAN is its ability to yield an analytical expression for $\eta^\psi$. For example: $\psi=-2.2823 + 3.7254 \exp\left(-1.0082 (0.6741-\eta)^2\right)$ implies $\eta^\psi\approx 0.3197$ for $\psi(\eta^\psi)=1$. However, scaling KANs to higher dimensions remains challenging due to numerical instability with deeper or wider networks and limited parallelism. Also, as KAN becomes deeper or wider, the symbolic formula becomes more difficult to simplify and interpret.

\begin{table}[!htb]
\caption{Free boundary models errors}\label{tab:free-boundary-models}

\centering
\resizebox*{\textwidth}{!}{%
\begin{tabular}{ccccc}
\toprule
PDE & Model & MSE & $\|u-\hat{u}\|_{L^\infty}$ & Free Boundary Violation\\
\midrule
Principal agent ($a=0.5$) & MLP & $3.28 \times 10^{-4}$ ($\pm$ $2.04 \times 10^{-4}$) & $2.66 \times 10^{-2}$ ($\pm$ $1.09 \times 10^{-2}$) & 0.00 ($\pm$ 0.00) \\
    & KAN & $2.23 \times 10^{-4}$ ($\pm$ $2.14 \times 10^{-5}$) & $2.07 \times 10^{-2}$ ($\pm$ $1.48 \times 10^{-3}$) & $1.39 \times 10^{-5}$ ($\pm$ $9.92 \times 10^{-7}$) \\
    & KAN Symbolic & $2.30 \times 10^{-4}$ ($\pm$ $2.10 \times 10^{-5}$) & $2.16 \times 10^{-2}$ ($\pm$ $1.57 \times 10^{-3}$) & $9.27 \times 10^{-4}$ ($\pm$ $2.99 \times 10^{-4}$) \\
    & KAN-KAN Symbolic & $2.35 \times 10^{-6}$ ($\pm$ $1.38 \times 10^{-6}$) & $5.79 \times 10^{-3}$ ($\pm$ $1.35 \times 10^{-3}$) & $9.27 \times 10^{-4}$ ($\pm$ $2.99 \times 10^{-4}$) \\
Principal agent ($a=2$) & MLP & $4.76 \times 10^{-3}$ ($\pm$ $4.42 \times 10^{-3}$) & $7.75 \times 10^{-2}$ ($\pm$ $2.77 \times 10^{-2}$) & 0.00 ($\pm$ 0.00) \\
    & KAN & $4.53 \times 10^{-5}$ ($\pm$ $3.64 \times 10^{-5}$) & $1.08 \times 10^{-2}$ ($\pm$ $2.59 \times 10^{-3}$) & $9.53 \times 10^{-3}$ ($\pm$ $1.49 \times 10^{-3}$) \\
    & KAN Symbolic & $1.41 \times 10^{-3}$ ($\pm$ $4.28 \times 10^{-3}$) & $2.23 \times 10^{-2}$ ($\pm$ $3.74 \times 10^{-2}$) & $5.46 \times 10^{-3}$ ($\pm$ $2.23 \times 10^{-3}$) \\
    & KAN-KAN Symbolic & $1.53 \times 10^{-3}$ ($\pm$ $4.84 \times 10^{-3}$) & $1.71 \times 10^{-2}$ ($\pm$ $4.22 \times 10^{-2}$) & $5.46 \times 10^{-3}$ ($\pm$ $2.23 \times 10^{-3}$) \\
Black-Scholes (American Option) & MLP & $6.49 \times 10^{-6}$ ($\pm$ $5.58 \times 10^{-6}$) & $5.47 \times 10^{-3}$ ($\pm$ $2.76 \times 10^{-3}$) & $2.92 \times 10^{-3}$ ($\pm$ $1.68 \times 10^{-3}$) \\
    & KAN & $1.48 \times 10^{-5}$ ($\pm$ $4.05 \times 10^{-6}$) & $9.83 \times 10^{-3}$ ($\pm$ $5.72 \times 10^{-4}$) & $9.72 \times 10^{-3}$ ($\pm$ $5.54 \times 10^{-4}$) \\
    & KAN Symbolic & $5.49 \times 10^{-5}$ ($\pm$ $3.32 \times 10^{-6}$) & $1.38 \times 10^{-2}$ ($\pm$ $3.90 \times 10^{-4}$) & $1.38 \times 10^{-2}$ ($\pm$ $3.90 \times 10^{-4}$) \\
    & KAN-KAN Symbolic & $5.38 \times 10^{-5}$ ($\pm$ $5.84 \times 10^{-6}$) & $1.81 \times 10^{-2}$ ($\pm$ $1.01 \times 10^{-3}$) & $1.38 \times 10^{-2}$ ($\pm$ $3.90 \times 10^{-4}$) \\
\bottomrule
\end{tabular}}
\end{table}

\begin{table}[!htb]
\caption{Brunnermeier \& Sannikov}\label{tab:free-boundary-1d}

\centering
\resizebox*{\textwidth}{!}{%
\begin{tabular}{ccccc}
\toprule
Model & MSE($q$, $\hat{q}$) & $\|q-\hat{q}\|_{L^\infty}$ & MSE($\psi$, $\hat{\psi}$) & $\|\psi-\hat{\psi}\|_{L^\infty}$\\
\midrule
MLP & $1.36 \times 10^{-6}$ ($\pm$ $7.03 \times 10^{-7}$) & $6.38 \times 10^{-5}$ ($\pm$ $4.51 \times 10^{-5}$) & $2.75 \times 10^{-3}$ ($\pm$ $8.91 \times 10^{-4}$) & $2.97 \times 10^{-2}$ ($\pm$ $5.87 \times 10^{-3}$) \\
KAN & $9.58 \times 10^{-8}$ ($\pm$ $2.74 \times 10^{-7}$) & $1.54 \times 10^{-5}$ ($\pm$ $4.20 \times 10^{-5}$) & $1.08 \times 10^{-3}$ ($\pm$ $3.94 \times 10^{-4}$) & $1.35 \times 10^{-2}$ ($\pm$ $4.87 \times 10^{-3}$) \\
KAN Symbolic & $7.84 \times 10^{-7}$ ($\pm$ $1.05 \times 10^{-7}$) & $1.78 \times 10^{-4}$ ($\pm$ $1.58 \times 10^{-5}$) & $4.50 \times 10^{-3}$ ($\pm$ $2.74 \times 10^{-4}$) & $7.25 \times 10^{-2}$ ($\pm$ $3.98 \times 10^{-3}$) \\
KAN-KAN Symbolic & $7.98 \times 10^{-7}$ ($\pm$ $1.07 \times 10^{-7}$) & $2.08 \times 10^{-4}$ ($\pm$ $2.89 \times 10^{-5}$) & $4.45 \times 10^{-3}$ ($\pm$ $3.98 \times 10^{-4}$) & $7.19 \times 10^{-2}$ ($\pm$ $8.18 \times 10^{-3}$) \\
\bottomrule
\end{tabular}}
\end{table}

\subsection{Neoclassical growth model}\label{sec:ncg}
\begin{figure}[!htb]
\centering
\begin{subfigure}[b]{0.24\linewidth}
\centering
\includegraphics[width=\linewidth]{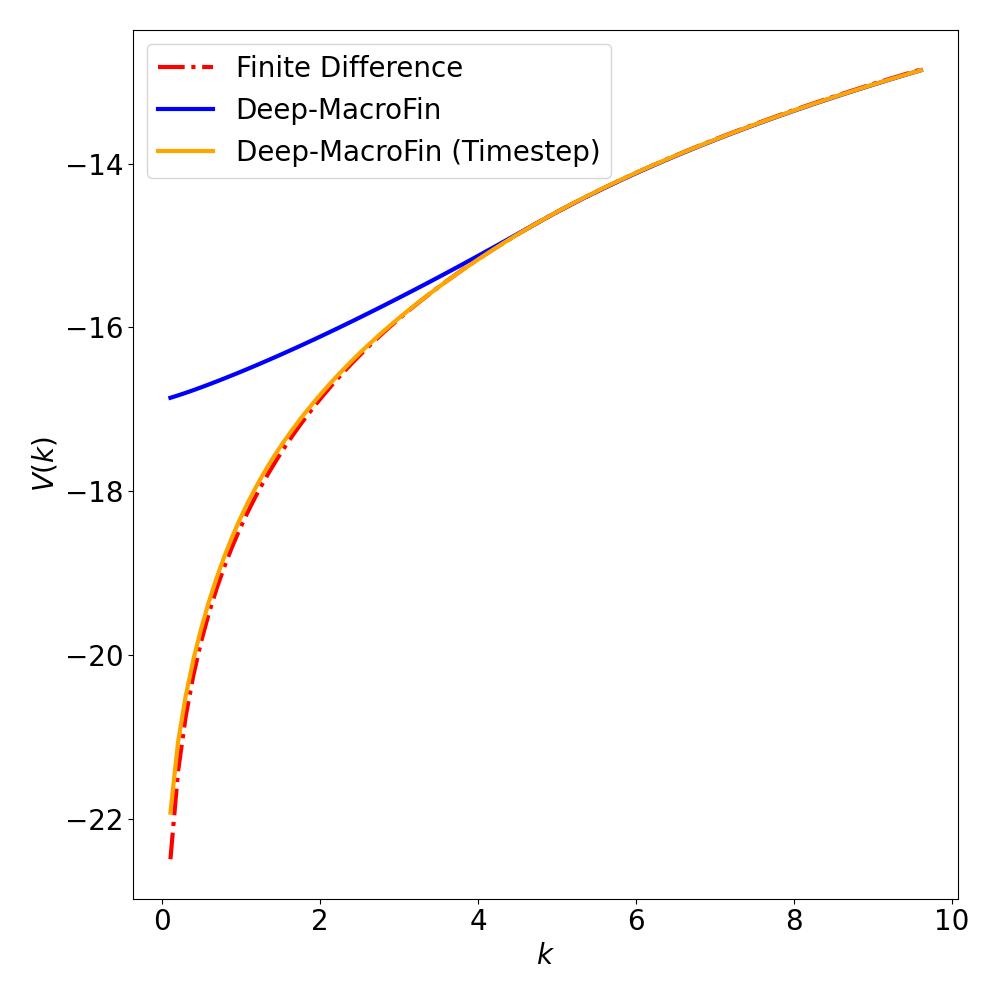}
\caption{CRRA ($\gamma=2$) $V$}
\end{subfigure}
\hfill
\begin{subfigure}[b]{0.24\linewidth}
\centering
\includegraphics[width=\linewidth]{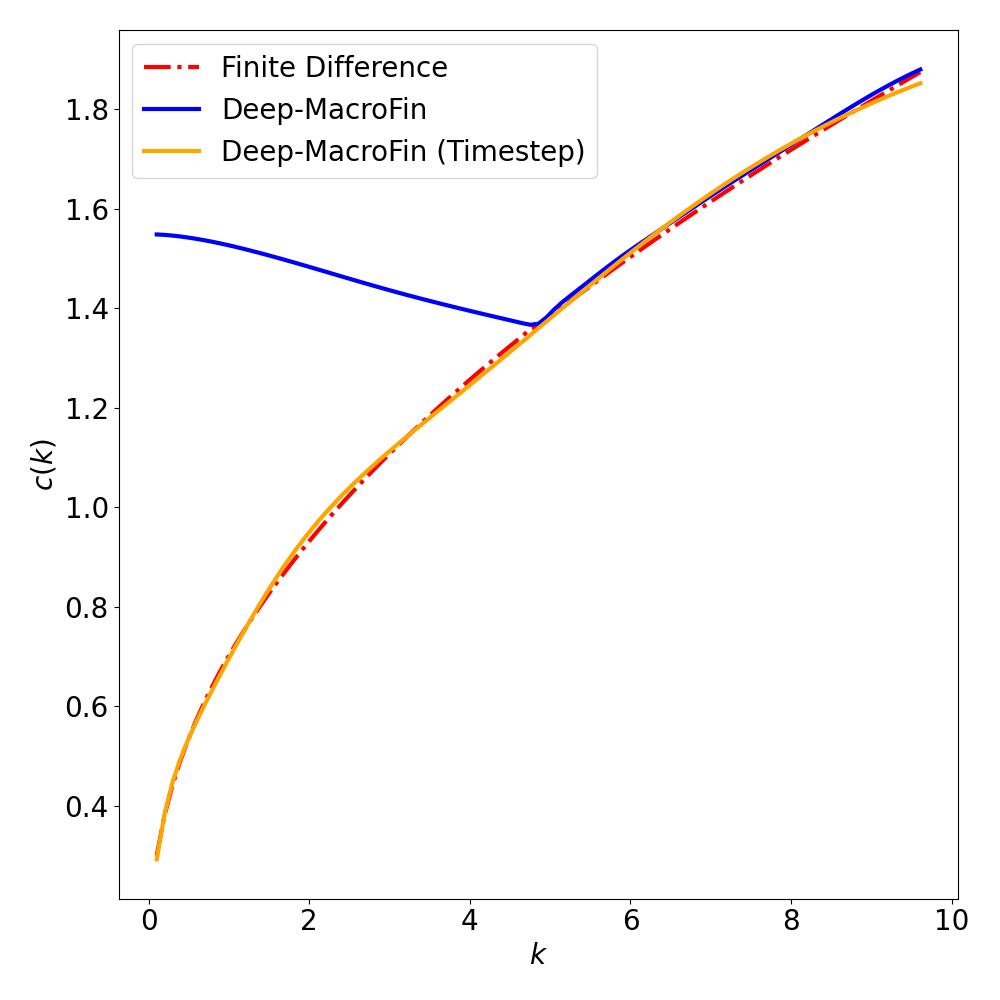}
\caption{CRRA ($\gamma=2$) $c$}
\end{subfigure}
\hfill
\begin{subfigure}[b]{0.24\linewidth}
\centering
\includegraphics[width=\linewidth]{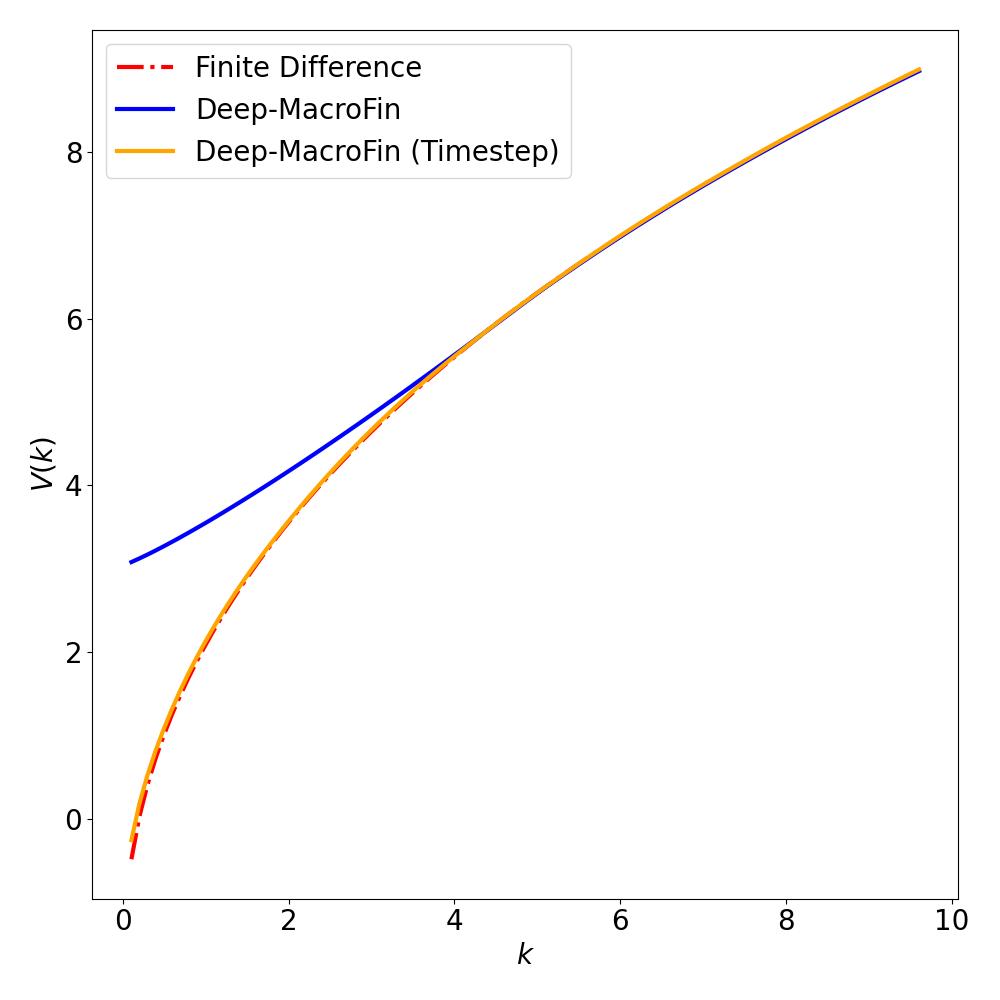}
\caption{Log ($\gamma=1$) $V$}
\end{subfigure}
\hfill
\begin{subfigure}[b]{0.24\linewidth}
\centering
\includegraphics[width=\linewidth]{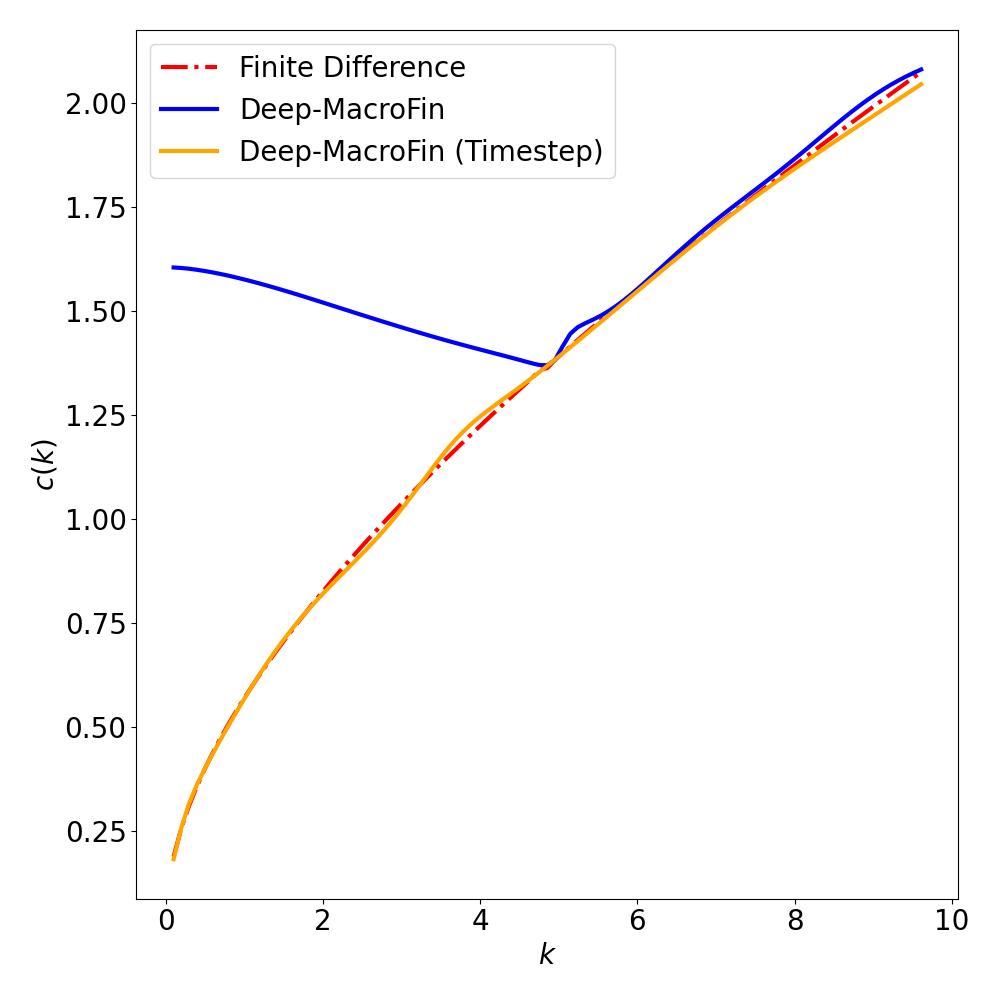}
\caption{Log ($\gamma=1$) $c$}
\end{subfigure}

\caption{Neoclassical growth model global solution}
\label{fig:ncg-global}
\end{figure}

The neoclassical growth model is a standard dynamic programming problem in macroeconomics. Its HJB equation is:
\begin{align*}
    \rho V(k) &= \max_c u(c) + V'(k) (k^\alpha - \delta k -c),
\end{align*}
where $u(c)=\begin{cases*}
    \frac{c^{1-\gamma}}{1-\gamma}, \gamma\neq 1 (\text{CRRA})\\
    \log(c), \gamma=1 (\text{Log})
\end{cases*}$ is the utility function. The FOC implies $u'(c)=V'(k)$. We parametrize both the value function $V$ and consumption $c$ using neural networks, and solve the system over the steady state $[k_{ss},2k_{ss}]$. Table~\ref{tab:neoclassical-growth} reports errors across 20 random seeds for both MLP and KAN models. The neural network models approximate $V$ and $c$ well, with errors around $10^{-3}$ for both CRRA and log utility. However, KAN's symbolic approximation often fails due to numerical instability, resulting in significantly higher errors in many trials. When extending the domain to $[0,2k_{ss}]$, standard training struggles to approximate $V$ and $c$ accurately outside the steady state.  To improve stability, we adopt the time-stepping scheme, by modifying the HJB equation to:
\begin{align*}
    \rho V(k, t) &= \max_c \partial_t V(k, t) + u(c) + \partial_k V (k, t) (k^\alpha - \delta k -c),
\end{align*}
This modification reduces the mean squared error (MSE) across the entire domain to $10^{-5}$, significantly improving accuracy, particularly outside the steady state, as shown in Figure~\ref{fig:ncg-global}. Furthermore, we extend this framework to a high-dimensional setting, with results presented in Appendix~\ref{appendix:ncg}.

\begin{table}[!htb]
\caption{Neoclassical growth model steady state errors}\label{tab:neoclassical-growth}

\centering
\resizebox*{\textwidth}{!}{\begin{tabular}{cccccc}
\toprule
PDE & Model & MSE ($V$) & $\|V-\hat{V}\|_{L^\infty}$ & MSE ($c$) & $\|c-\hat{c}\|_{L^\infty}$ \\
\midrule
CRRA ($\gamma=2$) & MLP & $9.54 \times 10^{-5}$ ($\pm$ $1.33 \times 10^{-4}$) & $1.17 \times 10^{-2}$ ($\pm$ $7.25 \times 10^{-3}$) & $2.03 \times 10^{-4}$ ($\pm$ $1.93 \times 10^{-4}$) & $2.83 \times 10^{-2}$ ($\pm$ $1.39 \times 10^{-2}$) \\
    & KAN & $5.89 \times 10^{-5}$ ($\pm$ $7.84 \times 10^{-5}$) & $9.60 \times 10^{-3}$ ($\pm$ $5.64 \times 10^{-3}$) & $7.18 \times 10^{-5}$ ($\pm$ $8.55 \times 10^{-5}$) & $1.71 \times 10^{-2}$ ($\pm$ $1.27 \times 10^{-2}$) \\
    & KAN Symbolic & 2.97 ($\pm$ 4.47) & 1.78 ($\pm$ 1.44) & $9.12 \times 10^{-1}$ ($\pm$ 3.81) & $3.71 \times 10^{-1}$ ($\pm$ $9.20 \times 10^{-1}$) \\
    & KAN-KAN Symbolic & 2.97 ($\pm$ 4.46) & 1.78 ($\pm$ 1.43) & $9.11 \times 10^{-1}$ ($\pm$ 3.81) & $3.61 \times 10^{-1}$ ($\pm$ $9.17 \times 10^{-1}$) \\
Log ($\gamma=1$) & MLP & $4.74 \times 10^{-5}$ ($\pm$ $3.08 \times 10^{-5}$) & $9.17 \times 10^{-3}$ ($\pm$ $2.51 \times 10^{-3}$) & $4.54 \times 10^{-5}$ ($\pm$ $1.94 \times 10^{-5}$) & $1.41 \times 10^{-2}$ ($\pm$ $2.47 \times 10^{-3}$) \\
    & KAN & $6.82 \times 10^{-5}$ ($\pm$ $8.24 \times 10^{-5}$) & $1.03 \times 10^{-2}$ ($\pm$ $5.05 \times 10^{-3}$) & $1.80 \times 10^{-4}$ ($\pm$ $1.87 \times 10^{-4}$) & $2.12 \times 10^{-2}$ ($\pm$ $1.01 \times 10^{-2}$) \\
    & KAN Symbolic & 1.64 ($\pm$ 5.05) & $5.57 \times 10^{-1}$ ($\pm$ 1.36) & $1.64 \times 10^{6}$ ($\pm$ $8.44 \times 10^{5}$) & $1.19 \times 10^{3}$ ($\pm$ $5.44 \times 10^{2}$) \\
    & KAN-KAN Symbolic & 1.63 ($\pm$ 5.05) & $5.50 \times 10^{-1}$ ($\pm$ 1.36) & $1.64 \times 10^{6}$ ($\pm$ $8.44 \times 10^{5}$) & $1.19 \times 10^{3}$ ($\pm$ $5.44 \times 10^{2}$) \\
\bottomrule
\end{tabular}}
\end{table}

\subsection{Lucas orchard}\label{sec:lucas-orchard}
\begin{figure}[!htb]
\centering
\begin{subfigure}[b]{0.24\linewidth}
\centering
\includegraphics[width=\linewidth]{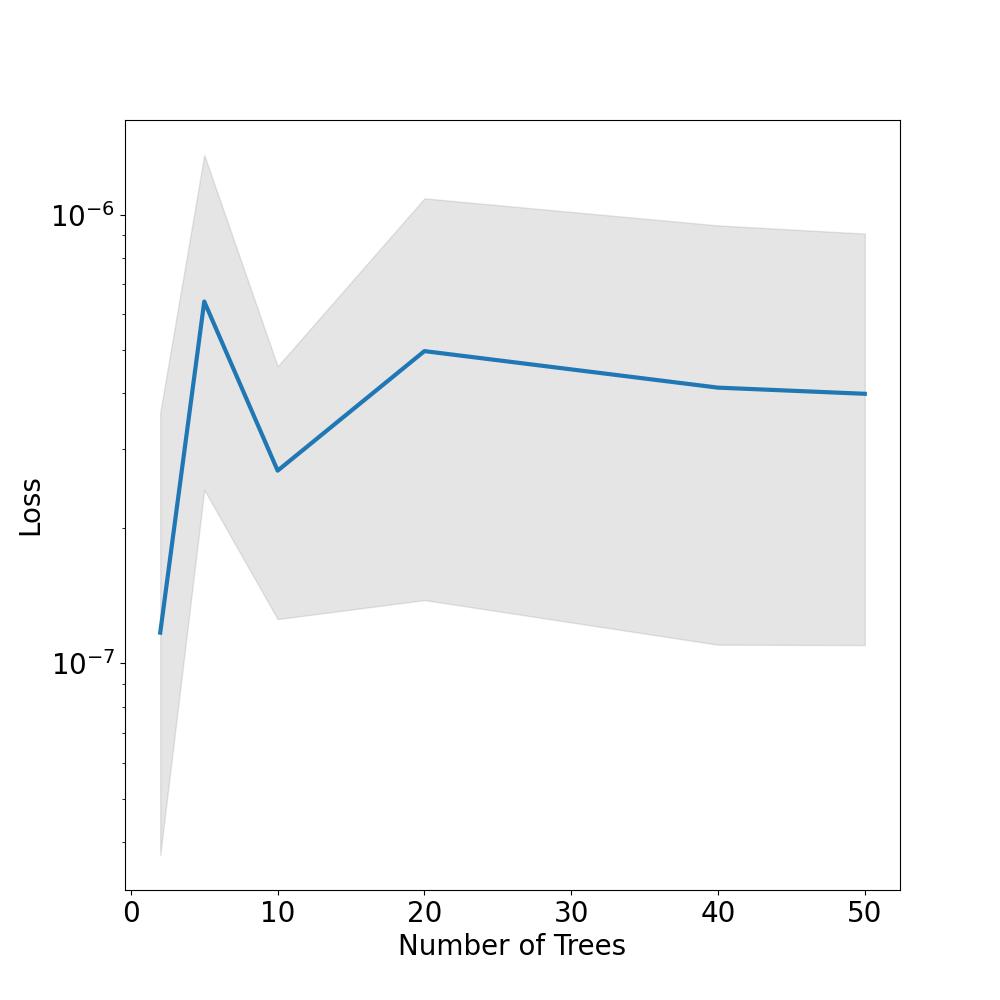}
\caption{Total Loss}
\end{subfigure}
\hfill
\begin{subfigure}[b]{0.24\linewidth}
\centering
\includegraphics[width=\linewidth]{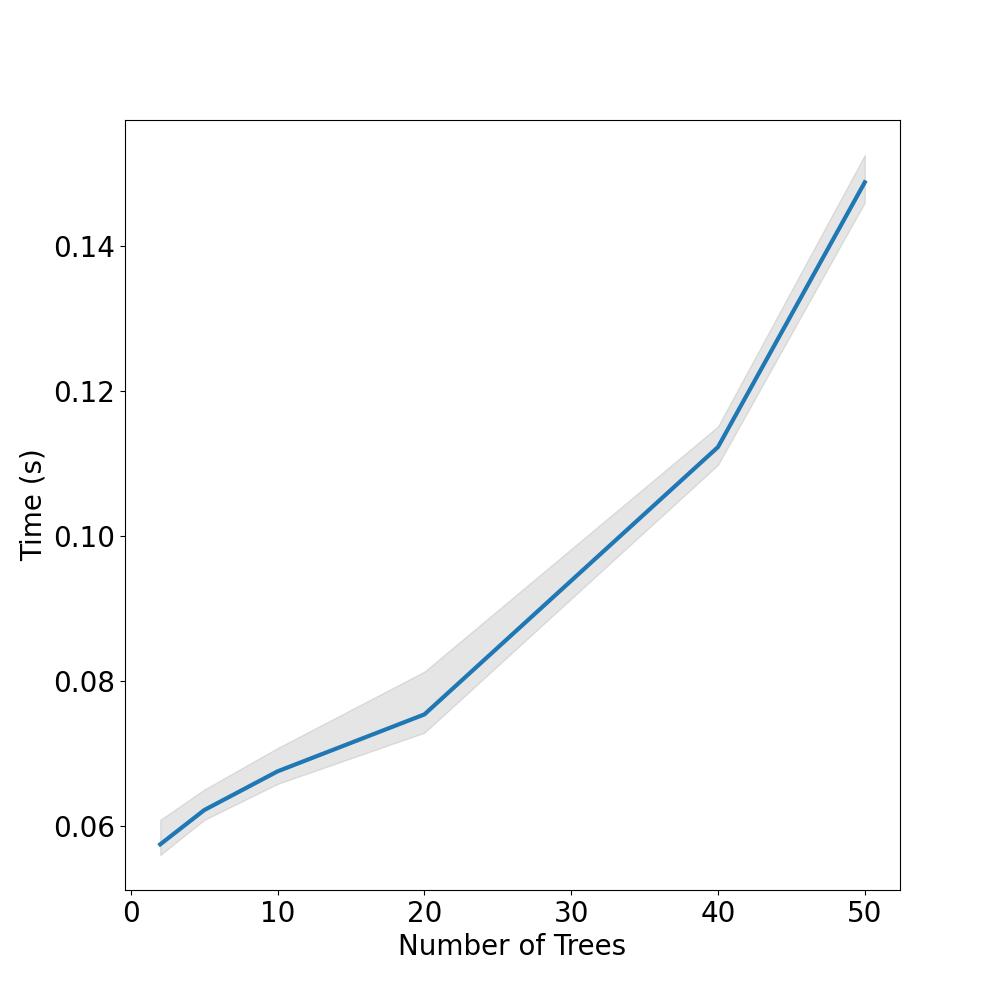}
\caption{Training Time}
\end{subfigure}
\hfill
\begin{subfigure}[b]{0.24\linewidth}
\centering
\includegraphics[width=\linewidth]{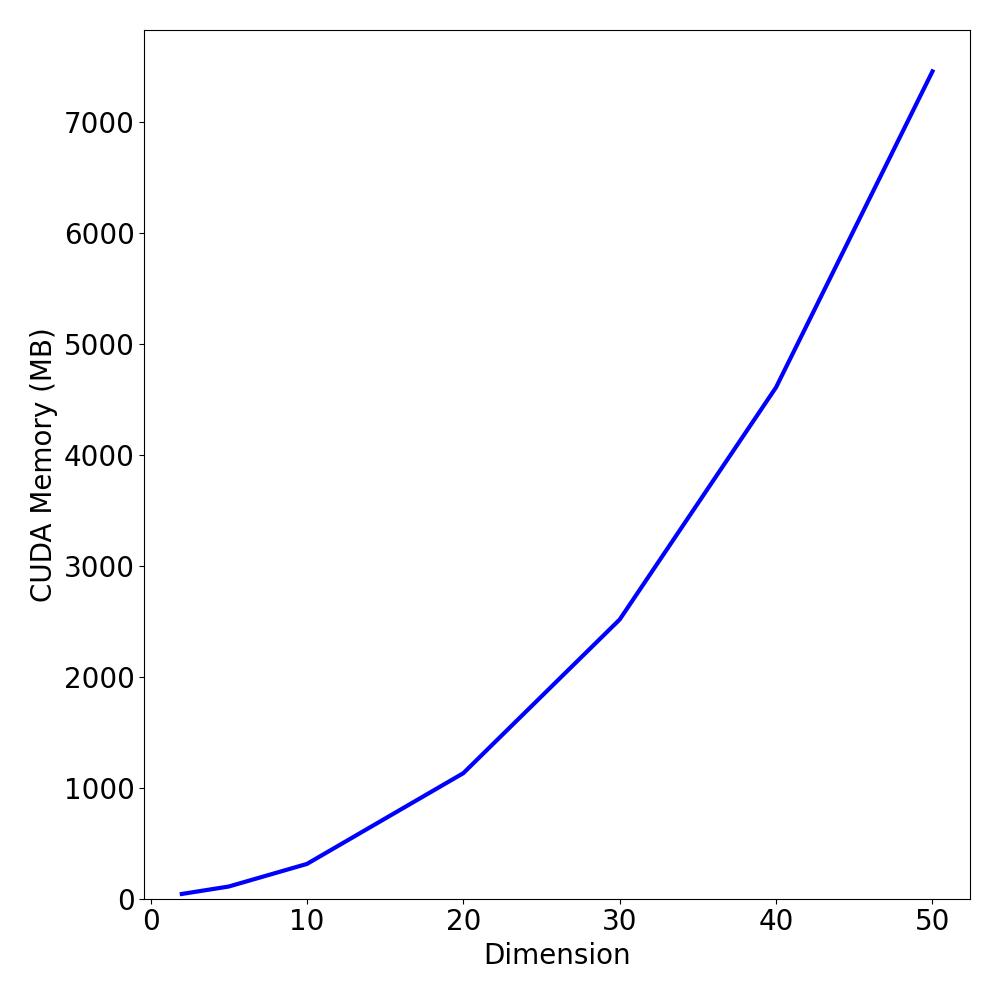}
\caption{Memory Usage}
\end{subfigure}
\hfill
\begin{subfigure}[b]{0.24\linewidth}
\centering
\includegraphics[width=\linewidth]{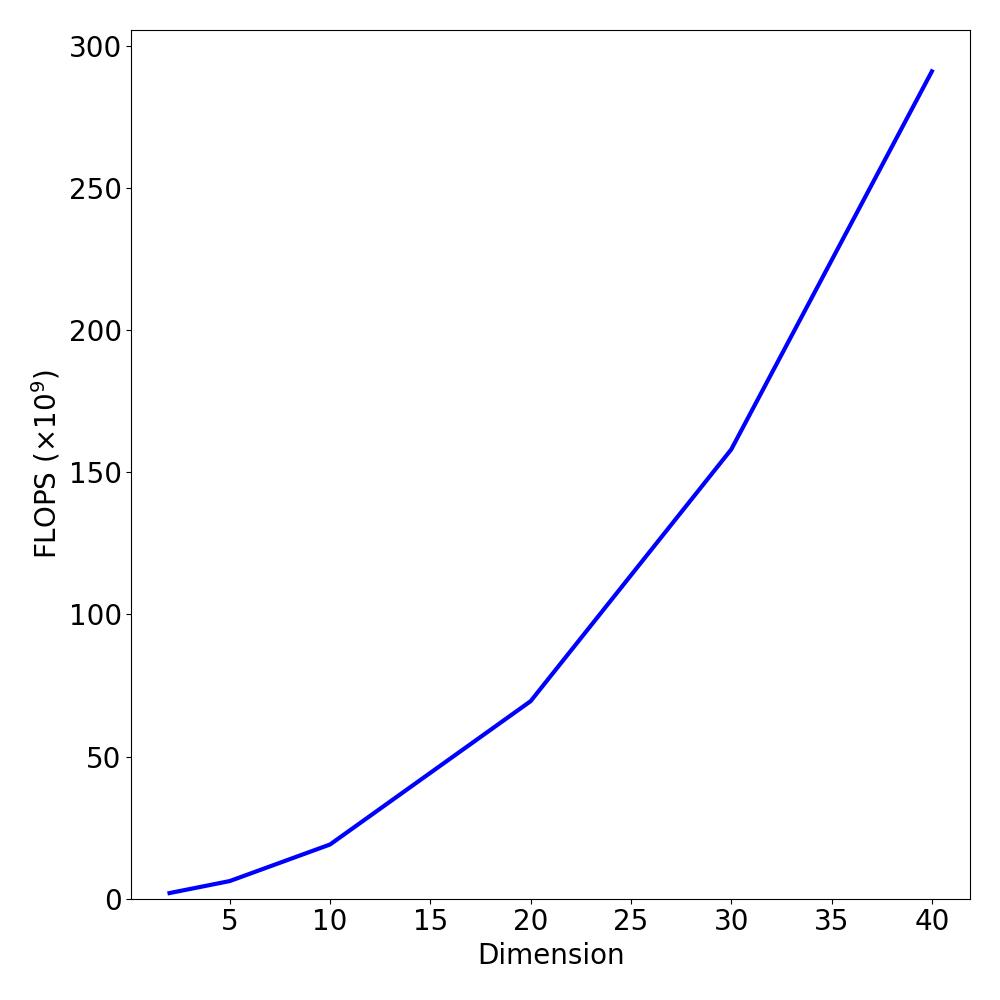}
\caption{FLOPs}
\end{subfigure}

\caption{Lucas orchard benchmark}
\label{fig:tree-benchmark}
\end{figure}

In the final example, we study the Lucas orchard asset pricing problem \cite{Martin2013}. For an economy with $N$ trees, the state variables are the wealth shares $z$ of each tree. We parametrize the value function $\kappa$ with neural networks and solve the HJB system using time-stepping for $N\in\bs{2,5,10,20,30,40,50}$:
\begin{align*}
    \mu^{\kappa}\kappa=&\partial_t \kappa + \nabla_{z} \kappa \mu_z + \frac{1}{2} \sigma_z^T H_z (\kappa) \sigma_z\\
    \sigma^{\kappa}\kappa=&\nabla_{z} \kappa  \sigma_z,
\end{align*}
where $\mu^\kappa$ and $\sigma^\kappa$ also depend on price $q$ of each tree, and $\mu_z$, $\sigma_z$ are geometric drift and diffusion of $z$. Figure~\ref{fig:tree-benchmark} shows the results, with shaded regions indicating the $[5\%,95\%]$ interval. FLOP tracking ends at $N=40$ due to PyTorch profiler overhead, but the model scales efficiently, using only 8GB VRAM for $N=50$, and can extend to higher dimensions. Validation against a numerical solution in 1D is provided in Appendix~\ref{appendix:trees}. With time-stepping and ergodic sampling, the 50-tree model converges within 10 minutes.

\section{Conclusion}
In this paper, we introduced Deep-MacroFin, a flexible and scalable framework for solving continuous-time economic models using deep learning. 
Designed to handle systems of differential equations, it outperforms existing libraries in both high-dimensional scalability and user customizability, offering a range of training algorithms and sampling strategies. We demonstrate that incorporating a time-stepping scheme from traditional numerical methods improves neural network solutions for continuous-time economic models with HJB equations.
We also evaluated the performance of KANs. While their symbolic representations provide accurate approximations for simple PDEs and support analytical insights, they face challenges in scaling to higher dimensions due to numerical instability and limited parallelism.
Our current limitations include the overhead from derivative pre-caching, leading to slight slowdowns in low dimensions, and the need for manual tuning of loss weights for improved convergence. 
Future work will focus on expanding the framework to more complex, high-dimensional, and dynamic economic models, as well as incorporating active learning and automated loss balancing to enhance training efficiency and accuracy.




\bibliography{main}

\begin{thebibliography}{47}
\providecommand{\natexlab}[1]{#1}
\providecommand{\url}[1]{\texttt{#1}}
\expandafter\ifx\csname urlstyle\endcsname\relax
  \providecommand{\doi}[1]{doi: #1}\else
  \providecommand{\doi}{doi: \begingroup \urlstyle{rm}\Url}\fi

\bibitem[Achdou et~al.(2017)Achdou, Han, Lasry, Lions, and Moll]{yves2017}
Y.~Achdou, J.~Han, J.-M. Lasry, P.-L. Lions, and B.~Moll.
\newblock Income and wealth distribution in macroeconomics: A continuous-time
  approach.
\newblock Working Paper 23732, National Bureau of Economic Research, 8 2017.
\newblock URL \url{http://www.nber.org/papers/w23732}.

\bibitem[Achdou et~al.(2022)Achdou, Han, Lasry, Lions, and Moll]{moll2020}
Y.~Achdou, J.~Han, J.-M. Lasry, P.~L. Lions, and B.~Moll.
\newblock {Income and Wealth Distribution in Macroeconomics: A Continuous-Time
  Approach}.
\newblock \emph{{The Review of Economic Studies}}, 89\penalty0 (1):\penalty0
  45--86, Jan. 2022.
\newblock \doi{10.1093/restud/rdab002}.
\newblock URL \url{https://hal.science/hal-03886376}.

\bibitem[Ansel et~al.(2024)Ansel, Yang, He, Gimelshein, Jain, Voznesensky, Bao,
  Bell, Berard, Burovski, Chauhan, Chourdia, Constable, Desmaison, DeVito,
  Ellison, Feng, Gong, Gschwind, Hirsh, Huang, Kalambarkar, Kirsch, Lazos,
  Lezcano, Liang, Liang, Lu, Luk, Maher, Pan, Puhrsch, Reso, Saroufim,
  Siraichi, Suk, Suo, Tillet, Wang, Wang, Wen, Zhang, Zhao, Zhou, Zou, Mathews,
  Chanan, Wu, and Chintala]{pytorch2}
J.~Ansel, E.~Yang, H.~He, N.~Gimelshein, A.~Jain, M.~Voznesensky, B.~Bao,
  P.~Bell, D.~Berard, E.~Burovski, G.~Chauhan, A.~Chourdia, W.~Constable,
  A.~Desmaison, Z.~DeVito, E.~Ellison, W.~Feng, J.~Gong, M.~Gschwind, B.~Hirsh,
  S.~Huang, K.~Kalambarkar, L.~Kirsch, M.~Lazos, M.~Lezcano, Y.~Liang,
  J.~Liang, Y.~Lu, C.~Luk, B.~Maher, Y.~Pan, C.~Puhrsch, M.~Reso, M.~Saroufim,
  M.~Y. Siraichi, H.~Suk, M.~Suo, P.~Tillet, E.~Wang, X.~Wang, W.~Wen,
  S.~Zhang, X.~Zhao, K.~Zhou, R.~Zou, A.~Mathews, G.~Chanan, P.~Wu, and
  S.~Chintala.
\newblock Pytorch 2: Faster machine learning through dynamic python bytecode
  transformation and graph compilation.
\newblock In \emph{29th ACM International Conference on Architectural Support
  for Programming Languages and Operating Systems, Volume 2 (ASPLOS '24)}. ACM,
  Apr. 2024.
\newblock \doi{10.1145/3620665.3640366}.
\newblock URL \url{https://pytorch.org/assets/pytorch2-2.pdf}.

\bibitem[Baker et~al.(2019)Baker, Alexander, Bremer, Hagberg, Kevrekidis, Najm,
  Parashar, Patra, Sethian, Wild, Willcox, and Lee]{scientific-ml}
N.~Baker, F.~Alexander, T.~Bremer, A.~Hagberg, Y.~Kevrekidis, H.~Najm,
  M.~Parashar, A.~Patra, J.~Sethian, S.~Wild, K.~Willcox, and S.~Lee.
\newblock Workshop report on basic research needs for scientific machine
  learning: Core technologies for artificial intelligence.
\newblock Technical report, USDOE Office of Science, 2 2019.
\newblock URL \url{https://www.osti.gov/biblio/1478744}.

\bibitem[Bischof and Kraus(2021)]{loss-balancing}
R.~Bischof and M.~Kraus.
\newblock Multi-objective loss balancing for physics-informed deep learning,
  2021.
\newblock URL \url{http://rgdoi.net/10.13140/RG.2.2.20057.24169}.

\bibitem[Boyce and DiPrima(2012)]{boyce2012FDM}
W.~Boyce and R.~DiPrima.
\newblock \emph{Elementary Differential Equations and Boundary Value Problems}.
\newblock Wiley, 2012.
\newblock ISBN 9781118157381.
\newblock URL \url{https://books.google.ca/books?id=vf_qMgEACAAJ}.

\bibitem[Brunnermeier and Sannikov(2014)]{Brunnermeier2014}
M.~K. Brunnermeier and Y.~Sannikov.
\newblock A macroeconomic model with a financial sector.
\newblock \emph{American Economic Review}, 104\penalty0 (2):\penalty0
  379–421, 2 2014.
\newblock \doi{10.1257/aer.104.2.379}.
\newblock URL \url{https://www.aeaweb.org/articles?id=10.1257/aer.104.2.379}.

\bibitem[Brunnermeier and Sannikov(2016)]{Brunnermeier2016}
M.~K. Brunnermeier and Y.~Sannikov.
\newblock Macro, money and finance: A continuous time approach.
\newblock Working Paper 22343, National Bureau of Economic Research, 6 2016.
\newblock URL \url{http://www.nber.org/papers/w22343}.

\bibitem[d'Avernas et~al.(2021{\natexlab{a}})d'Avernas, Petersen, and
  Vandeweyer]{pymacrofin}
A.~d'Avernas, D.~Petersen, and Q.~Vandeweyer.
\newblock Macro-financial modeling in python: Pymacrofin, 2021{\natexlab{a}}.
\newblock URL \url{https://adriendavernas.com/pymacrofin/index.html}.

\bibitem[d'Avernas et~al.(2021{\natexlab{b}})d'Avernas, Petersen, and
  Vandeweyer]{pymacrofinsolutionmethod}
A.~d'Avernas, D.~Petersen, and Q.~Vandeweyer.
\newblock A solution method for continuous-time general equilibrium models,
  2021{\natexlab{b}}.
\newblock URL \url{http://www.adriendavernas.com/papers/solutionmethod.pdf}.

\bibitem[Di~Tella(2017)]{ditella2017}
S.~Di~Tella.
\newblock Uncertainty shocks and balance sheet recessions.
\newblock \emph{Journal of Political Economy}, 125\penalty0 (6):\penalty0
  2038--2081, 2017.
\newblock \doi{10.1086/694290}.
\newblock URL \url{https://doi.org/10.1086/694290}.

\bibitem[Duffie and Epstein(1992)]{Duffie1992StochasticDU}
D.~Duffie and L.~G. Epstein.
\newblock Stochastic differential utility.
\newblock \emph{Econometrica}, 60:\penalty0 353--394, 1992.
\newblock URL \url{https://api.semanticscholar.org/CorpusID:51787219}.

\bibitem[Ebrahimi~Kahou et~al.(2021)Ebrahimi~Kahou, Fernández-Villaverde,
  Perla, and Sood]{hd-dp}
M.~Ebrahimi~Kahou, J.~Fernández-Villaverde, J.~Perla, and A.~Sood.
\newblock Exploiting symmetry in high-dimensional dynamic programming.
\newblock Working Paper 28981, National Bureau of Economic Research, July 2021.
\newblock URL \url{http://www.nber.org/papers/w28981}.

\bibitem[Epstein and Zin(1989)]{epstein-zin}
L.~G. Epstein and S.~E. Zin.
\newblock Substitution, risk aversion, and the temporal behavior of consumption
  and asset returns: A theoretical framework.
\newblock \emph{Econometrica}, 57\penalty0 (4):\penalty0 937--969, 1989.
\newblock ISSN 00129682, 14680262.
\newblock URL \url{http://www.jstor.org/stable/1913778}.

\bibitem[Fan et~al.(2023)Fan, Qiao, Jiao, Gu, Li, and
  Lu]{deepxde-macro-problem}
B.~Fan, E.~Qiao, A.~Jiao, Z.~Gu, W.~Li, and L.~Lu.
\newblock Deep learning for solving and estimating dynamic macro-finance
  models.
\newblock \emph{arXiv preprint arXiv:2305.09783}, 2023.
\newblock URL \url{https://arxiv.org/abs/2305.09783}.

\bibitem[Fernández-Villaverde et~al.(2020)Fernández-Villaverde, Nuño,
  Sorg-Langhans, and Vogler]{dp-dynamic}
J.~Fernández-Villaverde, G.~Nuño, G.~Sorg-Langhans, and M.~Vogler.
\newblock Solving high-dimensional dynamic programming problems using deep
  learning.
\newblock Working paper, September 2020.
\newblock URL
  \url{https://maximilianvogler.github.io/My_Website/Deep_Learning.pdf}.

\bibitem[Gomez(2017)]{gomez2017}
M.~Gomez.
\newblock Asset prices and wealth inequality.
\newblock 2017 Meeting Papers 1155, Society for Economic Dynamics, 2017.
\newblock URL \url{https://EconPapers.repec.org/RePEc:red:sed017:1155}.

\bibitem[Goodfellow et~al.(2016)Goodfellow, Bengio, and
  Courville]{deep-learning}
I.~Goodfellow, Y.~Bengio, and A.~Courville.
\newblock \emph{Deep Learning}.
\newblock MIT Press, 2016.
\newblock URL \url{http://www.deeplearningbook.org}.

\bibitem[Gopalakrishna(2021)]{ALIENs}
G.~Gopalakrishna.
\newblock Aliens and continuous time economies.
\newblock Swiss Finance Institute Research Paper Series 21-34, Swiss Finance
  Institute, 2021.
\newblock URL \url{https://EconPapers.repec.org/RePEc:chf:rpseri:rp2134}.

\bibitem[Grossmann et~al.(2007)Grossmann, Roos, and Stynes]{grossmann2007FDM}
C.~Grossmann, H.-G. Roos, and M.~Stynes.
\newblock \emph{Numerical Treatment of Partial Differential Equations}.
\newblock Springer Science \& Business Media, 2007.

\bibitem[Han et~al.(2018)Han, Jentzen, and E]{black-scholes-hd}
J.~Han, A.~Jentzen, and W.~E.
\newblock Solving high-dimensional partial differential equations using deep
  learning.
\newblock \emph{Proceedings of the National Academy of Sciences}, 115\penalty0
  (34):\penalty0 8505--8510, 2018.
\newblock \doi{10.1073/pnas.1718942115}.
\newblock URL \url{https://www.pnas.org/doi/abs/10.1073/pnas.1718942115}.

\bibitem[Hoffman and Frankel(2001)]{stability1}
J.~D. Hoffman and S.~P. Frankel.
\newblock \emph{Numerical methods for engineers and scientists}.
\newblock CRC Press, 2001.

\bibitem[Hornik et~al.(1989)Hornik, Stinchcombe, and White]{universal-approx}
K.~Hornik, M.~Stinchcombe, and H.~White.
\newblock Multilayer feedforward networks are universal approximators.
\newblock \emph{Neural Networks}, 2\penalty0 (5):\penalty0 359--366, 1989.
\newblock ISSN 0893-6080.
\newblock \doi{https://doi.org/10.1016/0893-6080(89)90020-8}.
\newblock URL
  \url{https://www.sciencedirect.com/science/article/pii/0893608089900208}.

\bibitem[Jaluria and Atluri(1994)]{stability2}
Y.~Jaluria and S.~N. Atluri.
\newblock Computational heat transfer.
\newblock \emph{Computational Mechanics}, 14\penalty0 (5):\penalty0 385--386,
  1994.

\bibitem[Kingma and Ba(2017)]{Adam}
D.~P. Kingma and J.~Ba.
\newblock Adam: A method for stochastic optimization.
\newblock \emph{ArXiv}, abs/1412.6980, 2017.

\bibitem[Kirk(1970)]{hjb1}
D.~E. Kirk.
\newblock \emph{Optimal Control Theory: An Introduction}.
\newblock Courier Corporation, 1970.

\bibitem[Liu and Nocedal(1989)]{lbfgs}
D.~C. Liu and J.~Nocedal.
\newblock On the limited memory method for large scale optimization.
\newblock \emph{Mathematical Programming B}, 45\penalty0 (3):\penalty0
  503--528, 1989.

\bibitem[Liu et~al.(2024)Liu, Wang, Vaidya, Ruehle, Halverson,
  Solja{\v{c}}i{\'c}, Hou, and Tegmark]{kan}
Z.~Liu, Y.~Wang, S.~Vaidya, F.~Ruehle, J.~Halverson, M.~Solja{\v{c}}i{\'c},
  T.~Y. Hou, and M.~Tegmark.
\newblock Kan: Kolmogorov-arnold networks.
\newblock \emph{arXiv preprint arXiv:2404.19756}, 2024.

\bibitem[Loshchilov and Hutter(2019)]{AdamW}
I.~Loshchilov and F.~Hutter.
\newblock Decoupled weight decay regularization.
\newblock \emph{ArXiv}, abs/1711.05101, 2019.

\bibitem[Lu et~al.(2021{\natexlab{a}})Lu, Jin, Pang, Zhang, and
  Karniadakis]{deeponet}
L.~Lu, P.~Jin, G.~Pang, Z.~Zhang, and G.~E. Karniadakis.
\newblock Learning nonlinear operators via deeponet based on the universal
  approximation theorem of operators.
\newblock \emph{Nature Machine Intelligence}, 3\penalty0 (3):\penalty0
  218–229, 3 2021{\natexlab{a}}.
\newblock ISSN 2522-5839.
\newblock \doi{10.1038/s42256-021-00302-5}.
\newblock URL \url{http://dx.doi.org/10.1038/s42256-021-00302-5}.

\bibitem[Lu et~al.(2021{\natexlab{b}})Lu, Meng, Mao, and Karniadakis]{deepxde}
L.~Lu, X.~Meng, Z.~Mao, and G.~E. Karniadakis.
\newblock Deepxde: A deep learning library for solving differential equations.
\newblock \emph{SIAM Review}, 63\penalty0 (1):\penalty0 208--228,
  2021{\natexlab{b}}.
\newblock \doi{10.1137/19M1274067}.
\newblock URL \url{https://doi.org/10.1137/19M1274067}.

\bibitem[Martin(2013)]{Martin2013}
I.~Martin.
\newblock The lucas orchard.
\newblock \emph{Econometrica}, 81:\penalty0 55--111, 1 2013.
\newblock ISSN 1468-0262.
\newblock \doi{10.3982/ECTA8446}.
\newblock URL \url{https://onlinelibrary.wiley.com/doi/full/10.3982/ECTA8446
  https://onlinelibrary.wiley.com/doi/abs/10.3982/ECTA8446
  https://onlinelibrary.wiley.com/doi/10.3982/ECTA8446}.

\bibitem[McCann and Zhang(2023)]{rochet-chone-comment}
R.~J. McCann and K.~S. Zhang.
\newblock Comment on "ironing, sweeping, and multidimensional screening'',
  2023.
\newblock URL \url{https://arxiv.org/abs/2311.13012}.

\bibitem[Mérigot and Oudet(2014)]{rochet-chone-solution}
Q.~Mérigot and E.~Oudet.
\newblock Handling convexity-like constraints in variational problems, 2014.
\newblock URL \url{https://arxiv.org/abs/1403.2340}.

\bibitem[Paszke et~al.(2017)Paszke, Gross, Chintala, Chanan, Yang, DeVito, Lin,
  Desmaison, Antiga, and Lerer]{pytorch-autodiff}
A.~Paszke, S.~Gross, S.~Chintala, G.~Chanan, E.~Yang, Z.~DeVito, Z.~Lin,
  A.~Desmaison, L.~Antiga, and A.~Lerer.
\newblock Automatic differentiation in pytorch.
\newblock In \emph{NIPS-W}, 2017.

\bibitem[Quarteroni and Valli(2008)]{quarteroni2008FEM}
A.~Quarteroni and A.~Valli.
\newblock \emph{Numerical Approximation of Partial Differential Equations},
  volume~23.
\newblock Springer Science \& Business Media, 2008.

\bibitem[Raissi et~al.(2017)Raissi, Perdikaris, and Karniadakis]{pinn2017}
M.~Raissi, P.~Perdikaris, and G.~E. Karniadakis.
\newblock Physics informed deep learning (part i): Data-driven solutions of
  nonlinear partial differential equations.
\newblock \emph{arXiv preprint arXiv:1711.10561}, 2017.

\bibitem[Raissi et~al.(2019)Raissi, Perdikaris, and Karniadakis]{pinn2019}
M.~Raissi, P.~Perdikaris, and G.~E. Karniadakis.
\newblock Physics-informed neural networks: A deep learning framework for
  solving forward and inverse problems involving nonlinear partial differential
  equations.
\newblock \emph{Journal of Computational Physics}, 378:\penalty0 686--707,
  2019.

\bibitem[Rochet and Choné(1998)]{rochet-chone}
J.-C. Rochet and P.~Choné.
\newblock Ironing, sweeping, and multidimensional screening.
\newblock \emph{Econometrica}, 66\penalty0 (4):\penalty0 783--826, 1998.
\newblock ISSN 00129682, 14680262.
\newblock URL \url{http://www.jstor.org/stable/2999574}.

\bibitem[Shen et~al.(2022)Shen, Shao, Zhou, Jiang, Luo, and
  Yang]{high-order-derivatives2}
S.~Shen, T.~Shao, K.~Zhou, C.~Jiang, F.~Luo, and Y.~Yang.
\newblock Hod-net: High-order differentiable deep neural networks and
  applications.
\newblock \emph{Proceedings of the AAAI Conference on Artificial Intelligence},
  36\penalty0 (8):\penalty0 8249--8258, 2022.
\newblock \doi{10.1609/aaai.v36i8.20799}.
\newblock URL \url{https://ojs.aaai.org/index.php/AAAI/article/view/20799}.

\bibitem[Shukla et~al.(2024)Shukla, Toscano, Wang, Zou, and
  Karniadakis]{kan-mlp}
K.~Shukla, J.~D. Toscano, Z.~Wang, Z.~Zou, and G.~E. Karniadakis.
\newblock A comprehensive and fair comparison between mlp and kan
  representations for differential equations and operator networks.
\newblock \emph{arXiv preprint arXiv:2406.02917}, 2024.
\newblock URL \url{https://arxiv.org/abs/2406.02917}.

\bibitem[Sirignano and Spiliopoulos(2018)]{Sirignano2018}
J.~Sirignano and K.~Spiliopoulos.
\newblock {DGM: A deep learning algorithm for solving partial differential
  equations}.
\newblock \emph{Journal of Computational Physics}, 2018.
\newblock ISSN 10902716.
\newblock \doi{10.1016/j.jcp.2018.08.029}.

\bibitem[Song et~al.(2024)Song, Wang, Yang, Taccari, and Chen]{loss-attention}
Y.~Song, H.~Wang, H.~Yang, M.~L. Taccari, and X.~Chen.
\newblock Loss-attentional physics-informed neural networks.
\newblock \emph{Journal of Computational Physics}, 501:\penalty0 112781, 2024.
\newblock ISSN 0021-9991.
\newblock \doi{https://doi.org/10.1016/j.jcp.2024.112781}.
\newblock URL
  \url{https://www.sciencedirect.com/science/article/pii/S0021999124000305}.

\bibitem[Wang et~al.(2023)Wang, Li, and Li]{deepxde-option-pricing}
X.~Wang, J.~Li, and J.~Li.
\newblock A deep learning based numerical pde method for option pricing.
\newblock \emph{Computational Economics}, 62:\penalty0 149--164, 2023.
\newblock \doi{https://doi.org/10.1007/s10614-022-10279-x}.

\bibitem[Yong and Zhou(1999)]{hjb2}
J.~Yong and X.~Y. Zhou.
\newblock \emph{Stochastic Controls: Hamiltonian Systems and HJB Equations}.
\newblock Springer, 1999.

\bibitem[Zhu and Yang(2021)]{high-order-derivatives}
Q.~Zhu and J.~Yang.
\newblock A local deep learning method for solving high order partial
  differential equations.
\newblock \emph{arXiv preprint arXiv:2103.08915}, 2021.
\newblock URL \url{https://arxiv.org/abs/2103.08915}.

\bibitem[Ŝolín(2005)]{solín2005FEM}
P.~Ŝolín.
\newblock \emph{Partial Differential Equations and the Finite Element Method}.
\newblock John Wiley \& Sons, 2005.

\end{thebibliography}


\clearpage
\newpage
\appendix

\section{Appendix: training algorithms}\label{appendix:training-algorithms}

Algorithm~\ref{algo:basic-training-step} shows the basic training algorithm, similar to all standard training procedures. Algorithm~\ref{algo:time-stepping} shows the training procedure for time-stepping scheme adopted from traditional numerical solutions to macro-finance model, as outlined in Section~\ref{sec:time-stepping-scheme}. In the time-stepping scheme, the initial guesses $\hat{V}_{i,\tau=0}$ and $\hat{E}_{i,\tau=0}$ can be set either as constants or based on precomputed guess function values. A good initial guess, closer to the true solution, may reduce the number of outer loop iterations, but it generally does not affect the overall convergence of the method.

\begin{algorithm}[!htb]
\caption{Basic training algorithm (single step)}\label{algo:basic-training-step}
\textbf{Input}: Neural networks $\bs{a_1,...,a_n, e_1,...,e_m}$, with parameters $\theta$
\begin{algorithmic}[1]
\raggedright
\STATE Sample a batch of state variables $X=(x_1,...,x_d)$
\FORALL{$v_i \in \{a_1,...,a_n, e_1,...,e_m\}$}
    \STATE Compute $v_i(\theta, X)$, and all associated derivatives
\ENDFOR

\FORALL{eq $\in$ equations}
    \STATE Update variables defined by eq.lhs using eq.rhs
\ENDFOR

\STATE Compute losses from conditions, constraints, endogenous / HJB equations and systems
\STATE Compute the total loss $\mathcal{L}(\theta, \mathcal{T})$ with \eqref{eq:total-loss}
\STATE Backward propagation and update $\theta$ to minimize $\mathcal{L}(\theta, \mathcal{T})$
\end{algorithmic}
\end{algorithm}

\begin{algorithm}[!htb]
\caption{Time-stepping scheme}\label{algo:time-stepping}
\raggedright
\textbf{Input}: $X$: state variables with time $t$,\\
$V_i: X \to \RR$: agent value variables,\\
$E_j: X \to \RR$: endogenous variables,\\
\textbf{Output}: Trained approximations $\hat{V}_i$, $\hat{E}_j$.
\begin{algorithmic}[1]
\raggedright
\STATE $\tau\gets 0, \hat{V}_{i,\tau=0} = 1$, $\hat{E}_{i,\tau=0}=1$ \COMMENT{Initialize as constant}
\WHILE {True}
    \STATE Sample $X=(x_0,...,x_n, t)$ uniformly random from domain ($x_0,...,x_n$ are defined by the problem domain, $t\in [0,1]$)
    \STATE{Embed boundary conditions: $V_{i, \tau+1}(t=1)=\hat{V}_{i, \tau}$, $E_{i,\tau+1}(t=1)=\hat{E}_{i,\tau}$} \COMMENT{At maximum time, the functions should satisfy the value from the previous step}
    \WHILE{True}
        \STATE{Compute variables using neural networks, with $\frac{\partial V_i}{\partial t}$, and $\frac{\partial E_i}{\partial t}$ integrated.}
        \STATE{Compute loss on boundary conditions, endogenous equations, HJB equations and systems}
        \STATE{Compute total loss}
        \IF {iter $\geq$ max\_inner\_loop OR inner loss converges}
            \STATE{break}
        \ENDIF
    \ENDWHILE
    \STATE{$\hat{V}_{i,\tau+1}\gets V_{i,\tau+1}(t=0)$, $\hat{E}_{i,\tau+1}\gets E_{i,\tau+1}(t=0), \tau\gets \tau+1$}
    \IF {iter $\geq$ max\_outer\_loop OR $\hat{V}_i$, $\hat{E}_i$ converge}
        \STATE{break}
    \ENDIF
\ENDWHILE
\end{algorithmic}
\end{algorithm}

\section{Appendix: model details and additional examples}\label{appendix:models}
\subsection{Basic models}\label{appendix:basic-models}

This appendix provides the details for models used in Section~\ref{sec:basic-problems}.

\paragraph{Cauchy-Euler} 
\begin{align*}
    x^2 y'' + 6xy' + 4y &=0, \quad x\in [1,2]\\
    y(1)=6, \quad &y(2)=\frac{5}{4},
\end{align*}
with solution $y=4x^{-4} + 2 x^{-1}$. The MSE is computed on 50 equally spaced inputs $x\in \bs{1.0, 1.02, ..., 2.0}$. A sample symbolic formula provided by KAN is:
\begin{align*}
    y=-16.7436 + \frac{3.3016}{\bb{-1 + \frac{0.0214}{\bb{1-\frac{0.0069}{0.0246x+0.0034}}^4 + \frac{0.3063}{\bb{-1 + \frac{0.0119}{(x+0.1466)^4}}^4}}}^4}.
\end{align*}

\paragraph{Diffusion}
\begin{align*}
    \frac{\partial y}{\partial t} &= \frac{\partial^2 y}{\partial x^2} - e^{-t} (\sin(\pi x) - \pi^2 \sin (\pi x)), \quad (x,t)\in [-1,1] \times [0,1]\\
    y(x,0) &= \sin(\pi x), \quad y(-1,t)=y(1,t)=0.
\end{align*}
The solution is $y=e^{-t}\sin(\pi x)$. The MSE is reported over a $50\times 50$ equally-spaced grid in the problem domain $(x,t)\in [-1,1] \times [0,1]$. A sample symbolic formula provided by KAN is:
\begin{align*}
    & -3.2527\exp\bb{0.0735\sin(2.3027x + 2.1834) + 0.0915 \exp(-0.9166t)} \\
    &+ 0.2259\exp\bb{-0.8219\sin(2.5074x + 3.8264) + 0.5529\sin(0.9181t-4.1949)} \\
    &- 0.1011\exp\bb{-0.3498\sin(3.8952x - 6.2036) + 1.2514 \sin(1.7026 t + 1.8198)}\\
    & + 0.5222\sin(3.8749\sin (0.9643x + 8.3807) - 1.8474 + 0.6493\exp(-2.3867t)) \\
    &- 0.597\sin\bb{1.1602\sin(1.848 x - 0.7975) - 1.2157\sin(1.1028t + 6.4883) + 3.8254} + 3.278.
\end{align*}
Note that the complexity of the KAN formula depends on the width and depth of the KAN layers. In this case, we use a 2-input, 1-output KAN model with a single hidden layer of width 5. The symbolic formula is a linear combination of five sine and exponential functions, with the inputs to each function being linear combinations of two sine and exponential functions, corresponding to the defined layer width. In comparison, the equation for Cauchy-Euler equation has a more nested structure, as it uses a 1-input, 1-output KAN model with two hidden layers.

\paragraph{Black-Scholes}
This section provides more details on the generalized $N$-asset Black-Scholes model:
\begin{align*}
    &\frac{\partial V}{\partial t} + r S \cdot \nabla V + \frac{\sigma^2}{2} S^T \rho \nabla_S^2 V S- r V = 0, (S,t)\in [0,1]^{N+1},\\
    &V(S,T) = \max\bs{\frac{1}{n}\sum_{i=1}^N S_i - K, 0}
\end{align*}
where $S=\bb{S_1,...,S_N}$ are the normalized underlying asset prices and $t$ is normalized by time to maturity $T$. The constants are $\sigma=0.2$ (volatility), $r=0.05$ (risk-free rate), $K=0.5$ (strike price), and $\rho$ is a correlation matrix of the underlying asset processes, with $\rho_{ij}=\begin{cases}
    1, i=j,\\
    0.5, i\neq j,
\end{cases}$. This generalizes the one-asset model from \cite{black-scholes-hd}, including the correlation factor and excluding default risk.
The one-asset (2D) model has an analytic solution:
\begin{align*}
    V(t, S) &= S \Phi(d_+) - K e^{-r(T-t)} \Phi(d_-),\\
    d_{\pm} &= \frac{\log (S/K) + (r \pm 0.5\sigma^2)(T-t)}{\sigma\sqrt{T-t}},
\end{align*}
where $\Phi$ is the CDF of standard normal distribution. A sample symbolic formula provided by KAN is:
\begin{align*}
    V(t,S) = -0.0654 + 0.5868\exp\bb{-8.7782 \bb{0.52 + 0.9879\exp\bb{-0.9284(0.1783-S)^2} - \exp(-0.0245t)}^2},
\end{align*}

\paragraph{Laplacian}
\begin{align*}
    \Delta u &= 0, x\in \Omega = [0,1]^n\\
    \text{Zero boundary: } u(x) &= 0, x\in \partial\Omega,\\
    \text{Summation boundary: } u(x) &= \sum_{i=1}^n x_i, x\in\partial\Omega
\end{align*} 
Since the boundary conditions are harmonic, the maximum principle gives the analytical solution $u(x)=0$ for zero boundary condition and $u(x) = \sum_{i=1}^n x_i$ for summation boundary condition.

Table~\ref{tab:efficiency-benchmark} provides the exact values of mean epoch time (in seconds), memory usage (in MB), and FLOPs (floating-point operations) for the Laplace equation in 2, 5, 10, 20, 50, and 100 dimensions. The time is computed using Python's built-in \texttt{time} package. Peak memory usage is recorded with \texttt{torch.cuda.max\_memory\_allocated()}, and FLOPs are estimated using the PyTorch Profiler\footnote{\url{https://pytorch.org/tutorials/recipes/recipes/profiler_recipe.html}}. 
DeepXDE's memory usage and FLOPs grow much faster than Deep-MacroFin's. This demonstrates the potential of extending Deep-MacroFin to more complex economic models to higher dimensions.

\begin{table}[!htb]
\caption{Time, memory and FLOPs benchmark with DeepXDE}\label{tab:efficiency-benchmark}

\centering
\resizebox*{\textwidth}{!}{
\begin{tabular}{lcccccc}
\toprule
    & \multicolumn{2}{c}{Mean Epoch Time (s)} & \multicolumn{2}{c}{CUDA Memory (MB)} & \multicolumn{2}{c}{FLOPS ($\times 10^9$)} \\
N-dim & Deep-MacroFin & DeepXDE & Deep-MacroFin & DeepXDE & Deep-MacroFin & DeepXDE \\
\cmidrule(lr){1-1} \cmidrule(lr){2-3} \cmidrule(lr){4-5} \cmidrule(lr){6-7}
2 & 0.0216 & 0.0122 & 104.21 & 39.03 & 0.29 & 0.35 \\
5 & 0.0259 & 0.0222 & 266.44 & 93.95 & 0.50 & 1.28 \\
10 & 0.0330 & 0.0389 & 477.83 & 257.27 & 0.87 & 4.19 \\
20 & 0.0469 & 0.0732 & 869.67 & 880.69 & 1.70 & 15.88 \\
50 & 0.0901 & 0.1772 & 1955.26 & 5243.34 & 4.89 & 112.15 \\
100 & 0.1637 & 0.4636 & 4421.12 & 21189.79 & 12.53 & 571.58 \\
\bottomrule
\end{tabular}}
\end{table}

\paragraph{Model setup}
Table~\ref{tab:model-setup} shows the model setup.

\begin{table}[!htb]
\caption{Model setup}\label{tab:model-setup}

\vspace{.1in}
\centering
\resizebox*{\textwidth}{!}{
\begin{tabular}{cccccccc}
\toprule
PDE & Model & Hidden & \#Params & Activation & Epochs & Optimizer & Learning rate \\
\midrule
Cauchy-Euler & DeepXDE & [30]*4 & 2881 & tanh & 5000 & Adam & $10^{-3}$\\
 & MLP & [30]*4 & 2881 & tanh & 5000 & Adam & $10^{-3}$\\
 & KAN & [2,1] & 94 & SiLU & 100 & L-BFGS & $1$\\
Diffusion & DeepXDE & [30]*4 & 2911 & tanh & 5000 & Adam & $10^{-3}$\\
 & MLP & [30]*4 & 2911 & tanh & 5000 & Adam & $10^{-3}$\\
 & KAN & [5] & 276 & SiLU & 100  & L-BFGS & $1$\\
Black-Scholes & DeepXDE & [30]*4 & 2911 & SiLU & 5000 & Adam & $10^{-3}$\\
 & MLP & [30]*4 & 2911 & SiLU & 5000 & Adam & $10^{-3}$\\
 & KAN & [1] & 56 & SiLU & 100 & L-BFGS & $1$\\
Laplacian (100D) & DeepXDE & [30]*4 & 5851 & SiLU & 10000 & Adam & $10^{-3}$\\
 & MLP & [30]*4 & 5851 & SiLU & 10000 & Adam & $10^{-3}$\\
\bottomrule
\end{tabular}}
\end{table}

\subsection{Free boundary problems}\label{appendix:free-boundary-model}

This appendix provides the details for models used in Section~\ref{sec:free-boundary}. 

\paragraph{Obstacle problem} Let $\Omega\subset \RR^n$ be a simply connected open set with smooth boundary $\partial\Omega$. Consider a smooth obstacle function $w\in C^2(\overline{\Omega})$ s.t. the solution $u$ must satisfy $u\geq w$ for all $x\in \Omega$. The goal is to minimize the Dirichlet energy over $\Omega$, subject to this inequality constraint and prescribed boundary conditions. The variational problem can be formulated as:
\begin{align*}
  \min_u E(u) &= \min_u \frac{1}{2} \int_\Omega \norm{\nabla u}^2\\
  \text{s.t. } u(x) &= g(x), \quad x\in \partial\Omega\\
  u(x) &\geq w(x), \quad x\in \Omega, w\in C^2(\overline{\Omega}) 
\end{align*}

Define the admissible set:
\[V = \bs{v\in H^1(\Omega) : v\geq w \text{ in } \Omega; v=g \text{ on }\partial \Omega}.\]
The goal is to find $u\in V$ s.t. $E(u)\leq E(v)$, $\forall v\in V$. For simplicity, assume $g=0$ constant.

\begin{figure}[!htb]
\centering

\begin{subfigure}[b]{0.24\linewidth}
\centering
\includegraphics[width=\linewidth]{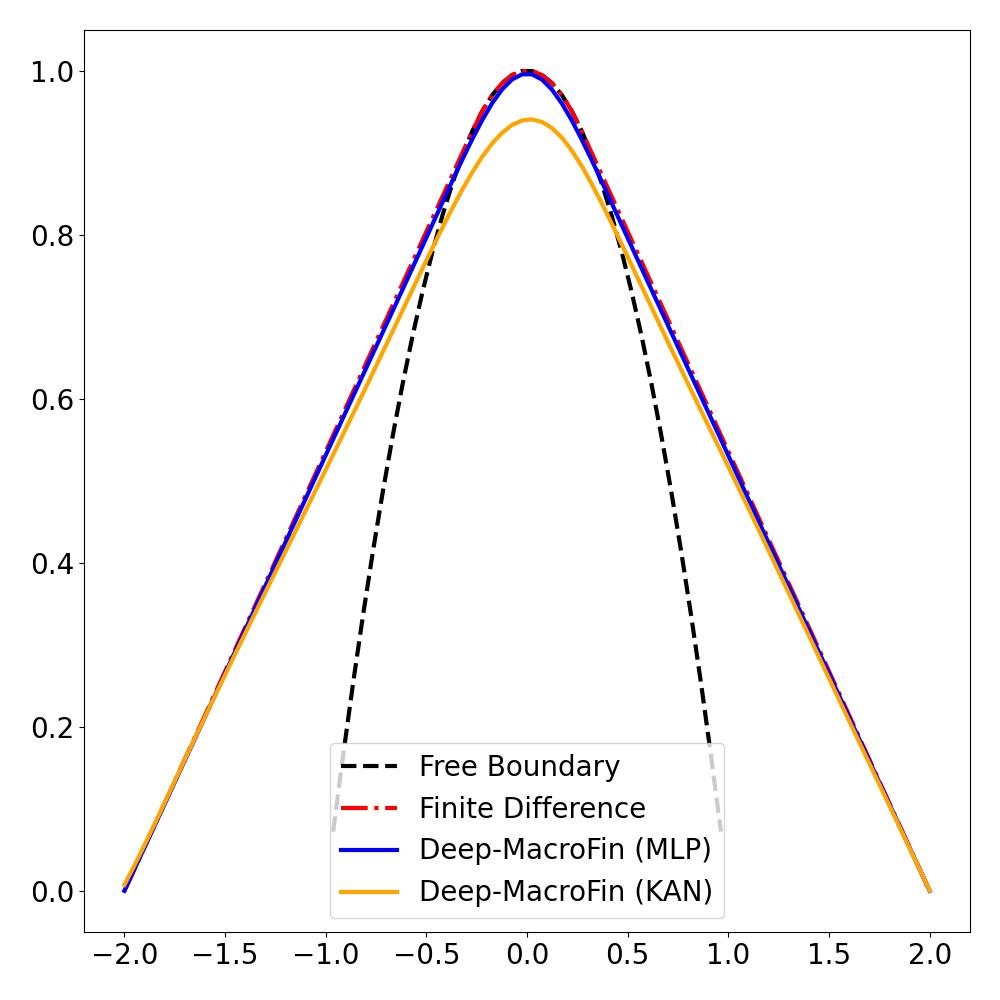}
\caption{1D}
\end{subfigure}
\begin{subfigure}[b]{0.48\linewidth}
\centering
\includegraphics[width=\linewidth]{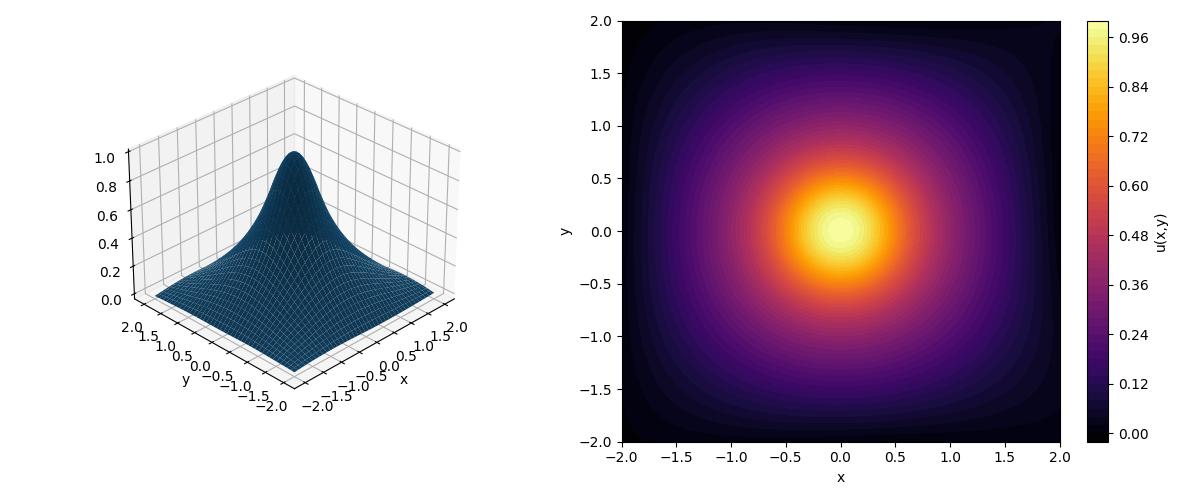}
\caption{2D}
\end{subfigure}

\caption{Obstacle problem}
\label{fig:obstacle-problem}
\end{figure}

Firstly, we derive the weak form and the associated Euler-Lagrange inequality for the variational problem. Let $u\in V$ be the minimizer, $v\in V$, $t\in \RR$. Apply a perturbation to $E(u)$:
\[E(u+t(v-u)) = \frac{1}{2} \innerprod{\nabla u + t\nabla (v-u), \nabla u + t \nabla (v-u)}.\]
Since $u$ is a minimizer, the derivative w.r.t. $t$ at $t=0$ should be non-negative:
\begin{align*}
  \frac{d}{dt} E(u+t(v-u))|_{t=0} &= \left.\frac{1}{2} \bb{ \innerprod{\nabla(v-u), \nabla u + t\nabla (v-u)} + \innerprod{\nabla u + t\nabla (v-u), \nabla(v-u)}} \right|_{t=0}\\
  &= \left.\innerprod{\nabla u + t\nabla (v-u), \nabla(v-u)} \right|_{t=0}\\
  &= \innerprod{\nabla u, \nabla (v-u)}\geq 0, \;\forall v\in V
\end{align*}
Now, choose a specific variation $v=u+\eta \psi$ for $\eta\in\RR$ small and $\psi\in C_C^\infty(\Omega)$ with $\psi\geq 0$ and $\psi=0$ on $\partial\Omega$. Substituting into the inequality yields:
\begin{align*}
  \innerprod{\nabla u,\nabla (v-u)} &= \int_\Omega \nabla u\nabla (v-u) = \int_\Omega \nabla u\nabla \eta \psi = \eta \int_\Omega \nabla u\nabla \psi\\
  &= \eta \bb{\int_{\partial\Omega} \nabla u \psi - \int_\Omega \nabla^2 u\psi}\\
  &= -\eta \int_\Omega \nabla^2 u \psi  = -\eta \innerprod{\nabla^2 u, \psi}\geq 0
\end{align*}
In the region where $u=w$, $\eta > 0$ so that $v\geq w$ is satisfied, this requires $\innerprod{\nabla^2 u, \psi} \leq 0$ for all $\psi$, so $\nabla^2 u \leq 0$ a.e. in $\Omega$. 
When $u>w$, $\eta$ can be both positive or negative, for the inequality to hold for all $\psi$, we need $\nabla^2 u =0$ a.e. in $\Omega$.
Thus, we obtain the following free boundary formulation of the obstacle problem:
\begin{align*}
  \nabla^2 u &=0,\; x\in \Omega \cap \bs{u>w} \\
  \nabla^2 u &\leq 0, \; x\in \Omega \cap \bs{u=w}\\
  u &\geq w,\; x\in \Omega\\
  u &= g, \; x\in\partial\Omega
\end{align*}
In summary, the solution $u$ lies above the obstacle $w$, is harmonic whenever it does not touch the obstacle ($u>w$), and it is superharmonic when it touches the obstacle ($u=w$). The interface between these two regions, where $u$ transitions from being strictly greater than $w$ to being equal to $w$, is known as the free boundary. We choose $w(x) = 1-x^2$ in 1D, and $w(x)=1-x_1^2-x_2^2$ in 2D and solve for $u$ over $[-2,2]^d$. The errors are computed over $50^d$ equally-spaced points on the domain with respect to numerical solutions from projected SOR method. In this example, MLP models perform better than KAN model, with smaller free boundary violation. Table~\ref{tab:obstacle-problem-errors} shows the errors, while Figure~\ref{fig:obstacle-problem} shows the graphical results of the tested models.

\begin{table}[!htb]
\caption{Obstacle problem errors}\label{tab:obstacle-problem-errors}

\centering
\resizebox*{\textwidth}{!}{%
\begin{tabular}{ccccc}
\toprule
PDE & Model & MSE & $\|u-\hat{u}\|_{L^\infty}$ & Free Boundary Violation\\
\midrule
Obstacle (1D) & MLP & $1.21 \times 10^{-4}$ ($\pm$ $8.58 \times 10^{-5}$) & $2.02 \times 10^{-2}$ ($\pm$ $5.47 \times 10^{-3}$) & $2.02 \times 10^{-2}$ ($\pm$ $5.47 \times 10^{-3}$) \\
    & KAN & $7.30 \times 10^{-4}$ ($\pm$ $9.68 \times 10^{-5}$) & $5.66 \times 10^{-2}$ ($\pm$ $3.09 \times 10^{-3}$) & $5.66 \times 10^{-2}$ ($\pm$ $3.09 \times 10^{-3}$) \\
    & KAN Symbolic & $1.34 \times 10^{-3}$ ($\pm$ $1.18 \times 10^{-4}$) & $8.36 \times 10^{-2}$ ($\pm$ $4.18 \times 10^{-3}$) & $5.83 \times 10^{-2}$ ($\pm$ $2.69 \times 10^{-3}$) \\
    & KAN-KAN Symbolic & $6.31 \times 10^{-4}$ ($\pm$ $4.64 \times 10^{-5}$) & $7.61 \times 10^{-2}$ ($\pm$ $4.19 \times 10^{-3}$) & $5.83 \times 10^{-2}$ ($\pm$ $2.69 \times 10^{-3}$) \\
Obstacle (2D) & MLP & $1.17 \times 10^{-3}$ ($\pm$ $1.72 \times 10^{-3}$) & $1.38 \times 10^{-1}$ ($\pm$ $1.20 \times 10^{-1}$) & $5.20 \times 10^{-2}$ ($\pm$ $1.17 \times 10^{-2}$) \\
    & KAN & $8.94 \times 10^{-3}$ ($\pm$ $2.64 \times 10^{-4}$) & $3.04 \times 10^{-1}$ ($\pm$ $7.72 \times 10^{-3}$) & $1.88 \times 10^{-1}$ ($\pm$ $4.34 \times 10^{-3}$) \\
    & KAN Symbolic & $8.84 \times 10^{-3}$ ($\pm$ $2.52 \times 10^{-4}$) & $3.24 \times 10^{-1}$ ($\pm$ $8.03 \times 10^{-3}$) & $1.69 \times 10^{-1}$ ($\pm$ $4.65 \times 10^{-3}$) \\
    & KAN-KAN Symbolic & $1.43 \times 10^{-5}$ ($\pm$ $1.46 \times 10^{-6}$) & $2.91 \times 10^{-2}$ ($\pm$ $1.41 \times 10^{-3}$) & $1.69 \times 10^{-1}$ ($\pm$ $4.65 \times 10^{-3}$) \\
\bottomrule
\end{tabular}}
\end{table}

\paragraph{Principal agent problem} Consider the Rochet–Choné formulation of the principal–agent problem \cite{rochet-chone,rochet-chone-comment}. The principal offers a menu of products represented by types $y\in [0,\infty)^n$ with $y=0$ representing the null product or outside option. 
Agents are heterogeneous and characterized by their type $x\in X\subset \RR^n$, distributed according to a measure $\mu \ll \calL$, where $\calL$ is the Lebesgue measure on $\RR^n$. There exists a probability density function $f(x) = \frac{d\mu}{dx}$.  
The principal selects a price menu $v(y)$ to maximize profit, subject to the constraint that outside option has value $v(0)=0$. Producing a product of type $y$ incurs a cost $c(y)$. 
The joint surplus generated from a transaction between an agent of type $x$ and a product $y$ is given by the bilinear form $b(x,y) = x\cdot y$. Agents choose products to maximize their utility:
\[u(x) = \sup_{y} b(x,y) - v(y),\]
which is the convex conjugate of the price function $v(y)$ w.r.t. the surplus $b(x,y)$.
Assuming the optimal menu $y(x) = \nabla u(x)$, the principal’s objective becomes the following variational problem:
\begin{align*}
  \max_{u,\nabla u, \nabla^2 u\geq  0} \Phi (u) &= \max_{u,\nabla u, \nabla^2 u\geq 0} \int_X (b(x,y) - u(x) - c(y)) d\mu \\
  & = \max_{u,\nabla u, \nabla^2 u\geq 0} \int_X  \bb{x\nabla u(x) - u(x) - c(\nabla u(x))} d\mu
\end{align*}
Choosing the quadratic production cost $c(y) = \frac{1}{2} \norm{y}^2$, this becomes:
\begin{align*}
  \max_{u,\nabla u, \nabla^2 u\geq 0} \Phi (u) &= \max_{u,\nabla u, \nabla^2 u\geq 0} \int_X  \bb{x\nabla u(x) - u(x) - \frac{1}{2} \norm{\nabla u}^2} d\mu
\end{align*}

In 1D with $X=[a,a+1]$, and uniform distribution $\frac{d\mu}{dx} = f(x) = \chi_{[a,a+1]} = \begin{cases}
    1, x\in [a,a+1]\\
    0, \text{ otherwise}
\end{cases}$, we get the free boundary problem:
\begin{align*}
    u'' &= 2, x\in [a,a+1] \cap \bs{u > 0}\\
    0\leq u'' &\leq 2, x\in [a,a+1] \cap \bs{u = 0}, \text{ equivalently } u''=0\\
    u(a) &= 0, u'(a+1)=a+1,
\end{align*}
The free boundary is at $x=\frac{a+1}{2}$. For $a\leq 1$, the solution is
\[u(x) = \begin{cases}
  0, x\in [a,\frac{a+1}{2}]\\
  \bb{x-\frac{a+1}{2}}^2, x\in [\frac{a+1}{2},a+1]
\end{cases}.\] 
When $a > 1$, the constraint $u\geq 0$ is inactive, and the solution becomes fully classical, $u(x) = \bb{x-\frac{a+1}{2}}^2 - \bb{\frac{a-1}{2}}^2$. 

In economic sense, the parameter $a$ is the lowest agent type considered by the principal.
When $a<1$, some low-type agents choose not to purchase (a buyer's market), and the region $\bs{x: u(x)=0}$ represents the set of excluded agents. The free boundary $x_0=\frac{a+1}{2}$ is the threshold type that is indifferent to participating. 
When $a\geq 1$, all agents participate (a seller's market), and the solution does not have a free boundary. Table~\ref{tab:principal-agent-problem-errors} reports the errors with respect to project SOR method in 1D. 

\begin{figure}[!htb]
\centering
\includegraphics[width=0.75\linewidth]{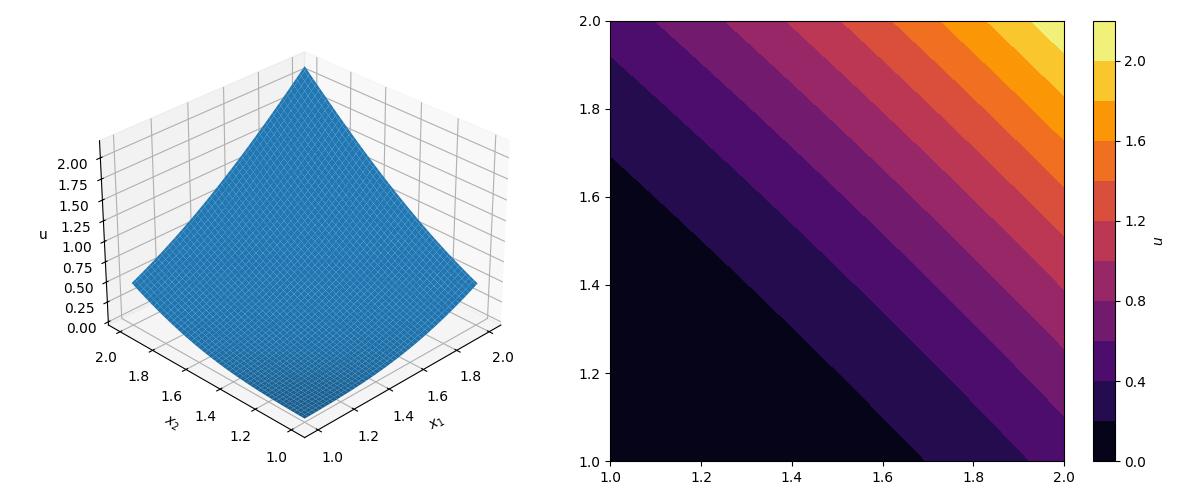}
\caption{Principal agent problem}
\label{fig:principal-agent-problem}
\end{figure}

In 2D, $X=[a,a+1]^2$. The free boundary is more difficult to be derived, and few numerical results are publicly available. In our implementation, we compute the functional $\Phi$ with simpson's method, set it as the HJB equation, and directly minimize the variational integral with the following constraints to ensure convexity and curvature.
\begin{align*}
    & \min \int_a^{a+1} \int_a^{a+1}  \bb{- x\nabla u(x) + u(x) + \frac{1}{2} \norm{\nabla u}^2} dx_1dx_2\\
    \text{s.t. } & u \geq 0, \nabla u \geq 0, \det H(u)\geq 0\\
    & 0\leq \frac{\partial^2 u}{\partial x_1^2} + \frac{\partial^2 u}{\partial x_2^2}\leq 3\\
    & u(0,0) = 0
\end{align*}
Figure~\ref{fig:principal-agent-problem} shows the solution in 2D using MLP, which aligns with numerical solutions as in \cite{rochet-chone-solution}. KAN is not trained due to inefficient computation and lack of stability.

\begin{table}[!htb]
\caption{Principal agent problem errors}\label{tab:principal-agent-problem-errors}

\centering
\resizebox*{\textwidth}{!}{%
\begin{tabular}{ccccc}
\toprule
PDE & Model & MSE & $\|u-\hat{u}\|_{L^\infty}$ & Free Boundary Violation\\
\midrule
Principal Agent ($a=0$) & MLP & $1.71 \times 10^{-4}$ ($\pm$ $1.24 \times 10^{-4}$) & $2.13 \times 10^{-2}$ ($\pm$ $9.89 \times 10^{-3}$) & 0.00 ($\pm$ 0.00) \\
    & KAN & $5.64 \times 10^{-5}$ ($\pm$ $5.11 \times 10^{-6}$) & $1.13 \times 10^{-2}$ ($\pm$ $5.70 \times 10^{-4}$) & $1.80 \times 10^{-6}$ ($\pm$ $3.77 \times 10^{-8}$) \\
    & KAN Symbolic & $6.15 \times 10^{-5}$ ($\pm$ $3.25 \times 10^{-6}$) & $1.32 \times 10^{-2}$ ($\pm$ $3.97 \times 10^{-4}$) & $1.71 \times 10^{-3}$ ($\pm$ $1.35 \times 10^{-3}$) \\
    & KAN-KAN Symbolic & $2.53 \times 10^{-6}$ ($\pm$ $5.45 \times 10^{-6}$) & $4.47 \times 10^{-3}$ ($\pm$ $1.74 \times 10^{-3}$) & $1.71 \times 10^{-3}$ ($\pm$ $1.35 \times 10^{-3}$) \\
Principal agent ($a=0.5$) & MLP & $3.28 \times 10^{-4}$ ($\pm$ $2.04 \times 10^{-4}$) & $2.66 \times 10^{-2}$ ($\pm$ $1.09 \times 10^{-2}$) & 0.00 ($\pm$ 0.00) \\
    & KAN & $2.23 \times 10^{-4}$ ($\pm$ $2.14 \times 10^{-5}$) & $2.07 \times 10^{-2}$ ($\pm$ $1.48 \times 10^{-3}$) & $1.39 \times 10^{-5}$ ($\pm$ $9.92 \times 10^{-7}$) \\
    & KAN Symbolic & $2.30 \times 10^{-4}$ ($\pm$ $2.10 \times 10^{-5}$) & $2.16 \times 10^{-2}$ ($\pm$ $1.57 \times 10^{-3}$) & $9.27 \times 10^{-4}$ ($\pm$ $2.99 \times 10^{-4}$) \\
    & KAN-KAN Symbolic & $2.35 \times 10^{-6}$ ($\pm$ $1.38 \times 10^{-6}$) & $5.79 \times 10^{-3}$ ($\pm$ $1.35 \times 10^{-3}$) & $9.27 \times 10^{-4}$ ($\pm$ $2.99 \times 10^{-4}$) \\
Principal agent ($a=1$) & MLP & $1.12 \times 10^{-3}$ ($\pm$ $1.10 \times 10^{-3}$) & $4.18 \times 10^{-2}$ ($\pm$ $2.02 \times 10^{-2}$) & 0.00 ($\pm$ 0.00) \\
    & KAN & $1.01 \times 10^{-4}$ ($\pm$ $3.13 \times 10^{-5}$) & $1.24 \times 10^{-2}$ ($\pm$ $1.67 \times 10^{-3}$) & $2.81 \times 10^{-3}$ ($\pm$ $4.00 \times 10^{-4}$) \\
    & KAN Symbolic & $9.95 \times 10^{-5}$ ($\pm$ $3.07 \times 10^{-5}$) & $1.19 \times 10^{-2}$ ($\pm$ $1.64 \times 10^{-3}$) & $4.06 \times 10^{-4}$ ($\pm$ $2.14 \times 10^{-4}$) \\
    & KAN-KAN Symbolic & $9.46 \times 10^{-7}$ ($\pm$ $1.22 \times 10^{-7}$) & $4.23 \times 10^{-3}$ ($\pm$ $5.08 \times 10^{-4}$) & $4.06 \times 10^{-4}$ ($\pm$ $2.14 \times 10^{-4}$) \\
Principal agent ($a=2$) & MLP & $4.76 \times 10^{-3}$ ($\pm$ $4.42 \times 10^{-3}$) & $7.75 \times 10^{-2}$ ($\pm$ $2.77 \times 10^{-2}$) & 0.00 ($\pm$ 0.00) \\
    & KAN & $4.53 \times 10^{-5}$ ($\pm$ $3.64 \times 10^{-5}$) & $1.08 \times 10^{-2}$ ($\pm$ $2.59 \times 10^{-3}$) & $9.53 \times 10^{-3}$ ($\pm$ $1.49 \times 10^{-3}$) \\
    & KAN Symbolic & $1.41 \times 10^{-3}$ ($\pm$ $4.28 \times 10^{-3}$) & $2.23 \times 10^{-2}$ ($\pm$ $3.74 \times 10^{-2}$) & $5.46 \times 10^{-3}$ ($\pm$ $2.23 \times 10^{-3}$) \\
    & KAN-KAN Symbolic & $1.53 \times 10^{-3}$ ($\pm$ $4.84 \times 10^{-3}$) & $1.71 \times 10^{-2}$ ($\pm$ $4.22 \times 10^{-2}$) & $5.46 \times 10^{-3}$ ($\pm$ $2.23 \times 10^{-3}$) \\
\bottomrule
\end{tabular}}
\end{table}

\paragraph{Black-Scholes (American option)} The American option setup is similar to the basic European option. An additional dividend yield parameter $q$ is added so that the free boundary is triggered. The goal is to solve for $V(S, t)$ with the following variational problem:
\begin{align*}
    &\min\bs{ \frac{\partial V}{\partial t} + (r-q) S \frac{\partial V}{\partial S} + \frac{1}{2}\sigma^2 S^2 \frac{\partial^2 V}{\partial^2 S} - rV, V - \max\bs{S-K, 0}} = 0\\
    \text{s.t. }& V(S, T) = \max\bs{S-K, 0}\\
    & V(0, t) = 0, \; V(1, t) = 1 - K
\end{align*}
The asset price $S$ and time $t$ are normalized to $[0,1]$. The constants are $K=0.3$, $r=0.03$, $q=0.08$, $\sigma=0.4$. The errors are computed over $t=0$.

\paragraph{Brunnermeier \& Sannikov} This model is based on Proposition 4 in \cite{Brunnermeier2014} and is implemented in PyMacroFin, starting from an initial guess. 
Following the setup in \cite{pymacrofinsolutionmethod}, there are two endogenous variables, $q$ and $\psi$. We use a quadratic investment function $\iota_t = \frac{q_t^2 - 1}{2\kappa}$, where $\kappa$ is the investment rate. The constant parameters are provided in Table~\ref{tab:free-boundary-model-constant-params}. The market clearing conditions and optimal portfolio choices lead to the following system of PDEs to solve when $\psi < 1$:
\begin{align*}
    (r(1-\eta) + \rho\eta)q &= \psi a_e + (1-\psi) a_h - \iota\\
    \sigma_t^q + \sigma &= \frac{\sigma}{1 - \frac{1}{q} \frac{\partial q}{\partial \eta} (\psi - \eta)}\\
    (\sigma + \sigma_t^q) ^2 \frac{q (\psi - \eta)}{\eta (1-\eta)} &= a_e - a_h
\end{align*}
When $\psi = 1$, the system of PDEs simplifies to: 
\begin{align*}
    (r(1-\eta) + \rho \eta) q &= a_e - \iota\\
    \sigma_t^q + \sigma &= \frac{\sigma}{1 - \frac{1}{q} \frac{\partial q}{\partial \eta} (1 - \eta)}
\end{align*}
The functions $q$ and $\psi$ are approximated using neural networks. During training, $\sigma_t^q$ is defined using $\sigma_t^q = \frac{\sigma}{1 - \frac{1}{q} \frac{\partial q}{\partial \eta} (\psi - \eta)}- \sigma$. Sum-of-squared errors are used for training, with a weight of 2 assigned to the endogenous equation involving $(\sigma+\sigma_t^q)^2$ for better convergence. The KAN models are 1-input and 1-output with no hidden dimensions so that it is easier to understand and demonstrate the symbolic formula.

\begin{table}[!ht]
\caption{Free boundary model constant parameters}\label{tab:free-boundary-model-constant-params}

\centering
\begin{tabular}{lll}
\toprule
Parameter & Definition & Value\\
\midrule
$\sigma$ & exogenous volatility of capital & 0.1\\
$\delta_e$ & depreciation rate of capital for experts & 0.05\\
$\delta_h$ & depreciation rate of capital for households & 0.05\\
$a$ & productivity of experts & 0.11\\
$a_h$ & productivity of households & 0.07\\
$\rho$ & discount rate of experts & 0.06\\
$r$ & discount rate of households & 0.05\\
$\kappa$ & adjustment cost parameter & 2\\
\bottomrule
\end{tabular}
\end{table}

\paragraph{Model setup}
Table~\ref{tab:free-boundary-model-setup} shows the model setup for all free boundary models.

\begin{table}[!htb]
\caption{Free boundary model setup}\label{tab:free-boundary-model-setup}

\centering
\resizebox*{\textwidth}{!}{%
\begin{tabular}{cccccccc}
\toprule
PDE & Model & Hidden & \#Params & Activation & Epochs & Optimizer & Learning rate \\
\midrule
Obstacle 1D & MLP & [50]*4 & 7801 & SiLU & 3000 & Adam & $10^{-3}$\\
 & KAN & [1] & 38 & SiLU & 100 & L-BFGS & $10^{-3}$\\
Obstacle 2D & MLP & [50]*4 & 7851 & SiLU & 3000 & Adam & $10^{-3}$\\
 & KAN & [1] & 56 & SiLU & 100 & L-BFGS & $10^{-3}$\\
Principle agent 1D & MLP & [50]*4 & 7801 & SiLU & 50000 & Adam & $10^{-3}$\\
 & KAN & [2] & 75 & SiLU & 200 & L-BFGS & $1$\\
Principle agent 2D & MLP & [50]*4 & 7851 & SiLU & 50000 & Adam & $10^{-3}$\\
Black-Scholes (American option) & MLP & [30]*4 & 2911 & SiLU & 10000 & Adam & $10^{-3}$\\
 & KAN & [1] & 56 & SiLU & 200 & L-BFGS & $1$\\
Brunnermeier \& Sannikov ($\psi < 1$) & MLP & [30]*4 & 2881 & SiLU & 10000 & Adam & $10^{-3}$\\
 & KAN & [] & 19 & SiLU & 100 & L-BFGS & $1$\\
Brunnermeier \& Sannikov ($\psi = 1$) & MLP & [30]*4 & 2881 & SiLU & 2000 & Adam & $10^{-3}$\\
 & KAN & [] & 19 & SiLU & 100 & L-BFGS & $1$\\
\bottomrule
\end{tabular}}
\end{table}

\subsection{Neoclassical growth}\label{appendix:ncg}

This appendix details the models and algorithms used in Section~\ref{sec:ncg}. The objective is to find the optimal consumption $c$ and the value function $V$ that solve the HJB equation:
\begin{align*}
    \rho V(k) &= \max_c u(c) + V'(k) (k^\alpha - \delta k -c),
\end{align*}
The utility function is given by 
\begin{align*}
    u(c)=\begin{cases*}
        \frac{c^{1-\gamma}}{1-\gamma}, \gamma\neq 1 (\text{CRRA})\\
        \log(c), \gamma=1 (\text{Log})
    \end{cases*}
\end{align*}
The first-order condition for optimality implies $u'(c)=V'(k)$. The steady state capital is at $k_{ss}=\bb{\frac{\alpha}{\rho+\delta}}^{\frac{1}{1-\alpha}}$. Parameter values are listed in Table~\ref{tab:neoclassical-growth-constant-parameters}. We parametrize $V$ and $c$ using neural networks and solve the following system of PDEs:
\begin{align*}
    c^{-\gamma} &= V'(k)\\
    \rho V(k) &=u(c) + V'(k) (k^\alpha - \delta k - c)\\
    V(k_{ss}) &= \begin{cases}
        \frac{(k_{ss}^\alpha-\delta k_{ss})^{1-\gamma}}{(1-\gamma)\rho}, \gamma\neq 1\\
        \frac{1}{\rho} \log(k_{ss}^\alpha-\delta k_{ss}), \gamma=1
    \end{cases}, \quad c(k_{ss}) = k_{ss}^\alpha-\delta k_{ss}
\end{align*} 
Our results are compared against the benchmark numerical scheme from \cite{moll2020}. Initially, we solve the system over the domain $[k_{ss},2k_{ss}]$ using the basic algorithm. However, this basic approach does not generalize well to a broader domain. To address this, we adopt a time-stepping algorithm similar to value function iteration methods in numerical dynamic programming. The procedure is outlined in Algorithm~\ref{algo:time-stepping}. For faster convergence, the initial boundary condition for $V$ is set to -18, and the initial guess for $c$ is set to 1.5. Table~\ref{tab:ncg-model-setup} shows the model setup.

\begin{table}[!htb]
\caption{Neoclassical growth model constant parameters}\label{tab:neoclassical-growth-constant-parameters}
\centering
\begin{tabular}{lll}
\toprule
Parameter & Definition & Value\\
\midrule
$\gamma$ & relative risk aversion& $\gamma=2$/$\gamma=1$ \\
$\rho$ & time preference & $\rho=0.05$\\
$\alpha$ & return to scale & $\alpha=0.3$\\
$\delta$ & capital depreciation & $\delta=0.05$\\
\bottomrule
\end{tabular}
\end{table}

\begin{table}[!htb]
\caption{Neoclassical growth model setup}\label{tab:ncg-model-setup}

\centering
\resizebox*{\textwidth}{!}{%
\begin{tabular}{cccccccc}
\toprule
PDE & Model & Hidden & \#Params & Activation & Epochs & Optimizer & Learning rate \\
\midrule
Basic ($V$) & MLP & [64]*4 & 12673 & tanh & 20000 & Adam & $10^{-3}$\\
 & KAN & [2,2] & 149 & SiLU & 200 & L-BFGS & 1\\
Basic ($c$) & MLP & [32]*4 & 3265 & tanh & 20000 & Adam & $10^{-3}$\\
 & KAN & [2] & 75 & SiLU & 200 & L-BFGS & 1\\
Timestepping ($V$) & MLP & [64]*4 & 12737 & tanh & 20$\times$3000 & Adam & $10^{-3}$\\
Timestepping ($c$) & MLP & [32]*4 & 3297 & tanh & 20$\times$3000 & Adam & $10^{-3}$\\
\bottomrule
\end{tabular}}
\end{table}

\subsubsection{A high-dimensional multi-location dynamic capital allocation model}
Following \cite{dp-dynamic}, we analyze a multi-location capital allocation model. Time is continuous and there are $L$ locations. Each location $l$ has a representative household with CRRA utility $u(c) = \frac{c^{1-\gamma}}{1-\gamma}$, and an idiosyncratic and exogenous productivity $z_l$. The social planner solves the following optimization problem:
\begin{align*}
    \rho V(k, z) &= \max_{c, i, x} \sum_{l=1}^L u(c_l) + \nabla_k V^T (i-\delta k+x) + \nabla_z V^T \mu_z + \frac{1}{2} \sigma_z^T H_z(V) \sigma_z\\
    \text{s.t. } \quad dz_l &= \bb{-\nu\log (z_l) + \frac{\sigma^2}{2}} z_l dt + \sigma z_l dW_t^l \\
    zk_l^\alpha &= c_l + i_l + 0.5\kappa_1 i_l^2 + 0.5\kappa_2x_l^2\\
    \sum_{l=1}^L x_l &= 0
\end{align*}

\begin{table}[!htb]
\caption{Capital allocation model constant parameters}\label{tab:constant-params-multiloc}
\centering
\begin{tabular}{lll}
\toprule
Parameter & Definition & Value\\
\midrule
$\gamma$ & relative risk aversion& $\gamma=2$ \\
$\rho$ & time preference & $\rho=0.04$\\
$\alpha$ & return to scale & $\alpha=0.33$\\
$\delta$ & capital depreciation & $\delta=0.07$\\
$\kappa_1,\kappa_2$ & adjustment cost & $\kappa_1=\kappa_2=1$\\
$\nu$ & autocorrelation of productivity & $e^{-\nu}=0.8$\\
$\sigma$ & variance of productivity & $\sigma=0.33$\\
\bottomrule
\end{tabular}
\end{table}

The constant parameters are given in Table~\ref{tab:constant-params-multiloc}. Define neural networks: $V:\RR^{2L}\to\RR^1$ (value function), $I:\RR^{2L}\to \RR^L$ (investments), $X:\RR^{2L} \to \RR^{L-1}$ (capital share). Note that the last dimension of capital share can be computed by $X_{L}=-\sum_{i=1}^{L-1}X_i$. Sample $z\in [0.5,2.5]^L$, $k\in [2.5, 5.5]^L$. Compute:
\begin{align*}
    \mu_z &= \bb{-\nu \log z + \frac{\sigma^2}{2}}z\\
    \sigma_z &= \sigma z\\
    C &= zk^\alpha - I - 0.5 \kappa_1 I^2 - 0.5 \kappa_2 X^2\\
    U &= \sum_{l=1}^L \frac{C_l^{1-\gamma}}{1-\gamma}
\end{align*}

Taking FOC on $I$ and $X$ gives:
\begin{align*}
    \forall l\in \bs{0,1,...,L},  C_l^{-\gamma}*\bb{1+\kappa_1 I_l} - \frac{\partial V}{\partial k_l} &= 0\\
    \forall l\in \bs{0,1,...,L-1}, C_l^{-\gamma}\bb{\kappa_2 X_l} - \frac{\partial V}{\partial k_l} &= C_L^{-\gamma}\bb{\kappa_2 X_L} - \frac{\partial V}{\partial k_L}
\end{align*}
Together with market clearing and HJB, we get a system of equation to solve:

\begin{align*}
    C^{-\gamma}\bb{1+\kappa_1 I} - \nabla_k V &= 0\\
    \forall l\in \bs{0,1,...,L-1}, C_l^{-\gamma}\bb{\kappa_2 X_l} - \frac{\partial V}{\partial k_l} &= C_L^{-\gamma}\bb{\kappa_2 X_L} - \frac{\partial V}{\partial k_L}\\
    \partial_t V + U + \nabla_k V^T (I-\delta k + X) + \nabla_z V^T \mu_z + \frac{1}{2}\sigma_z^T H(V) \sigma(z) &= \rho V
\end{align*}

\begin{figure}[!htb]
\centering
\begin{subfigure}[b]{0.24\linewidth}
\centering
\includegraphics[width=\linewidth]{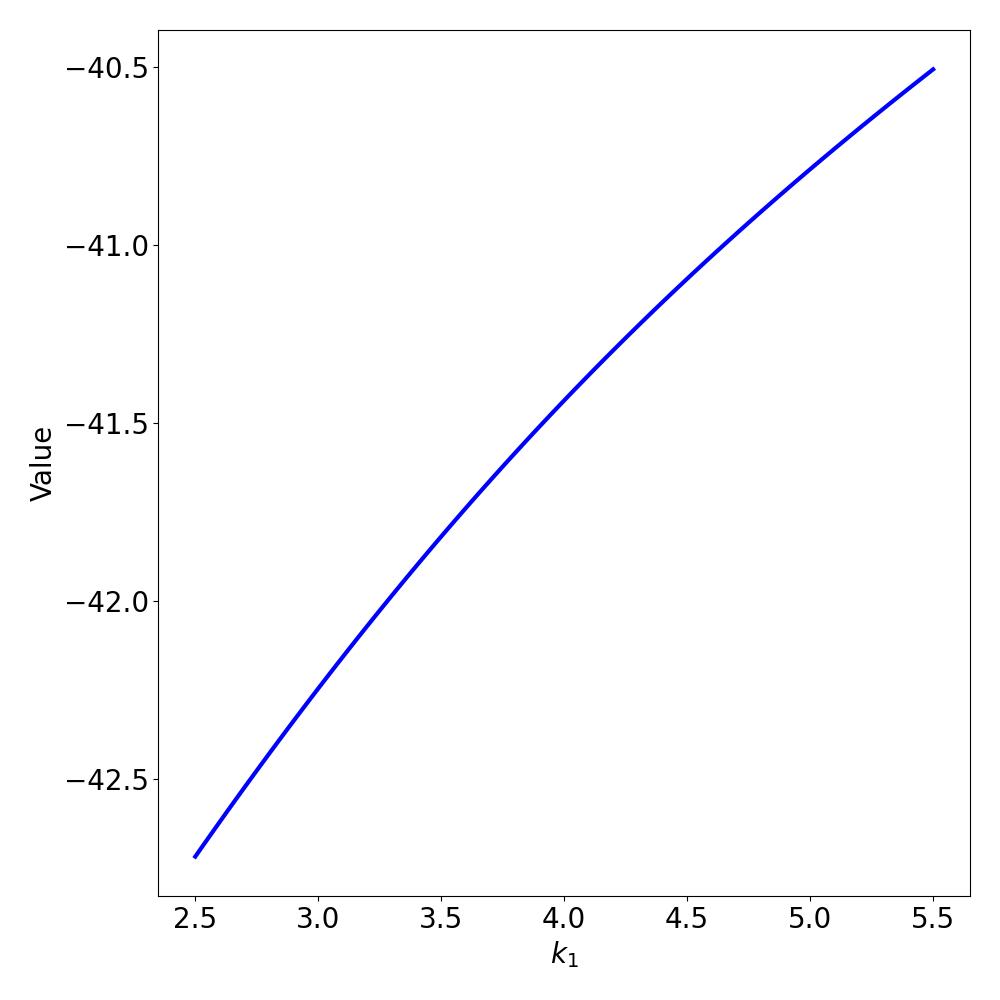}
\caption{Value function}
\end{subfigure}
\begin{subfigure}[b]{0.24\linewidth}
\centering
\includegraphics[width=\linewidth]{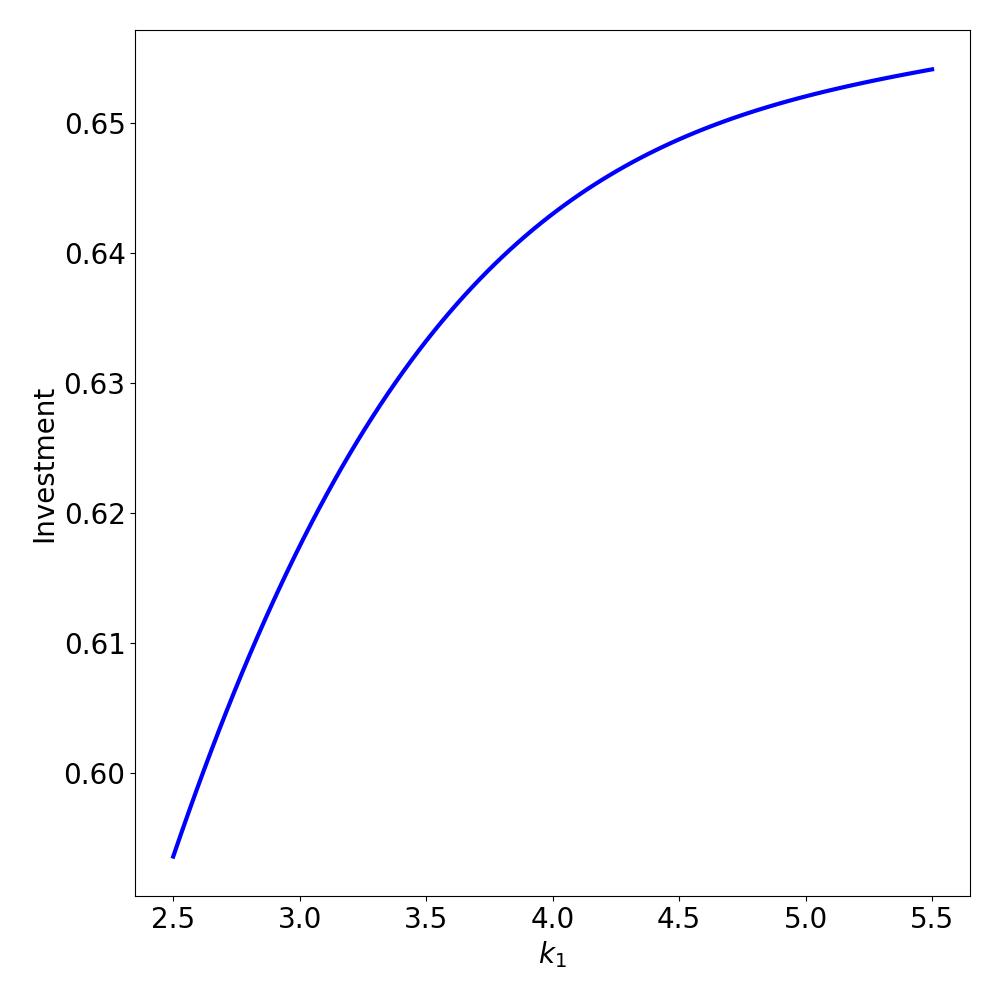}
\caption{Investment function}
\end{subfigure}
\begin{subfigure}[b]{0.24\linewidth}
\centering
\includegraphics[width=\linewidth]{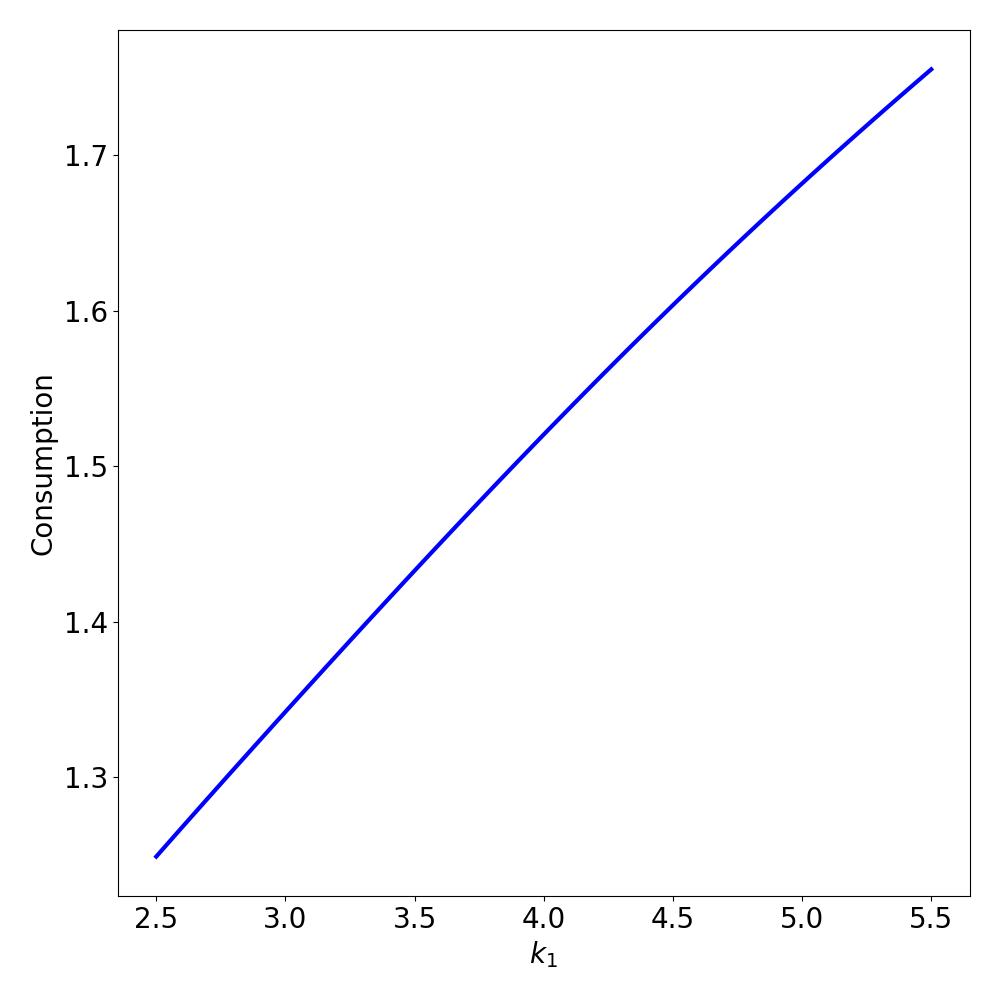}
\caption{Consumption function}
\end{subfigure}
\begin{subfigure}[b]{0.24\linewidth}
\centering
\includegraphics[width=\linewidth]{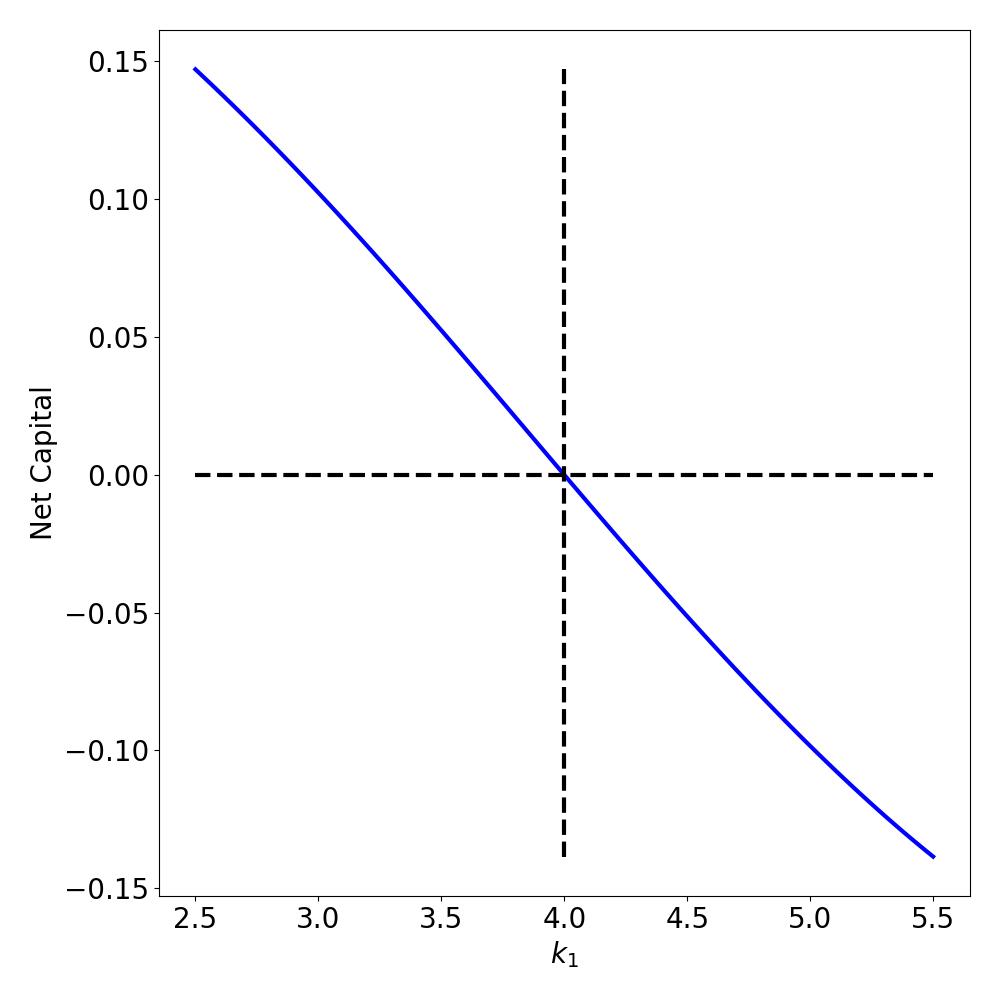}
\caption{Capital share}
\end{subfigure}

\caption{2-location capital allocation (HJB loss $2.47\times 10^{-5}$)}
\label{fig:ncg-4d}
\end{figure}

\begin{figure}[!htb]
\centering
\begin{subfigure}[b]{0.24\linewidth}
\centering
\includegraphics[width=\linewidth]{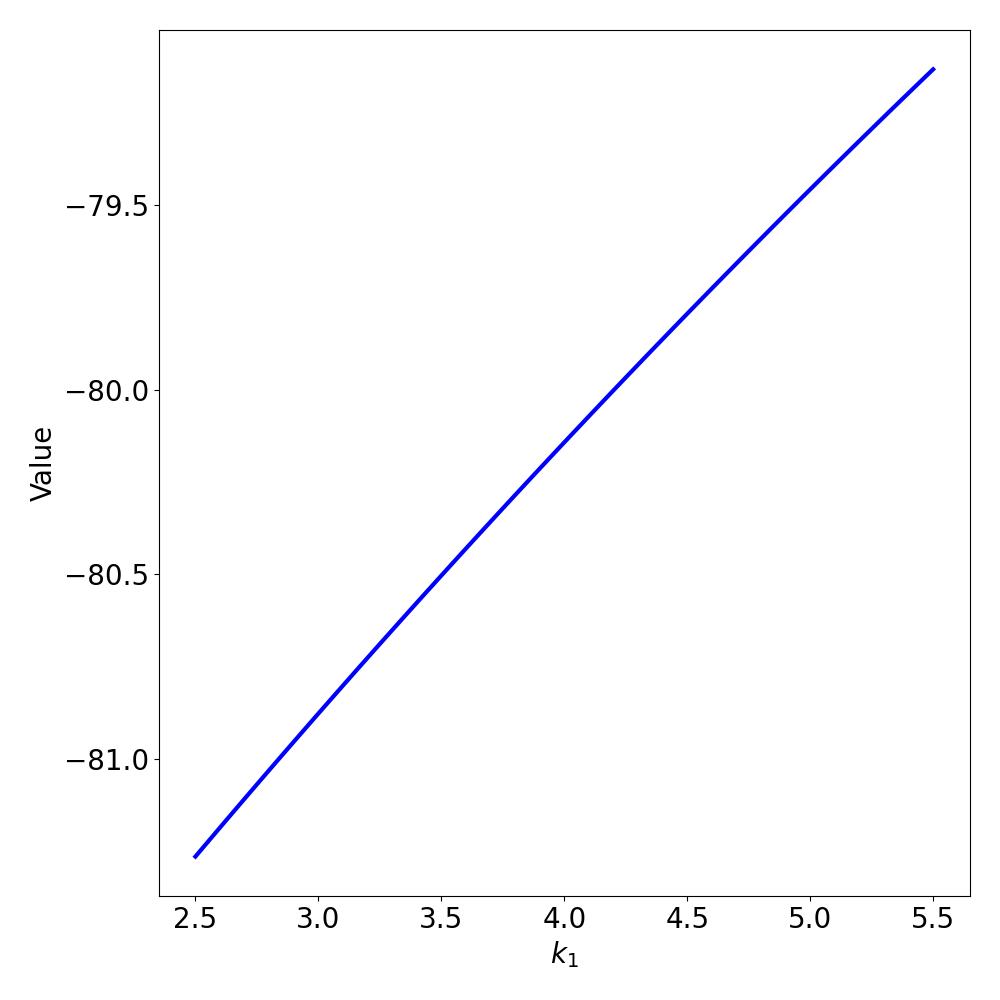}
\caption{Value function}
\end{subfigure}
\begin{subfigure}[b]{0.24\linewidth}
\centering
\includegraphics[width=\linewidth]{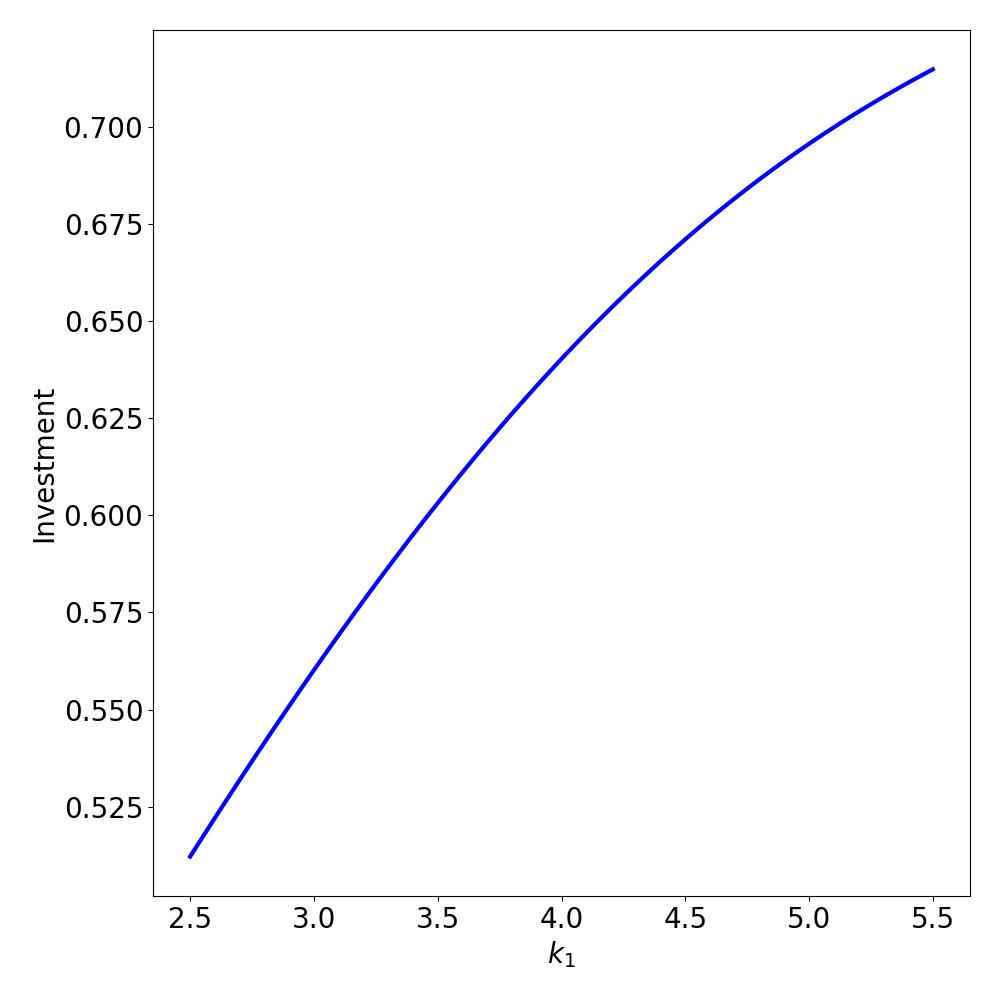}
\caption{Investment function}
\end{subfigure}
\begin{subfigure}[b]{0.24\linewidth}
\centering
\includegraphics[width=\linewidth]{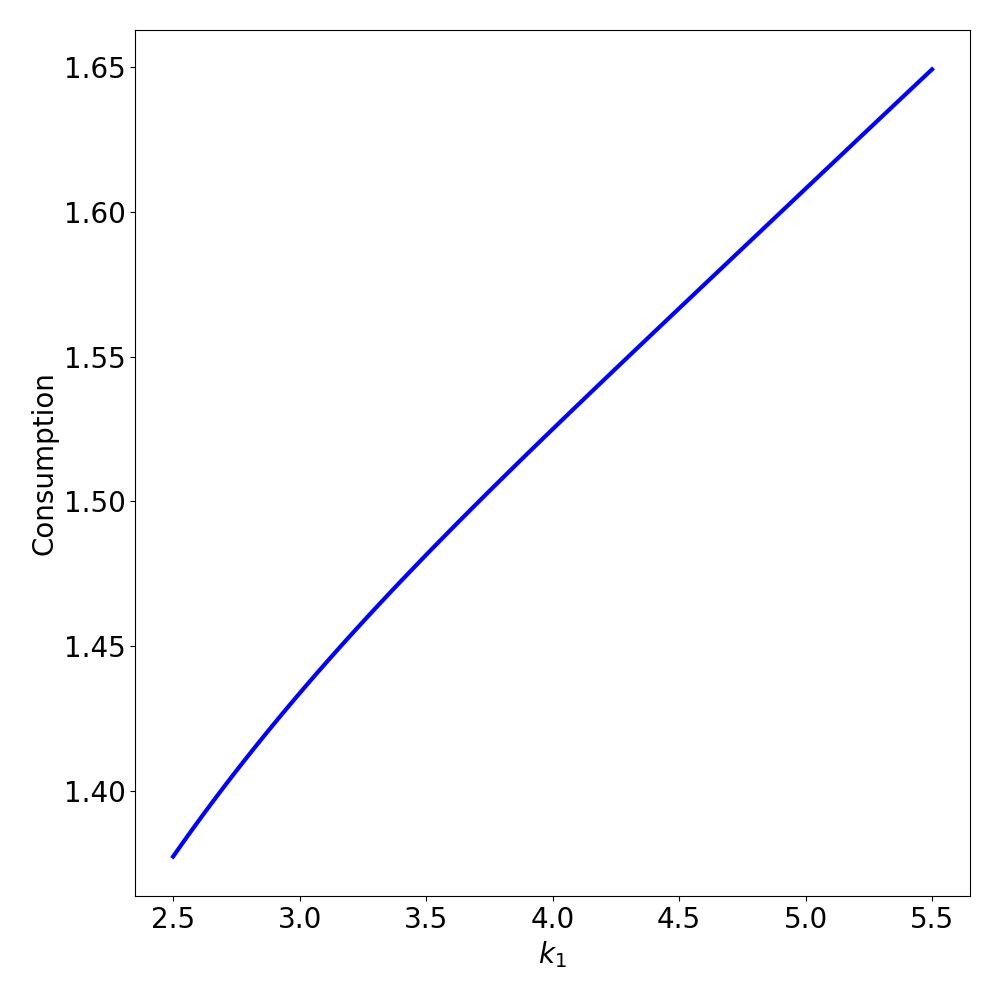}
\caption{Consumption function}
\end{subfigure}
\begin{subfigure}[b]{0.24\linewidth}
\centering
\includegraphics[width=\linewidth]{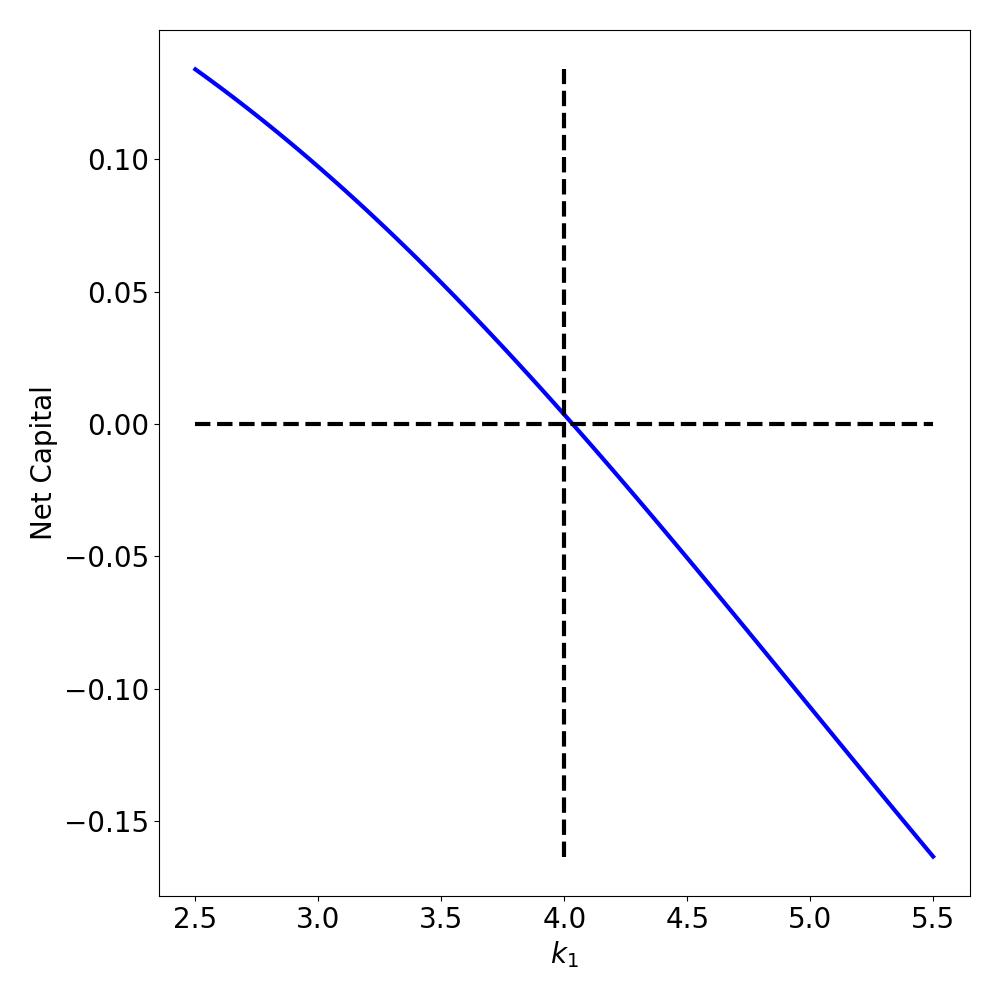}
\caption{Capital share}
\end{subfigure}

\caption{5-location capital allocation (HJB loss $2.63\times 10^{-4}$)}
\label{fig:ncg-10d}
\end{figure}

\begin{figure}[!htb]
\centering
\begin{subfigure}[b]{0.24\linewidth}
\centering
\includegraphics[width=\linewidth]{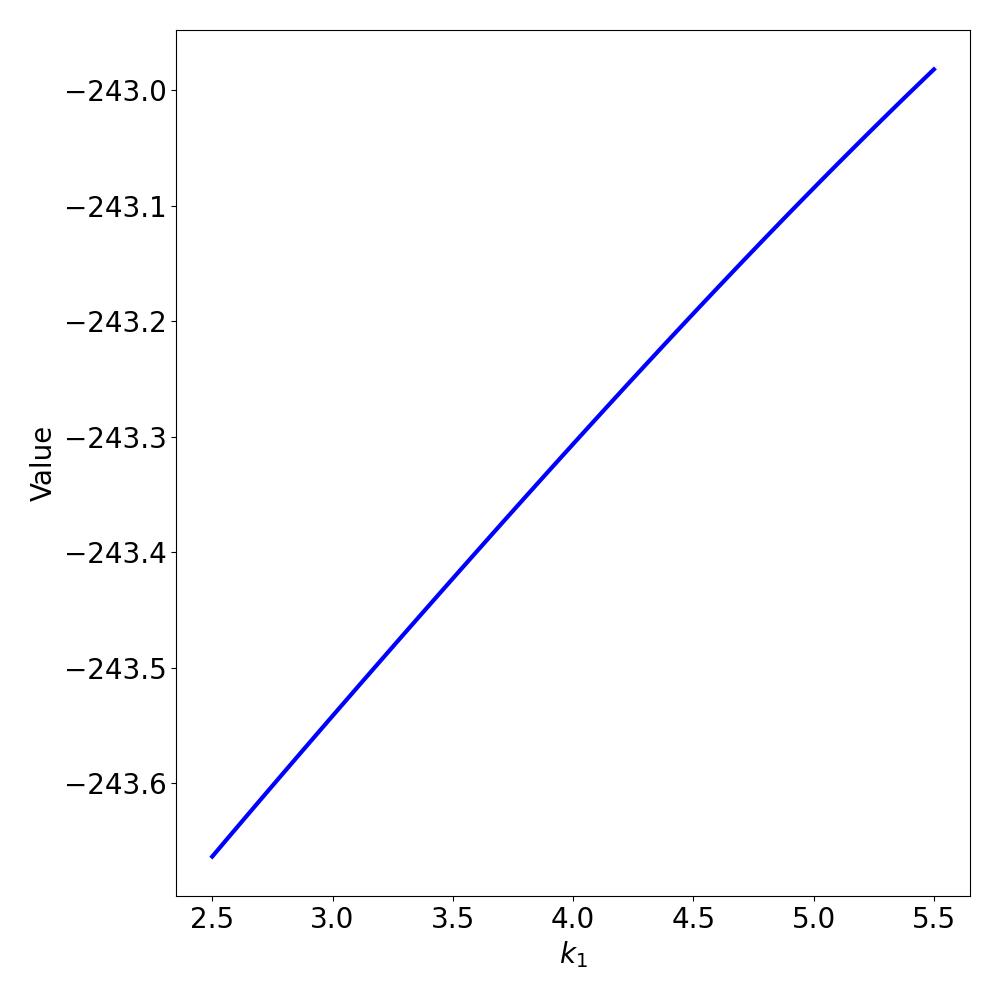}
\caption{Value function}
\end{subfigure}
\begin{subfigure}[b]{0.24\linewidth}
\centering
\includegraphics[width=\linewidth]{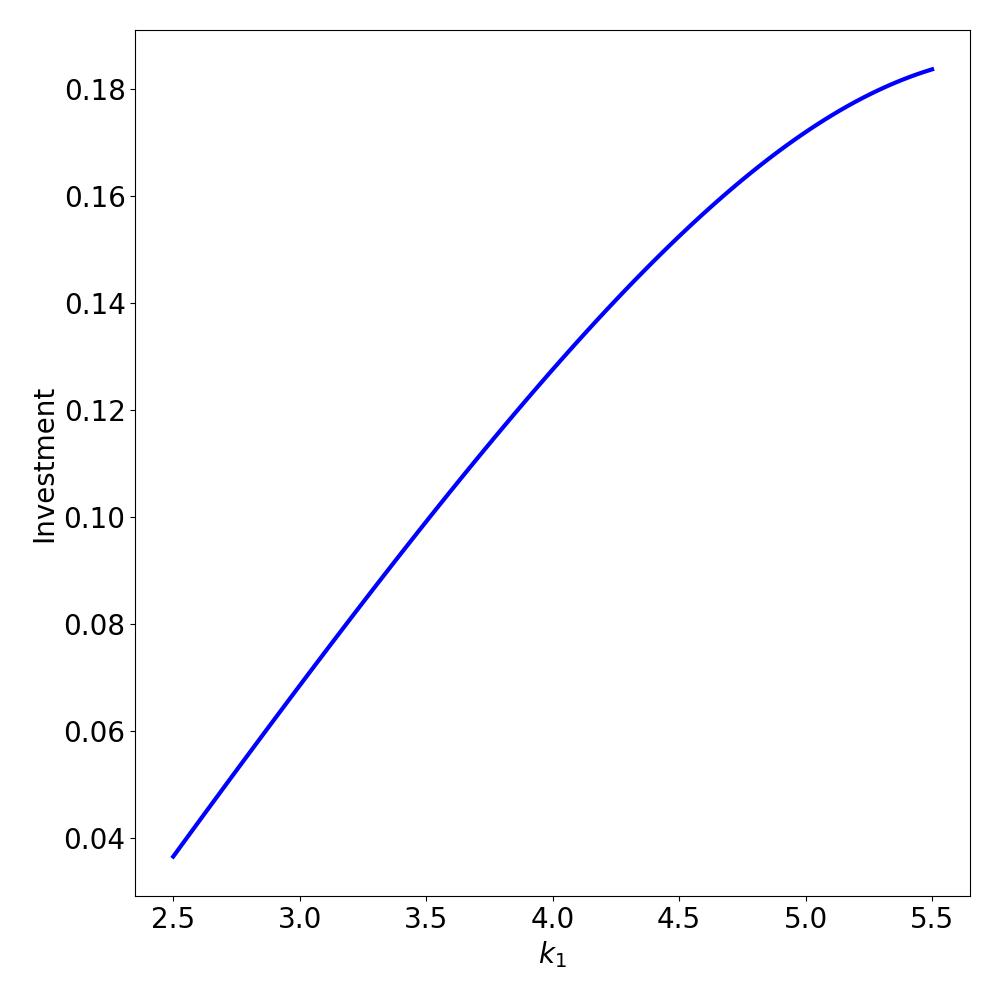}
\caption{Investment function}
\end{subfigure}
\begin{subfigure}[b]{0.24\linewidth}
\centering
\includegraphics[width=\linewidth]{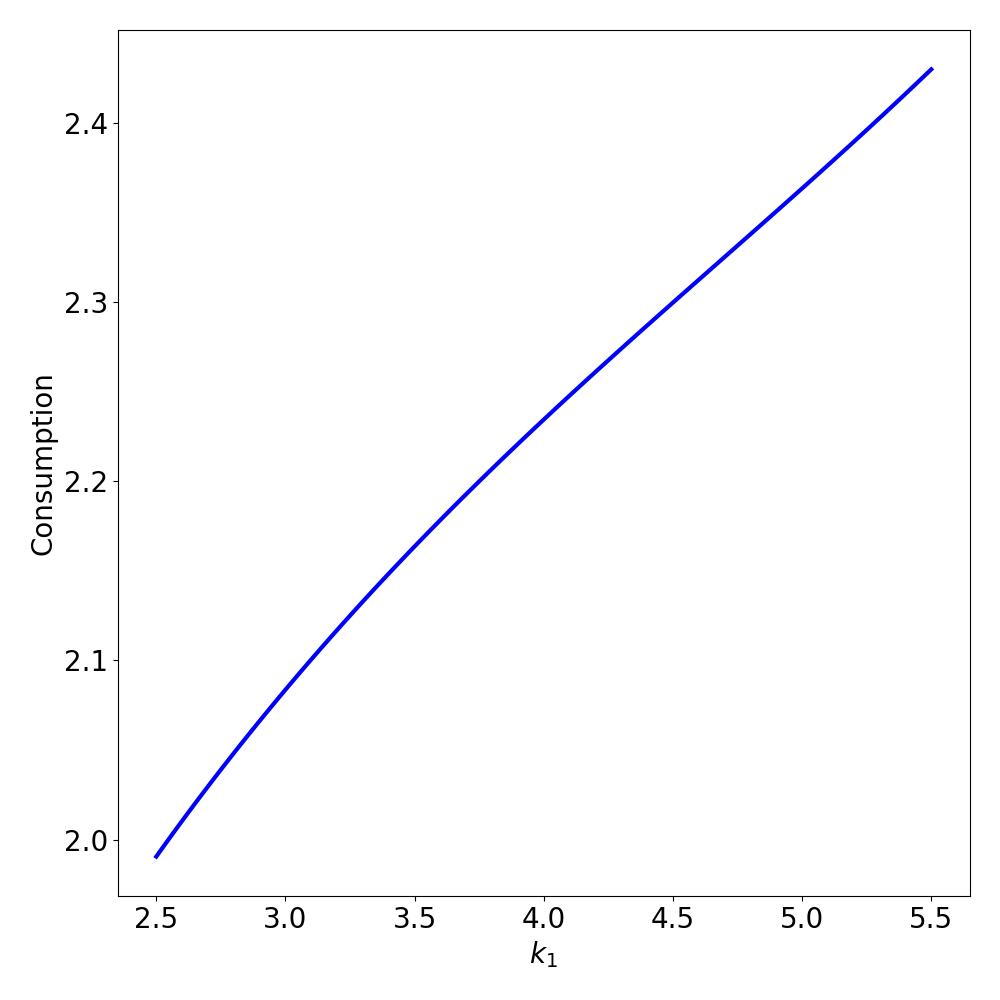}
\caption{Consumption function}
\end{subfigure}
\begin{subfigure}[b]{0.24\linewidth}
\centering
\includegraphics[width=\linewidth]{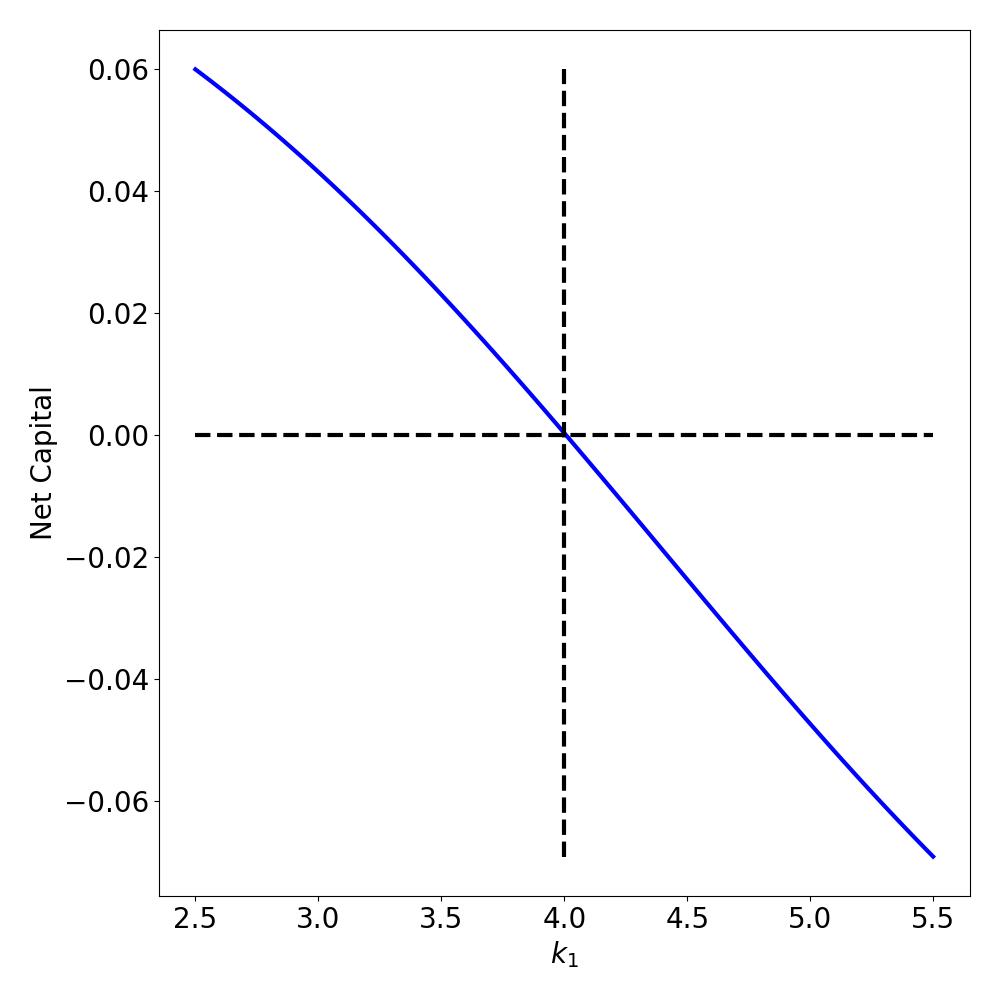}
\caption{Capital share}
\end{subfigure}

\caption{25-location capital allocation (HJB loss $8.19\times 10^{-5}$)}
\label{fig:ncg-20d}
\end{figure}

We train the model for $L=2, 5, 25$ (4, 10, 50-dimension). Table~\ref{tab:multi-loc-model-setup} shows the model setup. The difference in the number of parameters is due to different output sizes. For 2- and 5-location models, we solve for the global solution, while for the 25-location model, we solve for the location solution for $k_1\in [2.5, 5.5]$, $z_1\in [0.5, 2.5]$, with $k_i=4, z_i=1.5$ for $i\in \bs{2,...,25}$. The 25-location model achieves early termination at the 36th timestep, with difference in error $<10^{-4}$.
Figure~\ref{fig:ncg-4d}, \ref{fig:ncg-10d} and \ref{fig:ncg-20d} show partial results for $L=2, 5, 25$ respectively, displaying the value function, investment function, consumption function and net capital for Location 1. For visualization purposes, all non-displayed dimensions are fixed at their domain midpoints ($k_i=4$ and $z_i=1.5$). 
Value, investment and consumption functions have concave shapes with decaying growth rate as $k$ increases, consistent with diminishing returns to capital accumulation. Net capital is decreasing and crosses 0 at $k_1=4$, since capital in all other locations is fixed at 4, location 1 imports capital for $k<4$ and exports capital for $k>4$. Note that in the original formulation in \cite{dp-dynamic}, the FOCs misses a factor of 2, or equivalently, the adjustment cost term $\kappa_1 i_l^2 + \kappa_2 x_l^2$ is missing a factor of 0.5. As a result, there is a discrepancy between our solution and theirs.

\begin{table}[!htb]
\caption{Multi-location allocation model setup}\label{tab:multi-loc-model-setup}

\centering
\resizebox*{\textwidth}{!}{%
\begin{tabular}{cccccccc}
\toprule
$L$ & Variable & Hidden & \#Params & Activation & Epochs & Optimizer & Learning rate \\
\midrule
2 & $V$, $I$, $X$ & [64]*3, [30]*3, [30]*3 & 8769, 2102, 2071 & tanh & $20\times 10000$ & Adam & $10^{-3}$\\
5 & $V$, $I$, $X$ & [64]*3 for all & 9153, 9413, 9348 & tanh & $50\times 10000$ & Adam & $10^{-3}$\\
25 & $V$, $I$, $X$ & [128]*3 for all & 39809, 42905, 42776 & tanh & $50\times 10000$ & Adam & $10^{-3}$\\
\bottomrule
\end{tabular}}
\end{table}

\subsection{Lucas orchard}\label{appendix:trees}

This appendix provides the details for models and algorithms used in Section~\ref{sec:lucas-orchard}. Consider an economy with $N$ risky assets, each following a Lucas tree process. The dividend growth of each asset is assumed to be independent and identically distributed (i.i.d.) over time but potentially correlated across assets.
An agent can hold multiple assets, but total wealth is constrained to equal 1. Therefore, we parametrize the agent’s portfolio using $N-1$ variables $z:=(z_1,z_2,...,z_{N-1})\in \RR^{N-1}$, where $z_i$ denotes the share of wealth invested in asset $i$. The value function $\kappa: \RR^{N-1}\to\RR^N$ represents the value associated with each tree. The dynamics of $z$ are governed by a geometric Brownian motion:
\begin{align*}
    dz_t = \mu^z zdt + \sigma^z zdW_t,
\end{align*}
where the components of $z$ are normalized by the dividends $y_i$ of each tree, which themselves follow geometric Brownian motions with constant drift $\mu^y$ and volatility $\sigma^y$. In our experiments, we set $\mu^{y_i}=\sigma^{y_i} = 0.01 i$ for all models for illustration purpose. However, these parameters can be tunned to match the real-world scenarios.
Each tree has a price $q_j$ determined by the value function $\kappa_j$ and the portfolio share $z_j$. The agent chooses asset weights $w_j$ and consumption $c_t$, maximizing expected lifetime utility under a CRRA utility function with risk aversion parameter $\gamma=5$ and discount rate $\rho=0.05$:
\begin{align*}
    \max_{c_t, w_t} \EE \closedb{\int_0^\infty e^{-\rho t} \frac{c_t^{1-\gamma}}{1-\gamma}dt}
\end{align*}

We define the full wealth share vector $\overline{z} = \bb{z_1,...,z_{N-1},1-\sum_{i=1}^z z_i}$, and denote by $\mu^{\overline{z}}$ and $\sigma^{\overline{z}}$ the corresponding drift and diffusion coefficients. We then compute the following variables in the equilibrium:
{\allowdisplaybreaks
\begin{align*}
    q &= \frac{\overline{z}}{\kappa}\\
    \mu^{z_i} &= \mu^{y_i} - \mu^y\cdot \overline{z} + \sigma^y\cdot \overline{z}  (\sigma^y\cdot \overline{z} - \sigma^{y_i})\\
    \sigma^{z_i} &= \sigma^{y_i} - \sigma^y\cdot \overline{z}\\
    \mu_z &= \mu^{z}  \times z\\
    \sigma_z &= \sigma^{z}  \times z\\
    \mu_{z_N} &= -\sum_{i=1}^{N-1}\mu_{z_i}\\
    \sigma_{z_N} &= -\sum_{i=1}^{N-1}\sigma_{z_i}\\
    \mu^{z_N} &= \frac{\mu_{z_N}}{z_N}\\
    \sigma^{z_N} &= \frac{\sigma_{z_N}}{z_N}\\
    \mu^{q} &= \frac{1}{q} \times  \left(\nabla_{z} q \mu_{z} + \frac{1}{2} (\sigma_z)^T H_z(q) \sigma_z\right)\\
    \sigma^{q} &= \frac{1}{q} \times  \left(\nabla_{z} q  \sigma_z\right)\\
    r &= \rho + \gamma  \mu^{y} \cdot \overline{z} - \frac{1}{2}  \gamma  (\gamma + 1)  (\overline{z})^2 \cdot (\sigma^{y})^2\\
    \zeta &= \gamma  \overline{z}\cdot \sigma^{y}\\
    \mu^{\kappa} &= \mu^{\overline{z}} - \mu^{q} + \sigma^{q} \times (\sigma^{q} - \overline{z})\\
    \sigma^{\kappa} &= \sigma^{\overline{z}} - \sigma^{q}
\end{align*}
}
Then we solve for $\kappa$ with the following HJB equations:
\begin{align*}
    \mu^{\kappa}\kappa=&\partial_t \kappa + \nabla_{z} \kappa \mu_z + \frac{1}{2} \sigma_z^T H_z (\kappa) \sigma_z\\
    \sigma^{\kappa}\kappa=&\nabla_{z} \kappa  \sigma_z,
\end{align*}

\begin{figure}[!htb]
\centering
\begin{subfigure}[b]{0.3\linewidth}
\centering
\includegraphics[width=\linewidth]{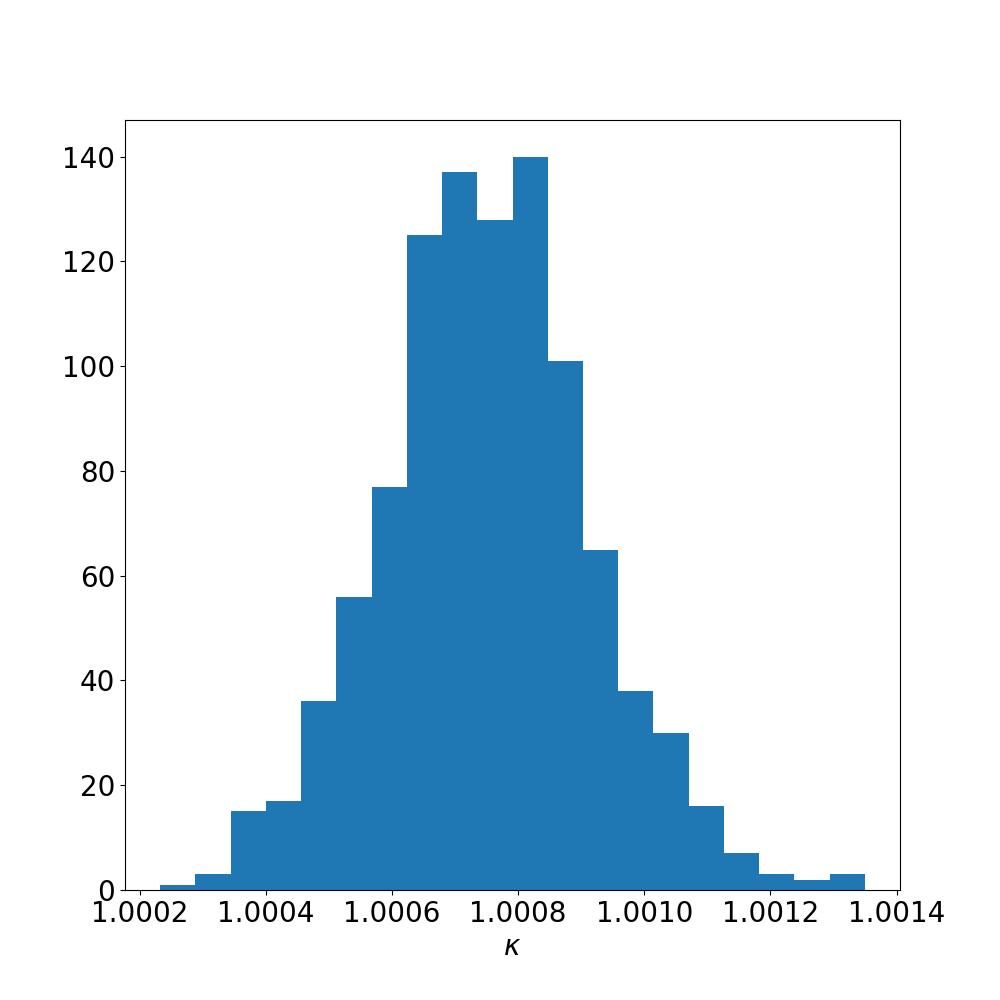}
\caption{$\kappa$}
\end{subfigure}
\begin{subfigure}[b]{0.3\linewidth}
\centering
\includegraphics[width=\linewidth]{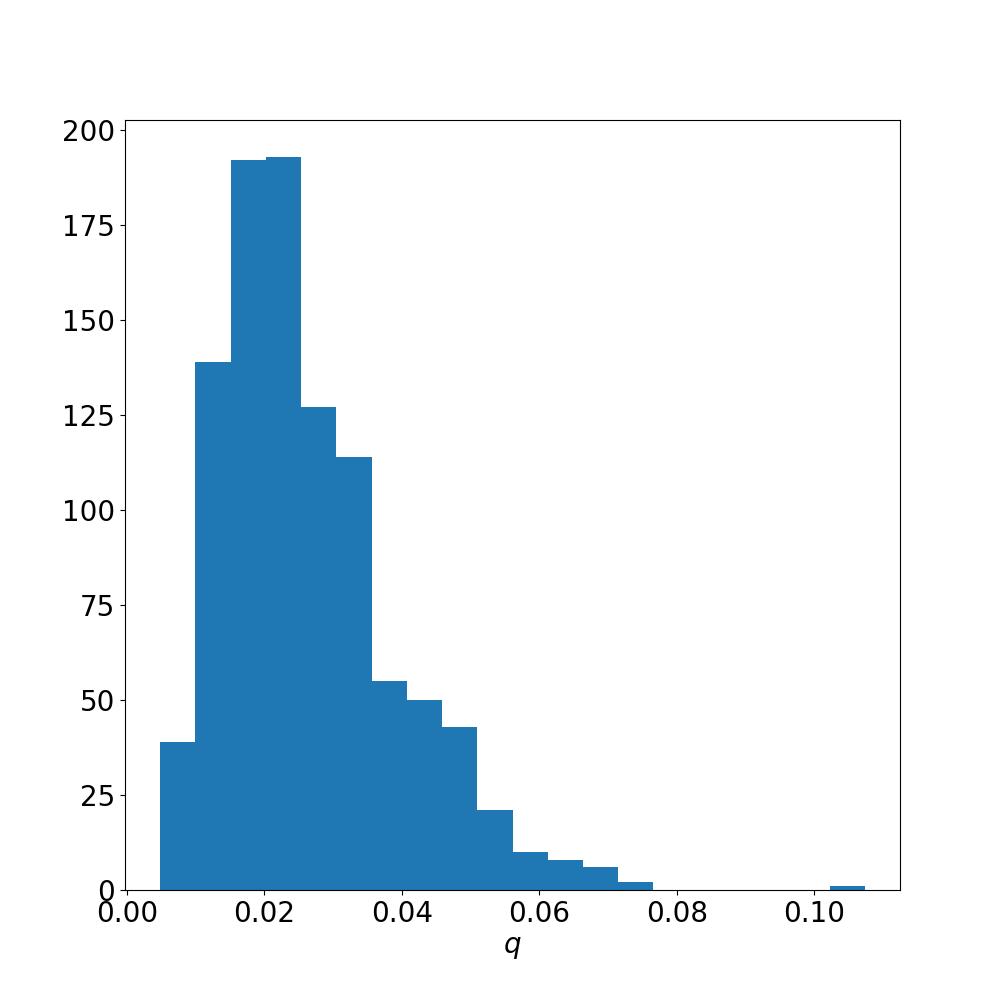}
\caption{$q$}
\end{subfigure}

\caption{50-Tree Distribution}
\label{fig:50-tree}
\end{figure}

During training, we sample from the ergodic distributions of dividends, with 
\begin{equation*}
    y_i\sim \text{LogNormal}(\mu^{y_i},\sigma^{y_i})
\end{equation*}
and then normalize to obtain the portfolio shares $z_i$. $\kappa$ is parametrized by 4-layer MLP with 80 hidden neurons in each layer. We use Adam optimizer with learning rate $\alpha=5\times 10^{-4}$. Table~\ref{tab:2-tree-validation} reports the errors between our solution and the PyMacroFin benchmark in the 1D case. The value function $\kappa$ is expected to be around 1, and since $q_i=\frac{z_i}{\kappa_i}$, under ergodic distribution, $q_i\approx \frac{1}{N}$ for $N$-tree models. Figure~\ref{fig:50-tree} shows the distribution of $\kappa$ and $q$ over the ergodic distribution, confirming the accuracy of our method.

\begin{table}[!htb]
\caption{2-Tree validation}\label{tab:2-tree-validation}
\centering
\resizebox*{\textwidth}{!}{%
\begin{tabular}{ccccc}
\toprule
    & $\kappa_1$ & $\kappa_2$ & $q_1$ & $q_2$ \\
\midrule
MSE & $3.93 \times 10^{-5}$ ($\pm$ $3.21 \times 10^{-5}$) & $3.96 \times 10^{-5}$ ($\pm$ $3.23 \times 10^{-5}$) & $1.82 \times 10^{-5}$ ($\pm$ $1.58 \times 10^{-5}$) & $1.33 \times 10^{-5}$ ($\pm$ $1.50 \times 10^{-5}$) \\
$\|\cdot\|_{L^\infty}$ & $1.16 \times 10^{-2}$ ($\pm$ $4.79 \times 10^{-3}$) & $1.16 \times 10^{-2}$ ($\pm$ $4.96 \times 10^{-3}$) & $9.47 \times 10^{-3}$ ($\pm$ $5.67 \times 10^{-3}$) & $8.80 \times 10^{-3}$ ($\pm$ $5.28 \times 10^{-3}$) \\
\bottomrule
\end{tabular}}
\end{table}

\subsection{One-Dimensional Economic Model}\label{appendix:1d-model}
This appendix provides the model details and results for another one dimensional economic model that numerical methods like PyMacroFin fails to solve due to singular matrix. Let $j\in\{h, e\}$ index two types of agents, where $h$ represents households and $i$ represents experts (intermediaries). We use $i$ here to avoid the confusion with $e$ for $\eta$ in the code. 

\subsubsection{Ito's Lemma Derivations}\label{sec:derivation}

Let $Z_t$ be a Brownian motion on $\bb{\Omega, \calF, \bs{\calF_t}_{t\geq 0}, \PP}$. The evolution of the capital is given by 
\begin{equation}
    \frac{dk_t}{k_t} = (\Phi(\iota_t) - \delta)dt + \sigma dZ_t,
\end{equation}
where $\delta$ and $\sigma$ represent depreciation rate and volatility of capital, respectively. $\Phi(\cdot)$ is a functional form for the investment function $\iota_t$.
Experts finance their holdings by short-selling the risk-free asset to households, assuming there are no transaction costs, and the agents are price-takers. The price of capital $q_t$, has dynamics:
\begin{equation}
    \frac{dq_t}{q_t} = \mu_t^q dt + \sigma_t^q dZ_t.
\end{equation}
Capital returns $dr_t^{kj}$ are based on dividend yield $\frac{a_j - \iota_t}{q_t} dt$ and capital gain $d (q_t k_t)$.
The aggregate amount of capital is $K_t$, and the economy's aggregate net worth is $q_t K_t$. Let the aggregate net worth of experts be $w_t^i$. Experts' wealth share, $\eta_t$, is $\eta_t = \frac{w_t^i}{q_t K_t} \in [0,1]$.

Given the dynamics of $k_t^a$ (capital),  $q_t^a$ (price of capital), $\xi_t^{j}$ (agent value function) and state variable $\eta_t$:
\begin{align}
    \frac{dk_t^a}{k_t^a} &= (\mu^a + \Phi(\iota_t^a))dt + \sigma^a dZ_t^a\\
    \frac{d q_t^a}{q_t^a} &= \mu_t^{qa} dt + \sigma_t^{qa} dZ_t^a,\\
    \frac{d \xi_t^j}{\xi_t^j} &= \mu_t^{\xi j} dt + \sigma_t^{\xi ja} dZ_t^a,\\
    \frac{d\eta_t}{\eta_t} &= \left( \underbrace{(1-\eta_t)(\mu_t^{n i} - \mu_t^{n h}) +(\sigma_t^{na})^2  - \sigma_t^{n ia}\sigma_t^{na}}_{\mu_t^{\eta}}) \right) dt + \underbrace{(1-\eta_t) ( \sigma^{nia}_t-  \sigma^{nha}_t)}_{\sigma_t^{\eta a}}dZ^a_t,
\end{align}
where $Z_t^a$ is a standard Brownian motion on $\bb{\Omega, \calF, \bs{\calF_t}_{t\geq 0}, \PP}$. $dZ_t^a$ is a Wiener process (with $\mu=0$, $\sigma=1$), $(dZ_t^a)^2 = dt$. $\iota_t^a$ is investment function of capital, and $\Phi(\iota_t^a)$ is a functional form for the investment function.

Rewrite the process of $\eta_t$ as $d \eta_t = \mu_t^{\eta} \eta_t dt + \sigma_t^{\eta a} \eta_t dZ_t^a$. Then
\begin{align*}
    (d \eta_t)^2 &= (\mu_t^{\eta} \eta_t dt + \sigma_t^{\eta a} \eta_t dZ_t^a)^2\\
    &= (\sigma_t^{\eta a} \eta_t)^2 dt + (\mu_t^{\eta} \eta_t)^2 (dt)^2 + 2 (\mu_t^{\eta} \eta_t)(\sigma_t^{\eta a} \eta_t) dt dZ_t^a \\
    &= (\sigma_t^{\eta a} \eta_t)^2 dt + o(dt),
\end{align*}
where $o(dt) = \bs{f : |f(\eta_t, t)| < \epsilon |dt|, \forall \epsilon > 0}$.

The price is a process dependent of the state variable $\eta_t$, $q_t^a = q_t^a(\eta_t)$. By Ito's Lemma,
\begin{align*}
    d(q_t^a (\eta_t)) &= \frac{\partial q_t^a}{\partial \eta_t} d \eta_t + \frac{1}{2} \frac{\partial^2 q_t^a}{\partial \eta_t^2} (d \eta_t)^2\\
    &= \frac{\partial q_t^a}{\partial \eta_t} (\mu_t^{\eta} \eta_t dt + \sigma_t^{\eta a} \eta_t dZ_t^a) + \frac{1}{2} \frac{\partial^2 q_t^a}{\partial \eta_t^2} (\sigma_t^{\eta a} \eta_t)^2 dt\\
    &= \bb{\frac{\partial q_t^a}{\partial \eta_t} \mu_t^{\eta} \eta_t + \frac{1}{2} \frac{\partial^2 q_t^a}{\partial \eta_t^2} (\sigma_t^{\eta a} \eta_t)^2} dt + \frac{\partial q_t^a}{\partial \eta_t} \sigma_t^{\eta a} \eta_t dZ_t^a
\end{align*}

Match the terms with $\frac{d q_t^a}{q_t^a} = \mu_t^{qa} dt + \sigma_t^{qa} dZ_t^a$,
\begin{align*}
    \mu_t^{qa} &= \frac{1}{q_t^a} \bb{\frac{\partial q_t^a}{\partial \eta_t} \mu_t^{\eta} \eta_t + \frac{1}{2} \frac{\partial^2 q_t^a}{\partial \eta_t^2} (\sigma_t^{\eta a} \eta_t)^2},\\
    \sigma_t^{qa} &= \frac{1}{q_t^a} \frac{\partial q_t^a}{\partial \eta_t} \sigma_t^{\eta a} \eta_t.
\end{align*}

Similarly, $\xi_t^j$ is a process dependent of $\eta_t$, $\xi_t^j = \xi_t^j(\eta_t)$. Matching terms with $\frac{d \xi_t^j}{\xi_t^j} = \mu_t^{\xi j} dt + \sigma_t^{\xi ja} dZ_t^a$,
\begin{align*}
    \mu_t^{\xi j} &= \frac{1}{\xi_t^j}\bb{\frac{\partial \xi_t^j}{\partial \eta_t}\mu_t^{\eta}\eta_t + \frac{1}{2}\frac{\partial^2 \xi_t^j}{\partial \eta_t^2}(\sigma_t^{\eta a}\eta_t)^2},\\
    \sigma_t^{\xi ja}  &= \frac{1}{\xi_t^j}\frac{\partial \xi_t^j}{\partial \eta_t}\sigma_t^{\eta a}\eta_t.
\end{align*}

Using Ito's product rule, the process of the value of the capital $q_t^a k_t^a$ is
\begin{align*}
    d(q_t^a k_t^a) &= q_t^a dk_t^a + k_t^a dq_t^a + dq_t^a dk_t^a,\\
    \text{or } \frac{d(q_t^a k_t^a)}{q_t^ak_t^a} &= \frac{dk_t^a}{k_t^a} + \frac{dq_t^a}{q_t^a} + \frac{dq_t^a}{q_t^a}\frac{dk_t^a}{k_t^a}\\
    &= (\mu^a + \Phi(\iota_t^a))dt + \sigma^a dZ_t^a + \mu_t^{qa} dt + \sigma_t^{qa} dZ_t^a + \sigma^a \sigma_t^{qa} dt + o(dZ_t^a)\\
    &= (\mu^a + \Phi(\iota_t^a) + \mu_t^{qa} + \sigma^a \sigma_t^{qa}) dt + (\sigma^a + \sigma_t^{qa})dZ_t^a,
\end{align*}
which is the capital gain rate.

Let $\alpha^a$ be the productivity rate. The dividend yield generated by the capital is $(\alpha^a-\iota_t^a)/q_t^a$. The return process is:
\begin{align}
    dr_t^{ka} &= \text{divident yield} + \text{capital gain rate} \nonumber\\
    &= \left(\underbrace{\mu^a + \Phi(\iota_t^a) + \mu_t^{qa} + \sigma^a \sigma_t^{qa} + \frac{\alpha^a-\iota_t^a}{q_t^a}}_{r_t^{ka}} \right) dt + (\sigma^a + \sigma_t^{qa})dZ_t^a.
\end{align}

\subsubsection{HJB Equation Optimality}
Consider the following HJB equation:
\begin{align}
    0 & = \sup_{w_t^{ja},\iota_t^{ja}, c_t^j} f(c_t^j, U_t^j) + \mathbb{E}_t(dU_t^j) \nonumber\\
     &= \sup_{w_t^{ja},\iota_t^{ja}, c_t^j} \bigg\{ \frac{f(c_t^j n^t_j, V_t^j)}{(\xi_t^j n_t^j)^{(1-\gamma^j)}} + \mu_t^{\xi j} +  \mu_t^{nj} - \frac{\gamma^j}{2} (\sigma_t^{n ja} )^2 - \frac{\gamma^j}{2} (\sigma_t^{\xi ja})^2 + (1-\gamma^j) \sigma_t^{\xi ja} \sigma_t^{n ja}  \bigg\},
\end{align}
where the normalized aggregator $f$ follows the recursive utility:
\begin{align}
f(c_t^jn_t^j, V_t^j)&= \left(  \frac{ 1- \gamma^j}{1-1/ \zeta^j} \right) \rho^j V_t^j  \left[ \left( \frac{c_t^jn_t^j}{ [(1-\gamma^j) V_t^j]^{1/(1-\gamma^j)}} \right)^{1-1/\zeta^j} - 1 \right].
\end{align}

The agents have Epstein-Zin preferences \cite{epstein-zin}. The value function can be verified as 
\begin{align}
    V_t^j = \frac{(n_t^j \xi_t^j)^{1-\gamma^j}}{1-\gamma^j}.
\end{align}

The HJB equation can be rewritten as
\begin{align*}
    &\frac{f(c_t^j n^t_j, V_t^j)}{(\xi_t^j n_t^j)^{(1-\gamma^j)}} + \mu_t^{\xi j} +  \mu_t^{nj} - \frac{\gamma^j}{2} (\sigma_t^{n ja} )^2 - \frac{\gamma^j}{2} (\sigma_t^{\xi ja})^2 + (1-\gamma^j) \sigma_t^{\xi ja} \sigma_t^{n ja}\\
    &= \frac{\rho^j}{1-1/\zeta^j} \left[ \left( \frac{c_t^j}{\xi_t^j} \right)^{1-1/\zeta^j} - 1 \right] + \mu_t^{\xi j} + \mu_t^{nj} - \frac{\gamma^j}{2} (\sigma_t^{n ja} )^2 - \frac{\gamma^j}{2} (\sigma_t^{\xi ja})^2 + (1-\gamma^j) \sigma_t^{\xi ja} \sigma_t^{n ja}.
\end{align*}

Given $\mu_t^{nj}=r_t - c_t^j + w_t^{ja}(r_t^{ka} - r_t)$ and $\sigma_t^{nja} = w_t^{ja}(\sigma^a + \sigma_t^{qa})$ from the budget constraint:
\begin{align}
    \frac{dn_t^j}{n_t^j} &= \left(\underbrace{r_t - c_t^j + w_t^{ja} (r^{ka}_t - r_t)}_{\mu_t^{nj}} \right) dt + \underbrace{w_t^{ja} (\sigma^a + \sigma_t^{qa})}_{\sigma_t^{nja}}dZ_t^a,
\end{align}
we get the HJB equation:
\begin{align*}
    F(w_t^{ja},\iota_t^{ja}, c_t^j) &= \frac{\rho^j}{1-1/\zeta^j} \left[ \left( \frac{c_t^j}{\xi_t^j} \right)^{1-1/\zeta^j} - 1 \right] + \mu_t^{\xi j} + r_t - c_t^j + w_t^{ja}(r_t^{ka} - r_t) \\
    & - \frac{\gamma^j}{2} (w_t^{ja})^2(\sigma^a + \sigma_t^{qa})^2 - \frac{\gamma^j}{2} (\sigma_t^{\xi ja})^2 + (1-\gamma^j) \sigma_t^{\xi ja} w_t^{ja}(\sigma^a + \sigma_t^{qa})\\
    &= \frac{\rho^j}{1-1/\zeta^j} \left[ \left( \frac{c_t^j}{\xi_t^j} \right)^{1-1/\zeta^j} - 1 \right] + \mu_t^{\xi j} + r_t - c_t^j \\
    & + w_t^{ja}\bb{\mu_t^{qa} + \mu^a + \Phi(\iota_t^a) + \sigma^a \sigma_t^{qa} + \frac{\alpha^a - \iota_t^a}{q_t^a} - r_t}\\
    & - \frac{\gamma^j}{2} (w_t^{ja})^2(\sigma^a + \sigma_t^{qa})^2 - \frac{\gamma^j}{2} (\sigma_t^{\xi ja})^2 + (1-\gamma^j) \sigma_t^{\xi ja} w_t^{ja}(\sigma^a + \sigma_t^{qa})
\end{align*}

The first-order condition (FOC) of $F(w_t^{ja},\iota_t^{ja}, c_t^j)$ requires $\nabla F(w_t^{ja},\iota_t^{ja}, c_t^j) = \left( \frac{\partial F}{\partial w_t^{ja}}, \frac{\partial F}{\partial \iota_t^{ja}}, \frac{\partial F}{\partial c_t^j} \right)^T = 0$. This gives the necessary condition for optimality as:
\begin{align*}
    \frac{\partial F}{\partial w_t^{ja}} &= (r_t^{ka}-r_t) - \gamma^j w_t^{ja}(\sigma^a + \sigma_t^{qa})^2 + (1-\gamma^j)\sigma_t^{\xi ja} (\sigma^a + \sigma_t^{qa}) = 0;\\
    \frac{\partial F}{\partial \iota_t^{ja}} &= w_t^{ja} \bb{\Phi'(\iota_t^a) - \frac{1}{q_t^a}} = 0;\\
    \frac{\partial F}{\partial c_t^j} &= \frac{\rho^j}{1-1/\zeta^j} \frac{1}{(\xi_t^j)^{1-1/\zeta^j}} (1-1/\zeta^j) (c_t^{j})^{-1/\zeta^j} - 1 = 0.
\end{align*}

$\frac{\partial F}{\partial w_t^{ja}} = 0$ gives $(r_t^{ka}-r_t) - \gamma^j w_t^{ja}(\sigma^a + \sigma_t^{qa})^2 + (1-\gamma^j)\sigma_t^{\xi ja} (\sigma^a + \sigma_t^{qa}) = 0$.

$\frac{\partial F}{\partial \iota_t^{ja}}=0$ while $w_t^{ja}\neq 0$ gives $\Phi'(\iota_t^a) - \frac{1}{q_t^a} = 0$.

$\frac{\partial F}{\partial c_t^j} = 0$ with $\zeta^j\approx 1$ gives $\rho^j (c_t^j)^{-1} - 1 \approx 0, c_t^j \approx \rho^j$.

We take the investment function as $\iota_t^a = \frac{q_t^a - 1}{\kappa}$. Then the functional form is $\Phi(\iota_t^a) = \frac{1}{\kappa} \log (1 + \kappa \iota_t^a)$.

We use agent $h$ to compute $r_t$, 
\begin{align}
    r_t &= r_t^{ka} - \gamma^hw_t^{ha}(\sigma^a + \sigma_t^{qa})^2 + (1-\gamma^h)\sigma_t^{\xi ha}(\sigma^a + \sigma_t^{qa}).
\end{align}

Then add sentiment to agent $i$,
\begin{equation}
    \hat{r_t^{ka}} = r_t^{ka} + \frac{\mu^O - \mu^a}{\sigma^a}(\sigma^a + \sigma_t^{qa}).
\end{equation}

It should satisfy the endogenous equation:
\begin{align}
    \hat{r_t^{ka}} - r_t &= \gamma^iw_t^{ia}(\sigma^a  + \sigma_t^{qa})^2 - (1-\gamma^i)\sigma_t^{\xi ia}(\sigma^{a}  + \sigma_t^{qa}).
\end{align}

\subsubsection{Market Clearing Conditions}
Consider the capital for each agent $k_t^h$ and $k_t^i$, their sum should equal the total capital $k_t^h + k_t^i = k_t^a$. Also, the wealth share is defined as $\eta_t = \frac{n_t^i}{n_t^h+n_t^i}$. Total budget should equal total capital gain, so $n_t^h+n_t^i = q_t^a k_t^a$. Then
\begin{align}
    \eta_t = \frac{n_t^i}{n_t^h+n_t^i} &= \frac{n_t^i}{q_t^a k_t^a}.
\end{align}

When market clears, quantity supplied is equal to the quantity demanded at the clearing price, and consumption from both types of agents equals surplus from the production.
\begin{align*}
    c_t^i n_t^i + c_t^h n_t^h &= (\alpha^a - \iota_t^a) (k_t^i + k_t^h)\\
    c_t^i n_t^i + c_t^h n_t^h &= (\alpha^a - \iota_t^a) k_t^a\\
    \frac{c_t^i n_t^i}{q_t^a k_t^a} + \frac{c_t^h n_t^h}{q_t^a k_t^a} &= \frac{\alpha^a - \iota_t^a}{q_t^a}\\
    c_t^i \eta_t + c_t^h (1-\eta_t) &= \frac{\alpha^a - \iota_t^a}{q_t^a}
\end{align*}

Consider the portfolio weights $w_t^{ja} = \frac{k_t^j q_t^a}{n_t^j}$.
\begin{align*}
    k_t^i + k_t^h &= k_t^a\\
    \frac{k_t^i}{k_t^a} + \frac{k_t^h}{k_t^a} &= 1\\
    \frac{k_t^i n_t^i q_t^a}{n_t^i q_t^a k_t^a} + \frac{k_t^h n_t^h q_t^a}{n_t^h q_t^a k_t^a} &= 1\\
    \frac{k_t^i q_t^a}{n_t^i} \eta_t + \frac{k_t^h q_t^a}{n_t^h} (1-\eta_t) &= 1\\
    w_t^{ia} \eta_t + w_t^{ha} (1-\eta_t) &= 1
\end{align*}

Therefore, from market clearing conditions, we get two endogenous equations:
\begin{align}
    \alpha^a - \iota_t^a&= (c_t^i \eta_t + c_t^h (1-\eta_t)) q_t^a\\
    1 &= w_t^{ia} \eta_t + w_t^{ha} (1-\eta_t)
\end{align}

\subsubsection{Model Details}
The definition of constant parameters are provided in Table~\ref{tab:1d-model-constant-params} and the definition of variables are provided in Table~\ref{tab:1d-model-variables}.

\begin{table}[!ht]
\caption{1D Economic Model Constant Parameters}\label{tab:1d-model-constant-params}

\vspace{.1in}
\centering
\begin{tabular}{lll}
\toprule
Parameter & Definition & Value\\
\hline\hline
$\gamma^j$ & relative risk aversion & $\gamma^i=2.0$, $\gamma^h=5.0$\\
$\rho^j$ & discount rate & $\rho^i=\rho^h=0.05$\\
$\zeta^j$ & intertemporal elasticity of substitution & $\zeta^i=\zeta^h=1.00005$\\
$\mu^a$ & growth rate of capital & 0.04\\
$\sigma^a$ & volatility of capital & 0.2\\
$\mu^O$ & sentiment factor & 0.04\\
$\alpha^a$ & productivity & 0.1\\
$\kappa$ & investment cost & 10000\\
\hline
\end{tabular}
\end{table}

\begin{table}[!ht]
\caption{1D Model Variables}\label{tab:1d-model-variables}

\vspace{.1in}
\centering
\begin{tabular}{ll}
\hline
Type & Definition\\
\hline\hline
State Variables & $\eta_t$ ($\eta$)\\
Agents & $\xi_t^i$, $\xi_t^h$\\
Endogenous Variables & $\mu_t^{\eta}$, $\sigma_t^{\eta a}$, $q_t^a$, $w_t^{ia}$, $w_t^{ha}$\\
\hline
\end{tabular}
\end{table}

Equations to define new variables:
\begin{align}
    \iota_t^a &= \frac{q_t^a - 1}{\kappa}\\
    \Phi(\iota_t^a) &= \frac{1}{\kappa}\log(1+\kappa\iota_t^a)\\
    c_t^j &= (\rho^j)^{\zeta^j}(\xi_t^j)^{1-\zeta^j}\\
    \sigma_t^{qa} &= \frac{1}{q_t^a} \frac{\partial q_t^a}{\partial \eta_t} \sigma_t^{\eta a} \eta_t\\
    \sigma_t^{nja} &= w_t^{ja}(\sigma^a + \sigma_t^{qa})\\
    \sigma_t^{\xi ja}  &= \frac{1}{\xi_t^j}\frac{\partial \xi_t^j}{\partial \eta_t}\sigma_t^{\eta a}\eta_t \\
    \sigma_t^{na} &= \eta_t\sigma_t^{nia} + (1-\eta_t)\sigma_t^{nha}\\
    \mu_t^{qa} &= \frac{1}{q_t^a} \left(\frac{\partial q_t^a}{\partial \eta_t} \mu_t^{\eta} \eta_t + \frac{1}{2} \frac{\partial^2 q_t^a}{\partial \eta_t^2} (\sigma_t^{\eta a} \eta_t)^2\right)\\ 
    r_t^{ka} &= \mu_t^{qa} + \mu^a + \Phi_t^a + \sigma^a\sigma^{qa} + \frac{\alpha^a - \iota_t^a}{q_t^a}\\
    r_t &= r_t^{ka} - \gamma^hw_t^{ha}(\sigma^a + \sigma_t^{qa})^2 + (1-\gamma^h)\sigma_t^{\xi ha}(\sigma^a + \sigma_t^{qa})\\
    \mu_t^{nj} &= r_t - c_t^j + w_t^{ja}(r_t^{ka} - r_t)\\
    \mu_t^{\xi j} &= \frac{1}{\xi_t^j}\left(\frac{\partial \xi_t^j}{\partial \eta_t}\mu_t^{\eta}\eta_t + \frac{1}{2}\frac{\partial^2 \xi_t^j}{\partial \eta_t^2}(\sigma_t^{\eta a}\eta_t)^2\right)\\
    \hat{r_t^{ka}} &= r_t^{ka} + \frac{\mu^O - \mu^a}{\sigma^a}(\sigma^a + \sigma_t^{qa})
\end{align}

Endogenous equations to pin down endogenous variables:
\begin{align}
    \mu_t^{\eta} &= (1-\eta_t)(\mu_t^{ni} - \mu_t^{nh}) +(\sigma_t^{na})^2  - \sigma_t^{nia}\sigma_t^{na}\\
    \sigma_t^{\eta a} &= (1-\eta_t)(\sigma_t^{nia} - \sigma_t^{nha})\\
    \hat{r_t^{ka}} - r_t &= \gamma^iw_t^{ia}(\sigma^a  + \sigma_t^{qa})^2 - (1-\gamma^i)\sigma_t^{\xi ia}(\sigma^{a}  + \sigma_t^{qa})\\
    1 &= w_t^{ia}\eta_t + w_t^{ha}(1-\eta_t)\\
    \alpha^a - \iota_t^a &= (c_t^i\eta_t + c_t^h(1 - \eta_t))q_t^a
\end{align}

HJB equations to solve for agent value functions:
\begin{align}
    0 = & \frac{\rho^i}{1-\frac{1}{\zeta^i}}\left( \left(\frac{c_t^i}{\xi_t^i} \right)^{1-1/\zeta^i}-1 \right) + \mu_t^{\xi i} +  \mu_t^{ni} - \frac{\gamma^i}{2}(\sigma_t^{nia})^2  - \frac{\gamma^i}{2}(\sigma_t^{\xi ia})^2 + (1-\gamma^i)\sigma_t^{\xi ia}\sigma_t^{nia}\\
    0 = & \frac{\rho^h}{1-\frac{1}{\zeta^h}}\left( \left(\frac{c_t^h}{\xi_t^h} \right)^{1-1/\zeta^h}-1 \right) + \mu_t^{\xi h} +  \mu_t^{nh} - \frac{\gamma^h}{2}(\sigma_t^{nha})^2  - \frac{\gamma^h}{2}(\sigma_t^{\xi ha})^2 + (1-\gamma^h)\sigma_t^{\xi ha}\sigma_t^{nha}
\end{align}

\subsubsection{Results}
\begin{figure}[!ht]
\centering
\begin{subfigure}[b]{0.3\linewidth}
\centering
\includegraphics[width=\linewidth]{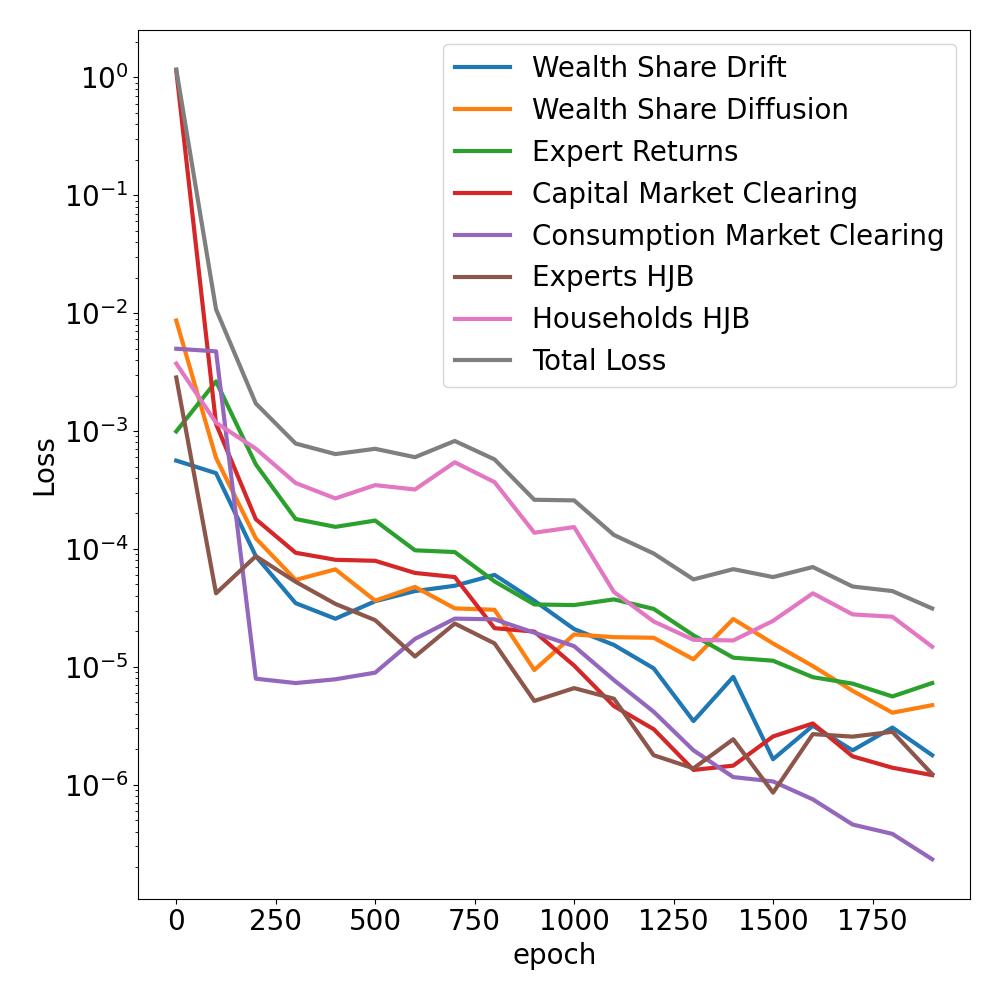}
\caption{Loss Progression}
\end{subfigure}
\hfill
\begin{subfigure}[b]{0.3\linewidth}
\centering
\includegraphics[width=\linewidth]{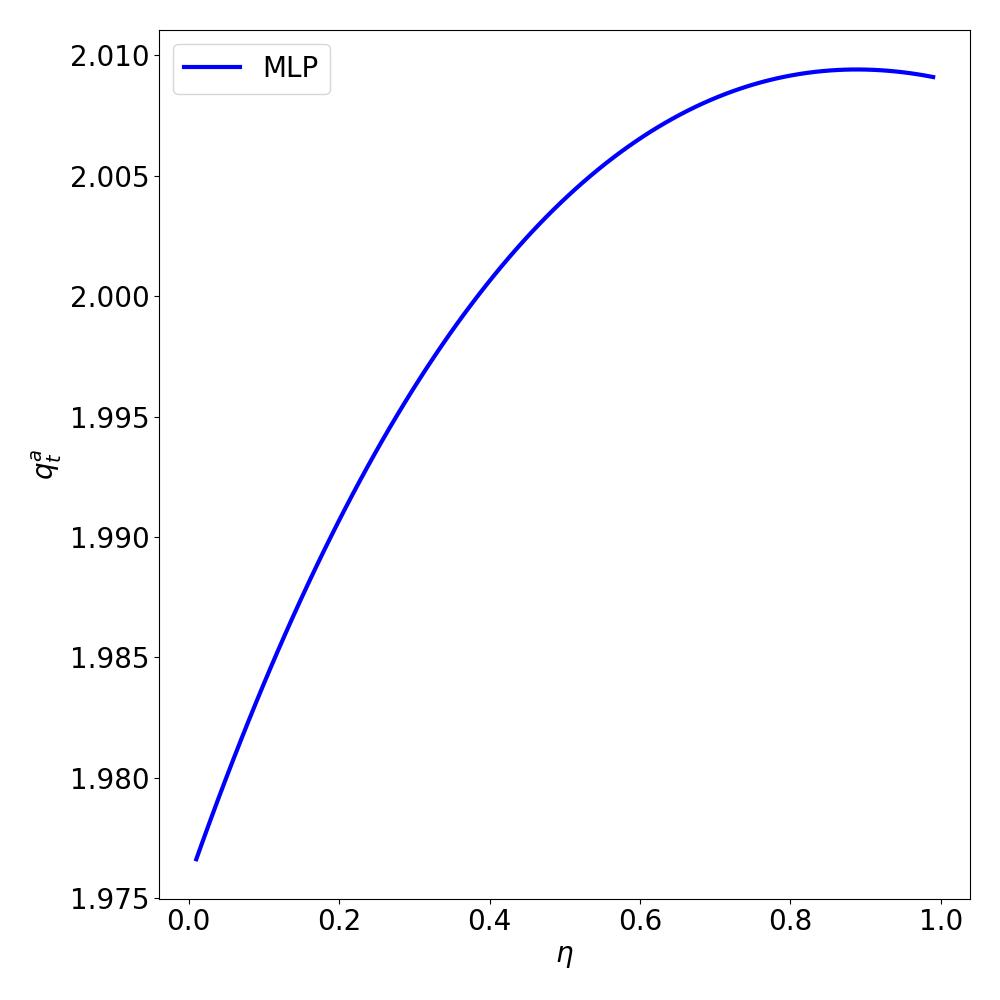}
\caption{Price}
\end{subfigure}
\hfill
\begin{subfigure}[b]{0.3\linewidth}
\centering
\includegraphics[width=\linewidth]{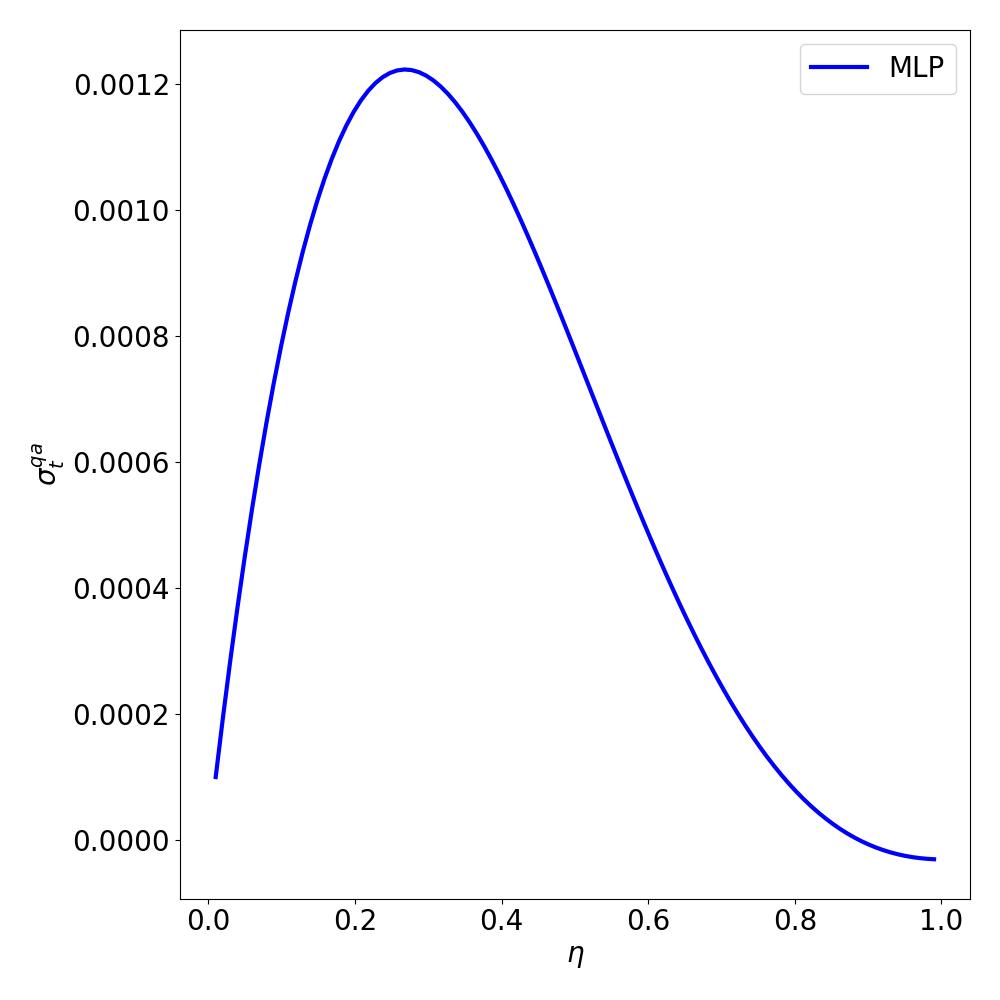}
\caption{Price Volatility}
\end{subfigure}

\vspace{.3cm}
\begin{subfigure}[b]{0.3\linewidth}
\centering
\includegraphics[width=\linewidth]{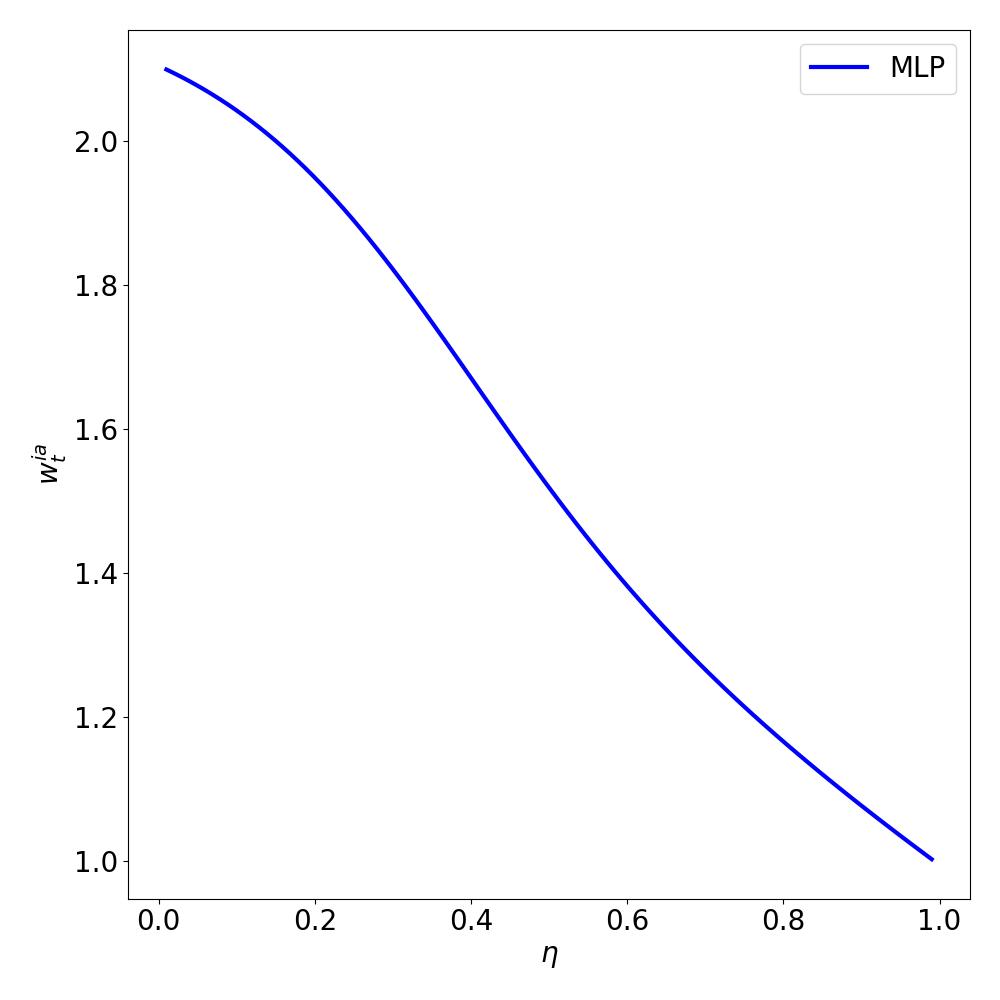}
\caption{Portfolio Choice: Experts}
\end{subfigure}
\hfill
\begin{subfigure}[b]{0.3\linewidth}
\centering
\includegraphics[width=\linewidth]{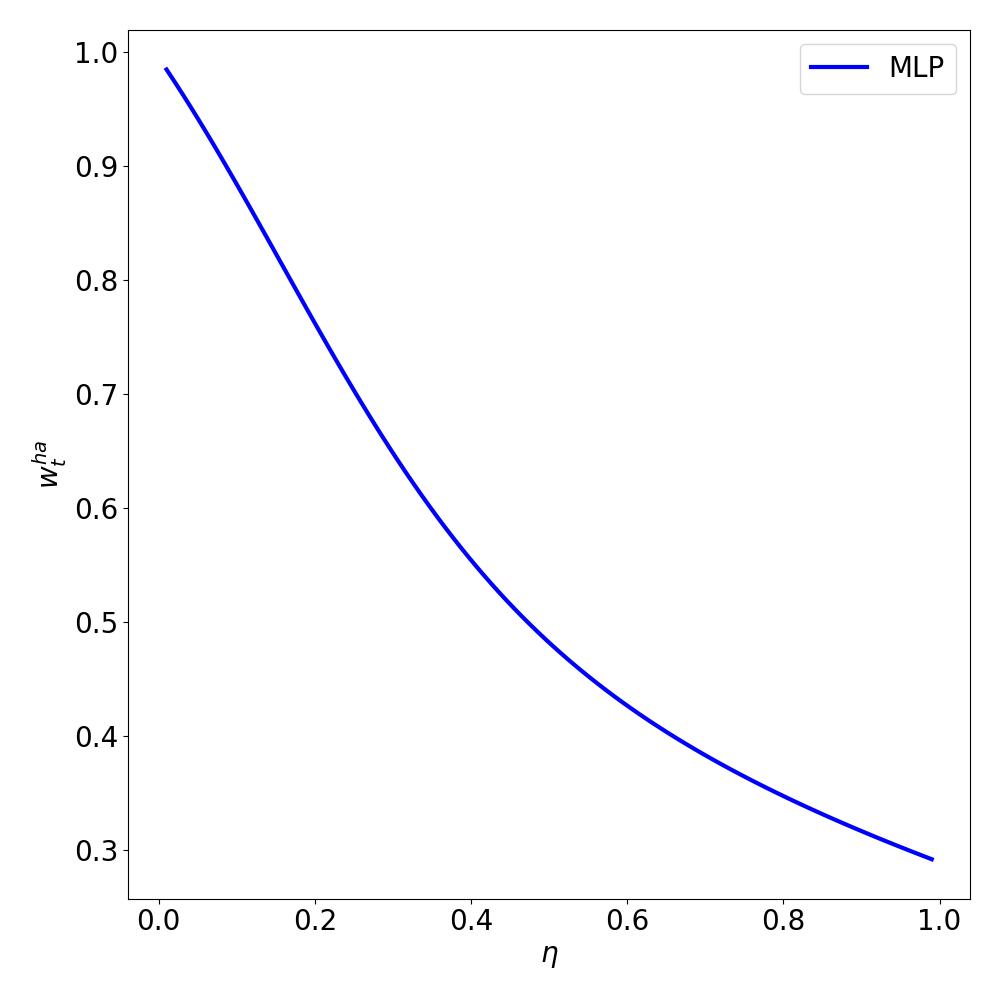}
\caption{Portfolio Choice: Households}
\end{subfigure}
\hfill
\begin{subfigure}[b]{0.3\linewidth}
\centering
\includegraphics[width=\linewidth]{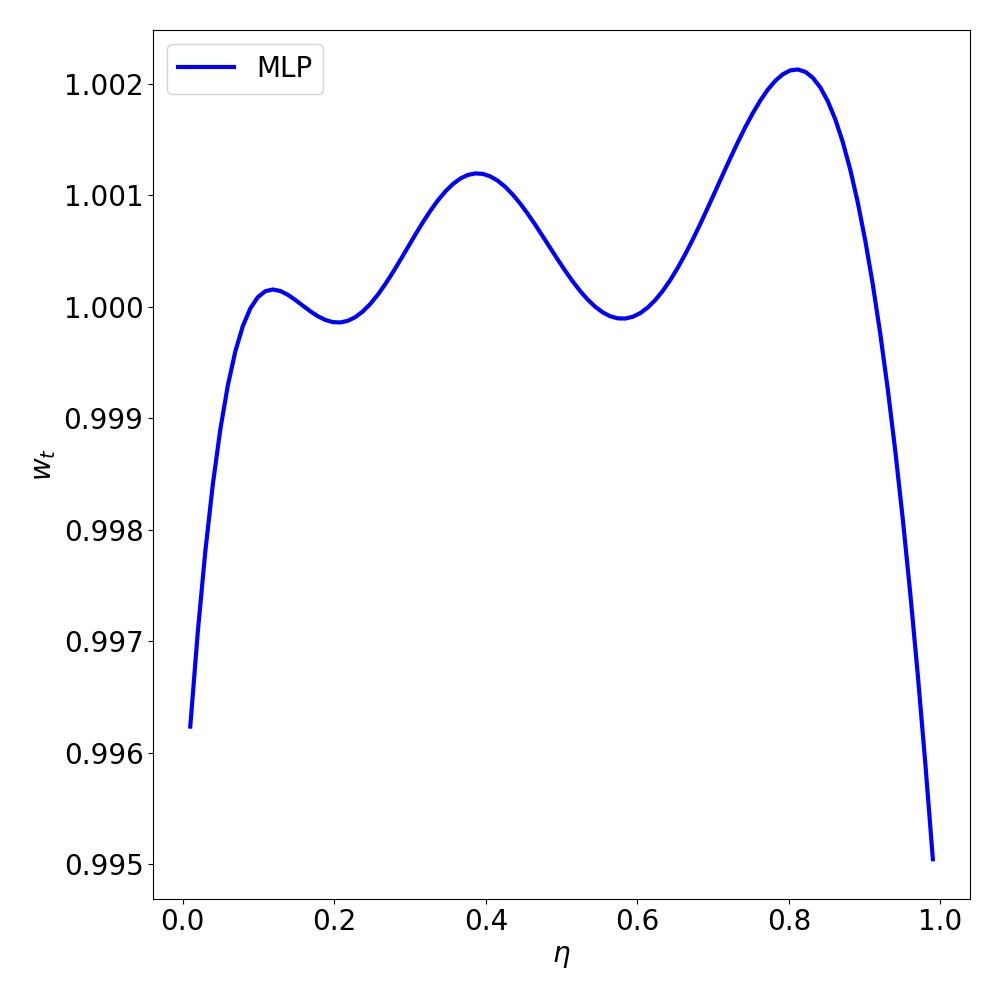}
\caption{Portfolio Choice Equilibrium}
\end{subfigure}
\caption{One-Dimensional Model Results}
\label{fig:1d-model}
\end{figure}

The agents and endogenous variables are configured as 4-layer MLPs with 30 hidden units per layer and tanh activation. $\xi_t^i$, $\xi_t^h$, and $q_t^a$ are constrained by SoftPlus (a smooth approximation to the ReLU function) to ensure positive outputs, thereby guaranteeing that the price of capital and agent wealth remain non-negative. The state variable $\eta_t$ is restricted to $[0.01, 0.99]$ to avoid instabilities at extreme share holdings. The system is trained for 2000 epochs using Adam optimizer with a learning rate of $10^{-3}$, under the endogenous and HJB constraints. 
As the share of wealth held by agent $i$ ($\eta_t$) increases, so does the price of capital $q_t^a$. The growth rate for capital price is decreasing, indicating diminishing returns.
The volatility of the price, represented by $\sigma_t^{qa}$, peaks when $\eta_t$ is around $0.3\sim 0.4$, reflecting the highest uncertainty. At extreme points, $\eta_t\to 0$ or $\eta_t\to 1$, $\sigma_t^{qa}\to 0$, indicating minimal price uncertainty when one type of agent holds all the wealth.
Both agents maintain a long position, as evidenced by the positive portfolio weights $w_t^{ia}$ and $w_t^{ha}$. When $\eta_t\sim 0$, only agent $h$ is contributing to the total capital, resulting in $w_t^{ha}(0)=1$. As $\eta_t$ increases, both weights decrease, with $w_t^{ia}$ decreasing more rapidly (from 2 to 1) than $w_t^{ha}$ (from 1 to 0.3). This ensures the market clearing condition, where the weighted sum of the capital portfolio held by intermediary $i$ and household $h$ always equals the total capital $k_t^a$. Figure~\ref{fig:1d-model} shows the results.

\subsection{Variable Parameters and Inverse Problems}
Consider the single-asset two-dimensional Black-Scholes equation. We are interested in finding solutions for various parameter values. To achieve this, we treat the parameters as pseudo state variables and include them as neural network inputs. This approach increases the dimensionality and complexity of the problem. However, once the neural network is trained, the solution becomes available for any point in the state and parameter spaces. Specifically, we solve the following four-dimensional problem with state variables $(S, t, \sigma, r) \in [0,1]\times [0,1]\times [0.1, 0.3]\times [0.01, 0.1]$:
\begin{align*}
    &\frac{\partial V}{\partial t} + r S \frac{\partial V}{\partial S} + \frac{\sigma^2}{2} S^2 \frac{\partial^2 V}{\partial S^2} - r V = 0,\\
    &V(S,T) = \max\bs{S - K, 0}
\end{align*}

\begin{figure}[!ht]
\centering
\begin{subfigure}[b]{0.45\linewidth}
\centering
\includegraphics[width=\linewidth]{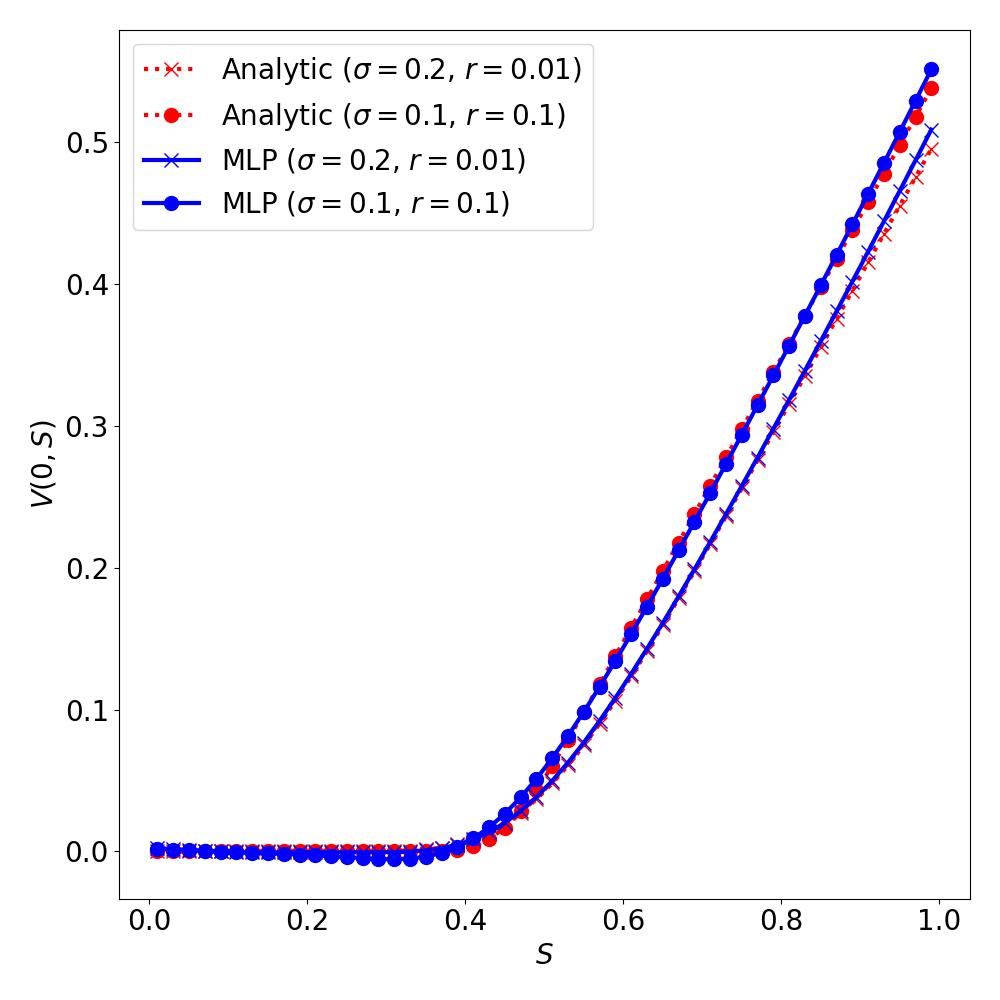}
\caption{Variable Parameters\label{fig:black-scholes-equation-variable-params}}
\end{subfigure}
\hfill
\begin{subfigure}[b]{0.45\linewidth}
\centering
\includegraphics[width=\linewidth]{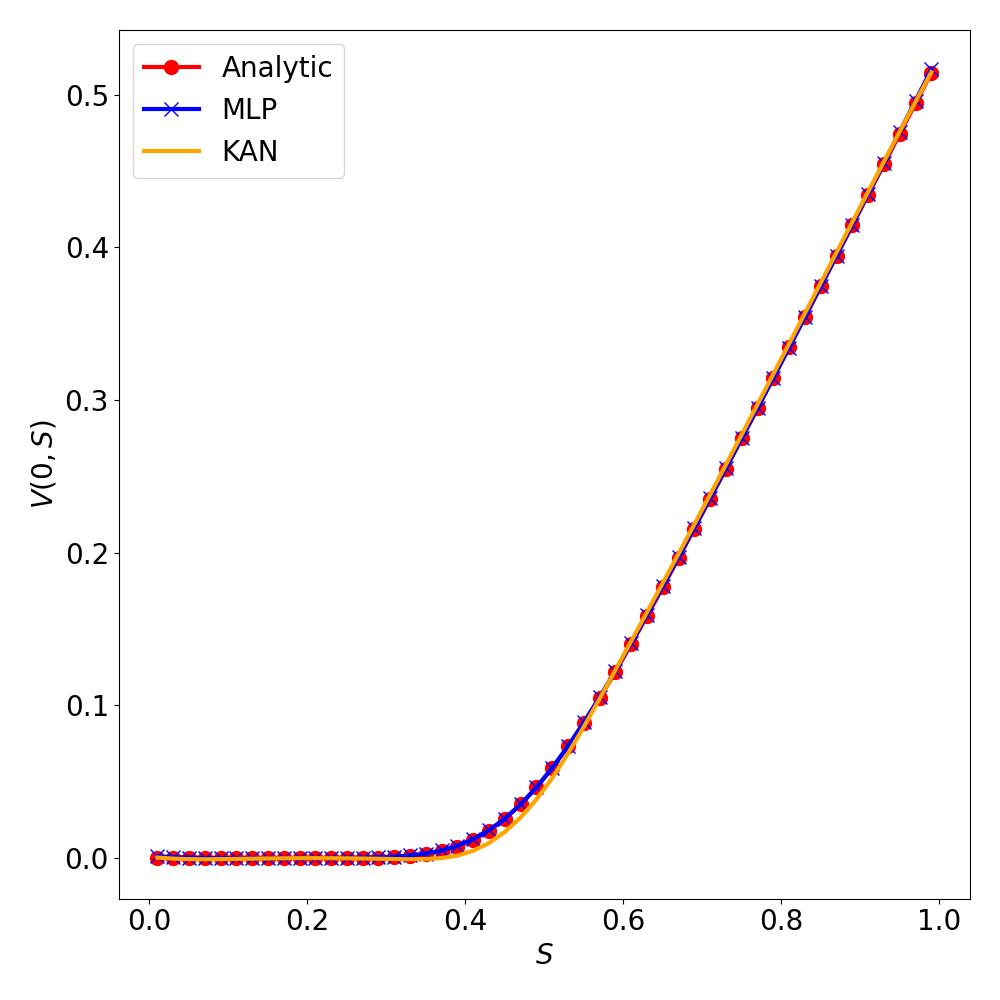}
\caption{Inverse Problems\label{fig:black-scholes-equation-inverse}}
\end{subfigure}
\caption{In (a), the fit of MLP (blue) models are shown for several parameters. The analytic solutions are plotted in red. In (b), the fit of MLP and KAN for the inverse problems are plotted against the analytic solution.}
\label{fig:black-scholes-equation-variable-params-inverse-problem}
\end{figure}

Figure~\ref{fig:black-scholes-equation-variable-params} illustrates the model fit for two parameter sets: $(\sigma=0.2, r=0.01)$ and $(\sigma=0.1, r=0.1)$, using a 4-Layer MLP with SiLU activation (2971 parameters). The MSEs are $1.50\times 10^{-5}$ and $2.59\times 10^{-5}$ respectively.

In some cases, the parameters themselves may be unknown, but partial observations are available. To address this, we can simultaneously solve for the exact solution and approximate the corresponding parameters. For instance, we set the ground truth values to $\sigma=0.2$ and $r=0.05$, with initial parameter guesses of $\hat{\sigma}=0.15$ and $\hat{r}=0.04$. The fit results are shown in Figure~\ref{fig:black-scholes-equation-inverse}. The parameters approximated by the MLP are $\sigma=0.19$ and $r=0.05$, while those approximated by the KAN are $\sigma=-0.1$ and $r=0.06$.

\end{document}